\documentclass[twoside,11pt]{article}
\usepackage{amsthm}
\usepackage[preprint]{jmlr2e}
\usepackage{xcolor}
\usepackage{amsmath}
\usepackage{booktabs}
\usepackage[bf]{caption} 
\usepackage{multirow}
\usepackage{wrapfig}
\usepackage{bm}

\DeclareMathOperator{\argmin}{argmin}

\newcommand{\R}{\mathbb{R}}

\usepackage[normalem]{ulem}

\newtheorem*{leakage_criterion*}{Leakage Criterion}

\usepackage{lastpage}
\jmlrheading{ }{ }{1-\pageref{LastPage}}{1/21; Revised 5/22}{9/22}{21-0000}{Parisini, Chakraborti, Harbron, MacArthur, Banerji}

\ShortHeadings{Leakage in Concept-Based Models}{Parisini, Chakraborti, Harbron, MacArthur, Banerji}
\firstpageno{1}

\begin{document}

\title{Leakage and Interpretability in Concept-Based Models}

\author{\name Enrico Parisini* \email eparisini@turing.ac.uk \\
       \addr The Alan Turing Institute \\
       London, UK \\
       \addr Kings College London\\
       London, UK 
       \AND
       \name Tapabrata Chakraborti \email tchakraborty@turing.ac.uk\\
       \addr The Alan Turing Institute\\
       London, UK\\
        \addr University College London\\
       London, UK
       \AND
       \name Chris Harbron \email  chris.harbron@roche.com\\
       \addr Roche Pharmaceuticals\\
       Welwyn Garden City, UK
       \AND
       \name Ben D. MacArthur* \email bmacarthur@turing.ac.uk\\
       \addr The Alan Turing Institute\\
       London, UK
       \AND
       \name Christopher R.S. Banerji* \email cbanerji@turing.ac.uk \\
       \addr The Alan Turing Institute\\
       London, UK\\
       \addr Kings College London\\
       London, UK       
       \AND
       \rm{*EP, BDM and CRSB should be considered joint corresponding authors. BDM and CRSB are joint senior authors.}}

\editor{ }

\maketitle

\begin{abstract}%
Concept-based Models aim to improve interpretability by predicting high-level intermediate concepts, representing a promising approach for deployment in high-risk scenarios. 
However, they are known to suffer from information leakage, whereby models exploit unintended information encoded within the learned concepts.
We introduce an information-theoretic framework to rigorously characterise and quantify leakage, and define two complementary measures: the concepts-task leakage (CTL) and interconcept leakage (ICL) scores. We show that these measures are strongly predictive of model behaviour under interventions and outperform existing alternatives. Using this framework, we identify the primary causes of leakage and, as a case study, analyse how it manifests in Concept Embedding Models, revealing interconcept and alignment leakage in addition to the concepts-task leakage present by design. Finally, we present a set of practical guidelines for designing concept-based models to reduce leakage and ensure interpretability.
\end{abstract}


\bigskip

\section{Introduction}

Explainability and transparency are essential aspects of model design, especially in high-risk scenarios such as biomedical applications.
Concept Bottleneck Models (CBMs) \citep{Kumar09, Lampert09,  koh20a} stand out in the landscape of interpretable models as they learn a set of high-level intermediate concepts associated with an outcome, in a manner which facilitates human oversight \citep{banerji2025train}. CBMs first predict the level of activation of each concept for a given input data point, then use these activations to predict a task.
Under the assumption that the set of annotated concepts are appropriately defined and learned by the model, a CBM is thus \emph{interpretable}, in the sense that its output can be deterministically and uniquely traced to the level of activation of semantically meaningful concepts (e.g. ``presence vs. absence" in the binary case). Their structure allows for direct human supervision and intervention, as opposed to \emph{post hoc} explainability approaches.\footnote{Other model classes that offer a degree of interpretability, beyond concept-based models, include prototype-based networks \citep{ProtoPNet} and B-cos networks \citep{Bcos}.}

In earlier CBMs, concept activations are real numbers (either probabilities or logits); more recently there have been efforts to generalise this approach to concept vector representations. Concept Embedding Models \citep[CEMs,][]{CEMs} are often regarded as state of the art in this respect, as they are capable of achieving higher task performance than CBMs, frequently matching that of end-to-end models.

A trained concept-based model, however, may be significantly less interpretable than expected, despite presenting the illusion of being so.
A key reason for this is the phenomenon of \emph{information leakage} \citep{Kazhdan2021IsDA, margeloiu2021, Mahinpei2021PromisesAP, Addressing_leakage, Makonnen25}, sometimes also described as shortcut learning \citep{enouen2025debugging}. Leakage occurs when a poorly designed concept-based model uses additional information to attain a higher task accuracy than a well-designed concept-based model, at the cost of undermining concept interpretability.
Storing such additional information in the learnt concepts is beneficial for the model with respect to loss function minimisation, as it typically allows the final head to leverage this information, attaining higher task performance while preserving high accuracy in concept predictions.
This ultimately results in a model that is not interpretable in the sense defined above. 

As an example, consider a concept-based model trained to make treatment recommendations for cancer patients from local biopsy data, using concepts routinely extracted from such data, such as tumor grade (i.e. how aggressive the tumor appears under the microscope). However, when metastases are present, treatment decisions depend on body-level information that cannot be inferred from local biopsy concepts alone. A concept-based model trained only on local biopsies may therefore be misspecified and, during training, can learn to encode task-relevant information (such as the presence of metastases) within its concept activations. In such a setting, a clinician can neither meaningfully supervise the model’s predictions (since decisions rely on hidden information not captured by the exposed concepts) nor safely intervene on an incorrectly predicted concept, because modifying that concept would change the output in a way that is causally unrelated to its clinical meaning. A concept-based model which displays leakage may therefore be unsafe for critical decision-making applications, such as clinical scenarios. 

Beyond leakage, several additional risks and limitations affect current concept-based architectures \citep{sinha2025comprehensivesurveyriskslimitations}. An important line of work in this context concerns robustness and security against adversarial attacks \citep{Robust1, Baniecki2025}, and recent studies have also shown that leakage itself can facilitate backdoor attacks that influence model predictions \citep{lai2025catconceptlevelbackdoorattacks}. The practical human-in-the-loop integration of interpretable models also raises several important challenges, including interventions, user interfaces, and interactive debugging \citep{Ramaswamy2022OverlookedFI, bontempelli2023conceptlevel}.

\paragraph{Contributions.} We propose and test a framework based on information theory to assess the interpretability of concept-based models. In particular, we provide the first precise definition of information leakage, and identify two main ways it can manifest - as information leaking into each learnt concept from either the class label, or the other concepts (Section \ref{sec_leakage}). After summarising the limitations of current leakage measures (Section \ref{sec_existing_measures_interpretability}), we accordingly define two measures, the concepts-task leakage (CTL) and the interconcept leakage (ICL) scores (Section \ref{sec_def_CTL_ICL}), and in Section \ref{sec_robustness} we show that they
\begin{itemize}
    \item are highly predictive of differences in performance upon intervention across models (the only robust indicator of interpretability in a CBM);
    \item substantially outperform previously proposed measures (including the OIS and NIS defined in \cite{Espinosa_Zarlenga_2023}) across models and datasets.
\end{itemize}
We then employ these measures to identify and assess the primary causes of leakage in CBMs (Section \ref{sec_causes_of_leakage}), including: over-expressive concept representations; an incomplete set of annotated concepts; a misspecified final head; and insufficient concept supervision. Building on this novel information-theoretic framework, in Section \ref{sec_interpretability_CEMs} we confirm that concepts-task leakage is structurally embedded in CEMs to boost task performance, consistent with their original architectural design and prior understanding \citep{CEMs, IntCEMs, zarlenga2025avoiding}. We further establish the presence of interconcept leakage in CEMs and identify alignment leakage, a novel form of concepts-task leakage that arises in architectures trained under concept-level interventions such as CEMs.
Finally, in Section \ref{sec_lessons} we propose a set of guidelines for the design of concept-based models based on preventing leakage, and we argue that evaluating leakage is a crucial step of model development to ensure interpretability. 

The complete codebase for computing leakage scores, evaluating concept-based models, and reproducing all experimental results is publicly available at \url{https://github.com/enricoparisini/xai-concept-leakage}.

\section{Concept bottleneck models}

\begin{figure}
\centering
    \includegraphics[width=1\textwidth]{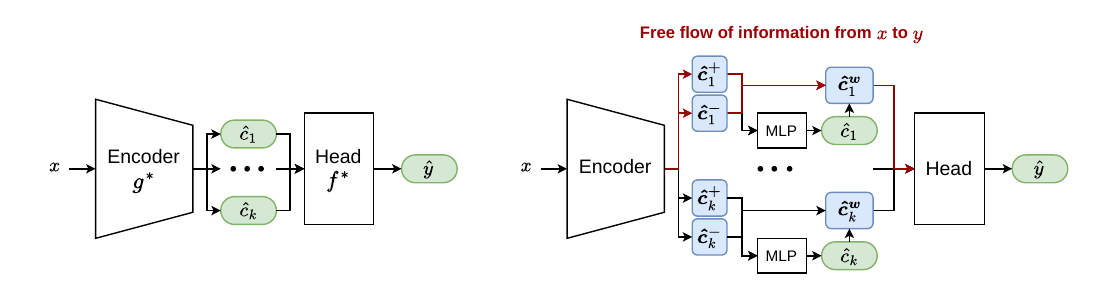}
    \caption{Scheme of CBM (\textit{left}) and CEM (\textit{right}) architectures. Quantities with green and blue backgrounds are predicted scalars and vectors respectively.}
    \label{figure_scheme_CBM_CEM}
\end{figure}

Consider a dataset $\left\{ \bm{x}^{(n)}, c^{(n)}, y^{(n)}\right\}_{n=1}^N$ with $N$ observations, where $\bm{x}^{(n)} \in \R^m$ is the input, $c^{(n)} \in \{0,1\}^k$ represents a set of $k$ annotated concepts, and $y^{(n)} \in \{1, \dots,  \ell\}$ is the target label.
A CBM (Figure \ref{figure_scheme_CBM_CEM}) is the composition of a concept encoder $g^*$ mapping the input to the predicted concept activations $\hat{c} = g^*(x)$, and a classifier head $f^*$ which maps the concepts to the predicted label $\hat{y} = f^*(\hat{c})$. This framework easily extends to regression tasks $y$ and to categorical or continuous concepts.

The presence of a concept bottleneck displaying the predicted concept activations for each input allows a human overseer to directly supervise the model reasoning without resorting to post-hoc approaches to explainability such as CAM \citep{CAM}, GradCAM \citep{GradCAM}, LIME \citep{LIME} or SHAP scores \citep{SHAP}.
Additionally, this setup enables human interventions: if one or more predicted concepts are identified as incorrect for a given input $\bm{x}^{(n)}$, a human overseer can update their values according to their expertise, resulting in an improved task prediction. Recent work argues that CBM interventions can be interpreted through a causal lens as approximate interventions on concept variables in a concept-to-label causal model \citep{shin2022a}. Counterfactual CBMs \citep{dominici2025counterfactual} make this connection to causality more explicit by learning to generate concept-level counterfactuals.

There are three possible strategies to train CBMs as introduced by \cite{koh20a}: 
\begin{itemize}
    \item \emph{independent training}, where the concept encoder and the final head are trained independently on ground-truth data. In this case,
    \begin{equation}
        g^* = \argmin_g \mathcal{L}_c(g(x), c)\,, \qquad 
        f^* = \argmin_f \mathcal{L}_y(f(c), y)\,,
    \end{equation}
    where $\mathcal{L}_c$ and $\mathcal{L}_y$ are suitable loss functions for the concepts and the label respectively;
    \item \emph{sequential training}, where the concept encoder is trained on ground-truth data, while the final head is trained on the predicted concepts. In this case,
    \begin{equation}
        g^* = \argmin_g \mathcal{L}_c(g(x), c)\,, \qquad 
        f^* = \argmin_f \mathcal{L}_y(f(g^*(x)), y)\,;
    \end{equation}
    \item \emph{joint training}, where the model is trained end-to-end. In this case,
    \begin{equation}
        g^*, f^* = \argmin_{g, f} \left[ \lambda \,\mathcal{L}_c(g(x), c) + \mathcal{L}_y(f(g(x)), y) \right]\,, \label{eq_loss_joint_CBMs}
    \end{equation}
    where $\lambda\geq0$ is a hyperparameter that quantifies the relative importance of concept and task learning. $\lambda = 0$ corresponds to an end-to-end model.
\end{itemize}
Common choices for concept encoding include binary probabilities, soft probabilities or logits. We will refer to \emph{hard} CBMs as independently trained models with binary concept encoding, while \emph{soft} CBMs denote jointly trained models with soft probability-based concept activations. Additionally, we will refer to jointly trained models using logit-based concept encoding as \emph{logit} CBMs.
 
More recently, CEMs (Figure \ref{figure_scheme_CBM_CEM}) were proposed in \cite{CEMs}. Here a pair of vectors $\left(\bm{\hat{c}^+}_i, \bm{\hat{c}^-}_i \right) \in \R^{2d}$ is predicted for each input and concept $c_i$, $i = 1 \dots k$ (where the hyperparameter $d$ indicates the embedding dimension). This pair of vectors is subsequently used to predict the activation 
$\hat{c}_i$ as a soft probability. The weighted vectors 
$\bm{\hat{c}^{w}}_i = \hat{c}_i \, \bm{\hat{c}^+}_i + (1 - \hat{c}_i)\,\bm{\hat{c}^-}_i$ are then constructed and concatenated before being fed into the final head to predict the label.
In this setup $\bm{\hat{c}^+}_i$ and $\bm{\hat{c}^-}_i$ are meant to represent concept $i$ being active or inactive respectively. Learning a two-fold representation for each concept enables interventions, by modifying the concept activations $\hat{c}_i$ and thus selecting either $\bm{\hat{c}^+}_i$ or $\bm{\hat{c}^-}_i$ as an input for the final head.

Training in CEMs is performed end-to-end. Similarly to joint CBMs, the loss function consists of reconstruction losses for both the concept activations and the class label, weighted by a parameter $\lambda \geq 0$ as in \eqref{eq_loss_joint_CBMs}. To make them more receptive to interventions (i.e. to realize any improvements in task performance upon intervention), CEMs are exposed to interventions during training: each activation $\hat{c}_i$ is set to its ground-truth value $c_i$ with probability defined by a parameter $p_{int} \in (0,1)$. 

Reasoning in CEMs takes place at the level of the vectors $\bm{\hat{c}^{w}}_i$, as opposed to CBMs where it is based on concept activations $\hat{c}_i$. As high-dimensional objects trained only to be predictive of concepts $c_i \in \{0,1\}$, the vectors 
$\bm{\hat{c}^{w}}_i \in \mathbb{R}^d$ are expressive representations capable of encoding large amounts of information about the task label. 
We demonstrate in Section \ref{sec_interpretability_CEMs} that this additional task-relevant information is the main driver of task performance. While \cite{CEMs} present CEMs as interpretable models that resolve the accuracy-interpretability trade-off, we argue that they are still affected by this trade-off: indeed, they favour accuracy over interpretability. Although requiring a case-by-case evaluation, other CEM-based architectures, such as \cite{IntCEMs, Hulp24, EvidentialCBMs24, SemisupervisedCBM24, VLG_CBM24}, are likely to face similar interpretability challenges.

\section{Definition of leakage}
\label{sec_leakage}

The interpretability of CBMs and CEMs assumes that learnt concepts are aligned with ground-truth human-interpretable concepts. However, on general grounds that is not the case. During training, concept-based models often encode additional input information into the learned concepts to enhance task accuracy. When this happens \emph{information leakage} occurs \citep{Kazhdan2021IsDA, margeloiu2021, Mahinpei2021PromisesAP, Addressing_leakage, Lockhart2022TowardsLT, ragkousis24}, and the model does not rely solely on the value of concepts
for its task prediction, but also on leaked information, which hinders interpretability. The reasoning of the model thus becomes effectively obscure to human overseers, to a degree specified by the amount of leakage. 

\begin{figure}
    \centering
    \includegraphics[width=0.8\textwidth]{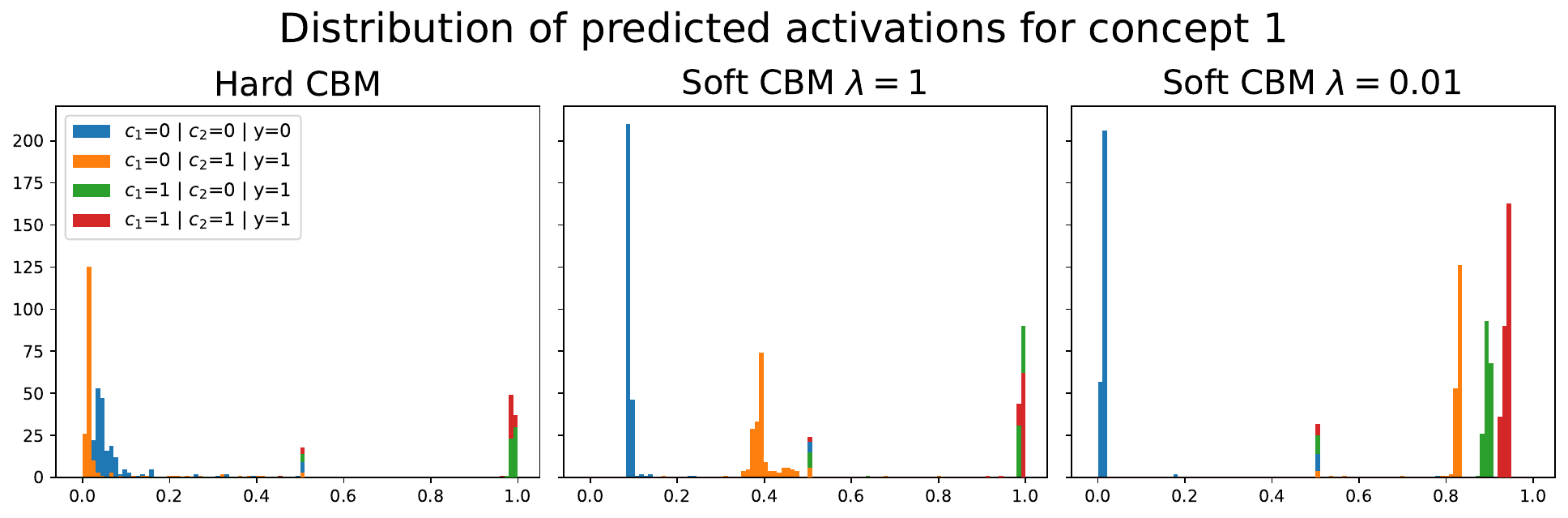}
    \caption{Distributions of predicted activations for concept 1 on the test set in CBMs with increasing leakage from left to right. Colours are based on the ground-truth values of concepts and task label, and to improve visibility, bars with lower counts are rendered in front of those with higher counts.}
    \label{figure_TT_leakage_concept_distributions}
\end{figure}

In practice, information leakage manifests as additional structure arising in the learnt concept distributions, as shown in
Figure \ref{figure_TT_leakage_concept_distributions}. In this synthetic example, models must learn two binary concepts $c_1,\,c_2$ and the binary task is to predict the function $y = c_1 \, \texttt{OR} \;c_2$ (see Appendix \ref{App_experiments_Datasets} for more details). We compare the distributions of activations for the first concept by a hard CBM and two soft CBMs, with either intermediate ($\lambda = 1$) or low ($\lambda = 0.01$) concept supervision. By definition no leakage can be present in hard CBMs, since concept learning is independent of task learning, and concepts are converted to binary values before being passed to the final head, preventing leaked information. Soft CBMs with lower concept supervision are instead more prone to leakage as accurate concept learning is not enforced during training
\citep[see][and our results in Section \ref{sec_causes_of_leakage}]{koh20a, Kazhdan2021IsDA}.

Neglecting the peak close to 0.5, which corresponds to observations the model is most uncertain about, the concept distribution in the hard CBM is very close to the binary ground-truth distribution for concept 1. In the soft CBM with $\lambda = 1$ additional structure is learnt based on the value of the other concept and of the class label, while for $\lambda = 0.01$ the concept itself is a very strong predictor of the value of both the class label and the other concept. In particular, the distribution is more similar to the ground-truth binary task distribution than to the concept distribution. 

Based on these observations, we identify two main types of leakage that concept-based models suffer from, and propose the following classification:

1) \emph{Concepts-task leakage:} additional information about the task is stored into the learnt concepts. For training to be successful, the annotated concepts must be predictive of the task label as measured by the ground-truth mutual information (MI, see more below) between each concept and the task label. If concepts-task leakage is present, then the learned concepts-task MI will be higher than ground-truth concepts-task MI. This is the predominant type of leakage and the main cause of non-interpretability, as it directly allows the model to achieve a better task performance (see experiments in Section \ref{sec_robustness}).

2) \emph{Interconcept leakage:} additional information about the other concepts is stored into each learned concept. In a given dataset, concepts are predictive of the value of other concepts as determined by their ground-truth pair-wise MI. This type of leakage manifests as learned concepts being more predictive of the value of the other learned concepts, resulting in a learned interconcept MI being higher than the ground-truth. This effect is usually a secondary factor of non-interpretability (see experiments in Section \ref{sec_robustness}), but it may be beneficial for models as an internal error-correcting tool: interconcept leakage provides redundant pathways for the model to attain high task performance even when concept predictions are poor \citep{ConceptCorrelation}.
 
This classification, highlighting the information-theoretic nature of leakage, forms the foundation of our quantitative approach to detect and measure it.

\section{Existing measures of interpretability and their shortcomings}
\label{sec_existing_measures_interpretability}

Previous works \citep{koh20a, Kazhdan2021IsDA, margeloiu2021, Mahinpei2021PromisesAP, Addressing_leakage, Espinosa_Zarlenga_2023} have proposed several measures of leakage, however none take into account its information-theoretic nature outlined in Section \ref{sec_leakage}.
As we highlight in this section, these existing measures consequently exhibit inherent limitations in sensitivity and robustness, limiting their usefulness in most practical scenarios.

\paragraph{Concept performance metrics.} Although performance metrics such as concept accuracy, F1 score and AUC are general quality indicators of concept learning, they are not sensitive to more subtle effects undermining interpretability, such as leakage. In particular, two models may exhibit the same evaluation scores while considerably differing in terms of leakage. This is apparent from the example in Figure \ref{figure_TT_leakage_concept_distributions}, where a classification threshold of 0.5 yields essentially the same concept accuracy and F1 score for the hard and $\lambda = 1$ soft CBM, while the latter encodes a non-trivial amount of additional structure. 
Concept AUC is a finer measure of interpretability than accuracy and F1 score as it captures more information about the concept distribution, and in this simple example it is able to discriminate between those two models. However it is not generally sensitive enough to quantify leakage in more complex setups as we illustrate in the following experiments.

\paragraph{OIS and NIS.} These scores were defined in \cite{Espinosa_Zarlenga_2023} to capture leakage in concept-based models. The Oracle Impurity Score (OIS) is obtained by training $k(k-1)$ additional neural networks $\psi_{i,j}$ (typically 2-layer perceptrons of modest size), whose objective is to predict the value of the ground-truth concept $j$ from the learnt representation of concept $i$. The $k\times k$ impurity matrix defined as 
${\pi_{ij}(\hat{c}, c) = \textrm{AUC}\left(\psi_{i,j}(\hat{c}_i), c_j \right)}$
is meant to estimate the predictivity of each learnt concept representation with respect to any other across the dataset. The OIS is the average difference between the predictivities of learnt and ground-truth concepts over pairs of concepts, 
\begin{equation}
    \text{OIS}(\hat{c}, c) = \frac{2}{k} \left|| \pi(\hat{c}, c) - \pi(c, c) \right||_F\,,
\end{equation}
where $|| \cdot ||_F$ indicates the Frobenius norm, and the normalisation ensures $ 0 \leq \text{OIS} \leq 1$. The OIS is thus intended to provide an estimate of interconcept leakage -- although, it is unable to account for concepts-task leakage, which is the prevalent effect as we illustrate in Section \ref{sec_robustness}. Although demonstrated to be superior to other concept-quality metrics in \cite{Espinosa_Zarlenga_2023}, the OIS is subject to additional limitations that hinder its applicability. In particular, in Section \ref{sec_robustness} we demonstrate that:
\begin{enumerate}
    \item its value is typically non-vanishing and relatively high for hard CBMs, indicating bias and a lack of sensitivity;
    \item due to the intrinsic stochasticity in its definition which involves training neural networks, it is extremely variable when repeatedly evaluated for a given model at test time. The resulting broad confidence intervals typically prevent drawing any statistically significant conclusion when comparing models with different amounts of leakage, ultimately limiting the utility of this score for model design.
\end{enumerate}

The Niche Impurity Score (NIS) was defined to capture leakage manifesting as additional information about concept $j$ stored into a joint subset of the learnt representations $\{i_1, \dots, i_p\}$ not containing concept $j$ \citep[see][for details]{Espinosa_Zarlenga_2023}. This score is meant to go beyond pair-wise interconcept leakage and estimate (generally subordinate) higher-order effects. As illustrated in Appendix \ref{App_NIS}, the NIS is generally non-vanishing and very high for hard CBMs, and moreover it appears to be anticorrelated with intervention performance and leakage, casting doubt on its suitability as a measure of leakage.

\begin{wrapfigure}{r}{0.45\textwidth}
\includegraphics[width=0.85\linewidth]{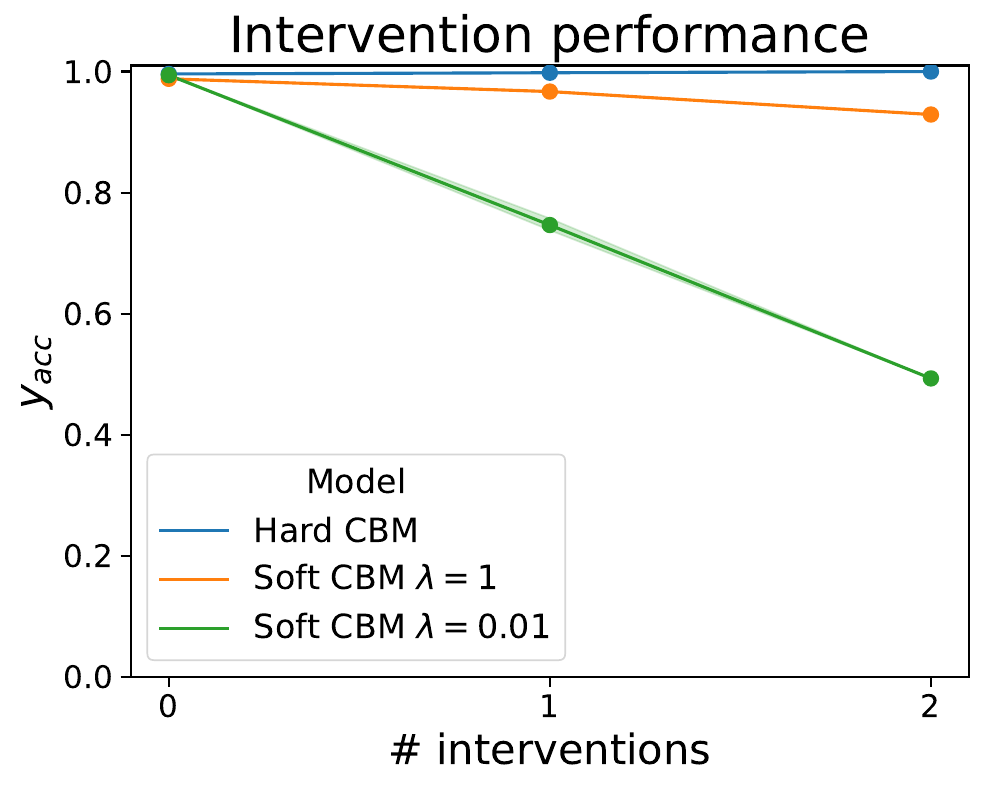} 
\caption{Performance upon random intervention of the three models analysed in Figure \ref{figure_TT_leakage_concept_distributions}. See Appendix \ref{App_experiments_Scores_evaluation} for more details.}
\label{figure_TT_intervention_2conceptsmodels_random}
\end{wrapfigure}

\paragraph{Performance upon intervention.} Interventions can be used to evaluate the interpretability of a model. They are performed  by substituting the ground-truth values of concepts to the corresponding predicted activation, with a given policy defining the order of the concepts to intervene on. The behaviour of the task accuracy after each intervention measures the extent to which the learnt and ground-truth concepts are aligned. A task accuracy $y_{acc}$ that decreases is a coarse indication of leakage since it suggests that the final head expects additional information that is not present in the ground-truth concept distribution.
Figure \ref{figure_TT_intervention_2conceptsmodels_random} shows the intervention performance of the same three models as in Figure \ref{figure_TT_leakage_concept_distributions}. As expected, models with higher leakage have worse intervention performance, while the hard CBM displays a monotonically increasing task accuracy.

As a metric of leakage, however, intervention performance has several limitations: 
\begin{enumerate}
    \item Because it is based on evaluation scores, it does not address the information-theoretical nature of leakage, resulting in a limited resolution. It is unable to distinguish between interconcept and concepts-task leakage, and it does not directly quantify interpretability at the level of individual concepts. It is therefore often insufficiently informative to guide model design or hyperparameter selection.
    \item Its evaluation is computationally expensive, especially in real-world applications in which models have a large number of concepts and are evaluated using state-of-the-art policies with a high overhead such as CooP \citep{CooP}.
    \item In more complex architectures such as CEMs, it is not a measure of leakage, and models with higher leakage may display better performance upon interventions (see Appendix \ref{App_further_CEM_noninterpret}).
\end{enumerate}

Collectively these issues indicate that poor intervention performance is a measure of leakage with perfect specificity (leakage has always occurred when intervention performance is poor), but with poor sensitivity (when intervention performance is good, leakage is not guaranteed to be absent).
Although the leakage scores we propose in Section \ref{sec_def_CTL_ICL} address these issues, intervention performance generally still remains a coarse but useful indicator of leakage in CBMs. 

Let us denote by $y_{acc}^{*(k)}$ the task accuracy after all the $k$ concepts have been intervened on in the model under evaluation, which coincides with the accuracy of the final head on the ground-truth concepts. As a benchmark of the proposed leakage scores we will use the intervention score
\begin{equation} \label{eq_def_S_int}
    \textrm{S}_{int}  = y_{acc}^{(k)} - y_{acc}^{*(k)}\,,
\end{equation}
where $y_{acc}^{(k)}$ is the reference accuracy of the final head when separately trained on the ground-truth concepts. It represents the maximal task accuracy attainable by the final head and it measures the fundamental misspecification of the final head with respect to the ground-truth functional dependence $y = f(c)$.
$\textrm{S}_{int} $ thus quantifies the further decrease in task accuracy caused by leakage accounting for final head misspecification. Its value is independent of the intervention policy adopted. Note that $y_{acc}^{(k)}$ coincides with $y_{acc}^{*(k)}$ for a successfully trained hard model, thus $\textrm{S}_{int}  = 0$  for hard models by construction.

\section{Concepts-task and interconcept leakage scores}
\label{sec_def_CTL_ICL}

Information leakage is information-theoretic in nature and, as such, it is most appropriately captured by metrics based on quantities from information theory. 
As discussed in Section \ref{sec_leakage}, the leaked information and the corresponding additional structure in the learnt concept distributions can be measured as modifications in the ground-truth interconcept and concepts-task MIs respectively.
This motivates the definition of the following set of scores based on information theory to quantify leakage and the interpretability of concept-based models:

\paragraph{Concepts-task leakage scores.} Denoting by $H(z)$ the entropy of the random variable $z$ and by $I(z, w)$ the MI between the random variables $z$ and $w$, we define the concepts-task leakage score (CTL) for concept $i$ as the difference between predicted and ground-truth MIs between concept $i$ and the task label,
\begin{equation} \label{eq_CTL_i}
\mathrm{CTL}_i(\hat{c}_i, c_i, y) = \max\left( 0, \frac{I(\hat{c}_i, y)}{H(y)} - 
\frac{I(c_i, y)}{H(y)}
\right) .
\end{equation}
This score quantifies the additional information that the learnt concept $i$ encodes about the task label with respect to the ground-truth.
MIs are appropriately normalised by the entropy of the class label to ensure $ 0 \leq \mathrm{CTL}_i \leq 1$.  When the difference inside the brackets is negative, the learnt concepts are less predictive of the task than expected from the ground truth, indicating poor concept learning rather than leakage. We further define the average concepts-task leakage score as
\begin{equation} \label{eq_CTL}
\mathrm{CTL}(\hat{c}, c, y) = \frac{1}{k} \sum_{i = 1 }^k \mathrm{CTL}_i(\hat{c}_i, c_i, y)\,,
\end{equation}
as the average extra information about the task stored into the learnt concepts. Note that a weighted average over concepts can be adopted in use cases where concept importances are non-uniform.

\paragraph{Interconcept leakage scores.} In a similar fashion, we define the pairwise interconcept leakage score  (ICL) between concepts $i$ and $j$ as
\begin{equation} \label{eq_ICL_ij}
\mathrm{ICL}_{ij}(\hat{c}, c) =  \max\left( 0, \frac{I(\hat{c}_i, \hat{c}_j)}{\sqrt{H(\hat{c}_i) \, H(\hat{c}_j)}}
-
\frac{I(c_i, c_j)}{\sqrt{H(c_i) \, H(c_j)}}
\right) ,
\end{equation}
which measures the additional predictivity of the learnt concept $i$ for concept $j$ with respect to ground-truth. Again, normalisation by the entropies ensures that $0 \leq \mathrm{ICL}_{ij} \leq 1$. Note that $\mathrm{ICL}_{ii} = 0$ since $I(z, z) = H(z)$. We also define the per-concept and average interconcept leakage scores as
\begin{equation} \label{eq_ICL_i_ICL}
    \mathrm{ICL}_i(\hat{c}, c) = 
\frac{1}{k-1} \sum_{\substack{j = 1}}^k \mathrm{ICL}_{ij}(\hat{c}, c)\,, \qquad \quad 
\mathrm{ICL}(\hat{c}, c) = 
\frac{1}{k} \sum_{\substack{i = 1}}^k \mathrm{ICL}_i(\hat{c}, c)\,,
\end{equation}
to quantify the extra information that each concept encodes about the other concepts, and the average additional interconcept predictivity respectively.

The CTL and ICL scores are meant to summarise complementary aspects of the leakage present in a given concept-based model, as well as its overall interpretability. Based on these measures, we define the following criterion to detect leakage:
\begin{leakage_criterion*}
\label{leakage_criterion}
When comparing two models (or two model classes) $A$ and $B$, a sufficient condition for $A$ having higher leakage than $B$ is that either of the following conditions are satisfied,
\begin{itemize}
    \item both the \(\mathrm{CTL}\) and \( \mathrm{ICL}\) scores are higher in $A$ than in $B$ with high statistical confidence,
    \item either the \(\mathrm{CTL}\) or the \( \mathrm{ICL}\) score is higher in $A$ with high statistical confidence, while the other score takes statistically compatible values in $A$ and $B$.
\end{itemize}
\end{leakage_criterion*}
Note that this is a conservative criterion covering the majority of cases, while it cannot be used in the cases where one measure is higher in A and the other is lower in B (or vice versa) with high statistical confidence.

These measures also convey more detailed information about each type of leakage at the level of single concepts. This is particularly useful for model design and when performing interventions (as they essentially indicate the risk of intervening on each concept). The framework above can be applied to concept-based models with a range of concept encodings, including binaries, soft probabilities, logits and vector representations. Alternative normalisations of MI can be used, although we found those adopted in \eqref{eq_CTL_i} and \eqref{eq_ICL_ij} to be particularly transparent from an information theory perspective, and robust in the experiments we carried out. We estimate MI and entropy using the Kraskov-Stögbauer-Grassberger (KSG) estimator \citep{KSG03}, based on $k$-nearest neighbour statistics.

\section{Robustness of leakage scores}
\label{sec_robustness}

\begin{figure}[t]
\begin{minipage}[c]{0.32\textwidth} 
\centering \small \sffamily
$\quad \;$ TabularToy(0.25)
\vspace{0.2cm}
\end{minipage}
\begin{minipage}[c]{0.32\textwidth} 
\centering \small \sffamily
$\quad$ dSprites(0)
\vspace{0.2cm}
\end{minipage}
\begin{minipage}[c]{0.32\textwidth}
\centering \small \sffamily
$\quad \;$ 3dshapes(0)
\vspace{0.2cm}
\end{minipage}

\begin{minipage}[c]{0.32\textwidth} 
\centering
\includegraphics[width=1\textwidth]{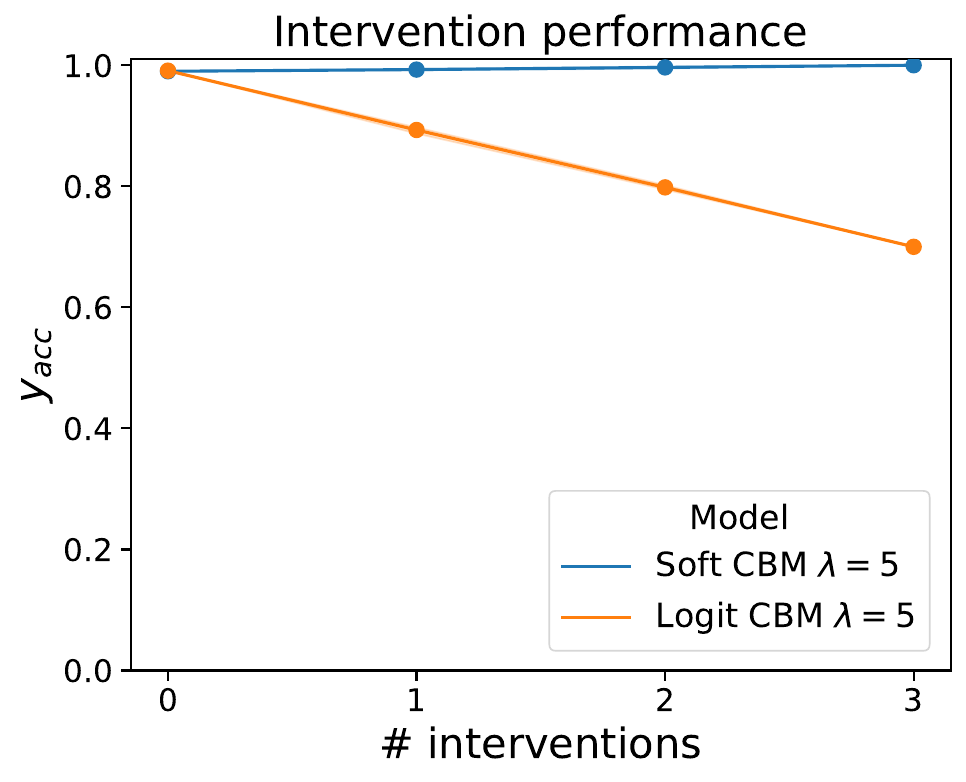} 
\end{minipage}
\begin{minipage}[c]{0.32\textwidth} 
\centering
    \includegraphics[width=1\textwidth]{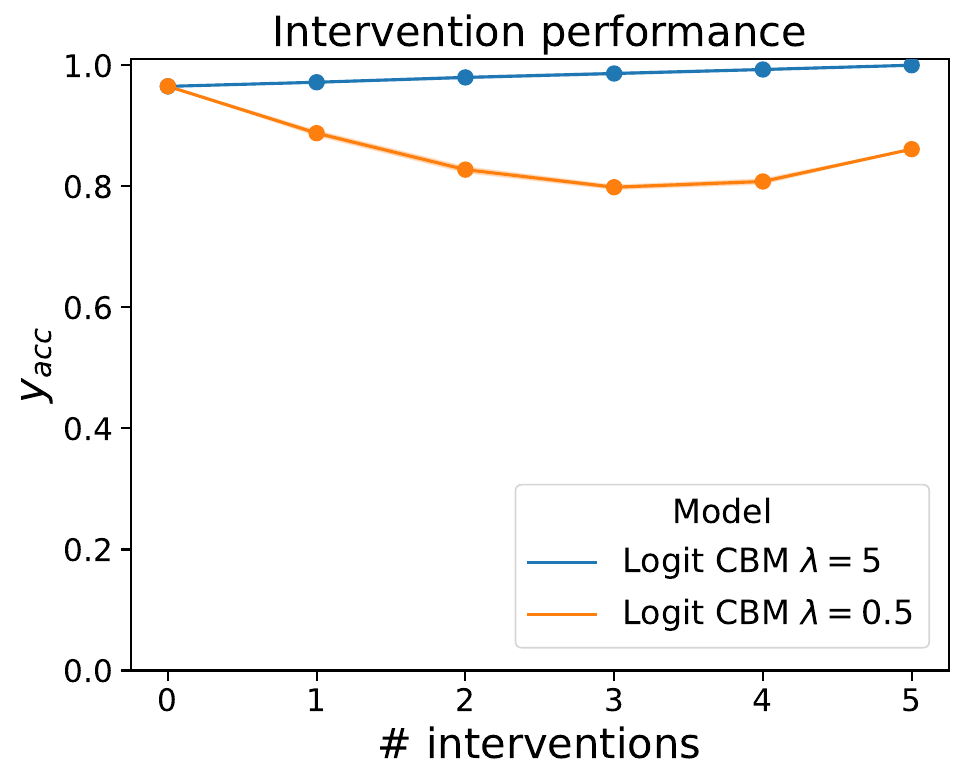} 
\end{minipage}
\begin{minipage}[c]{0.32\textwidth} 
\centering
    \includegraphics[width=1\textwidth]{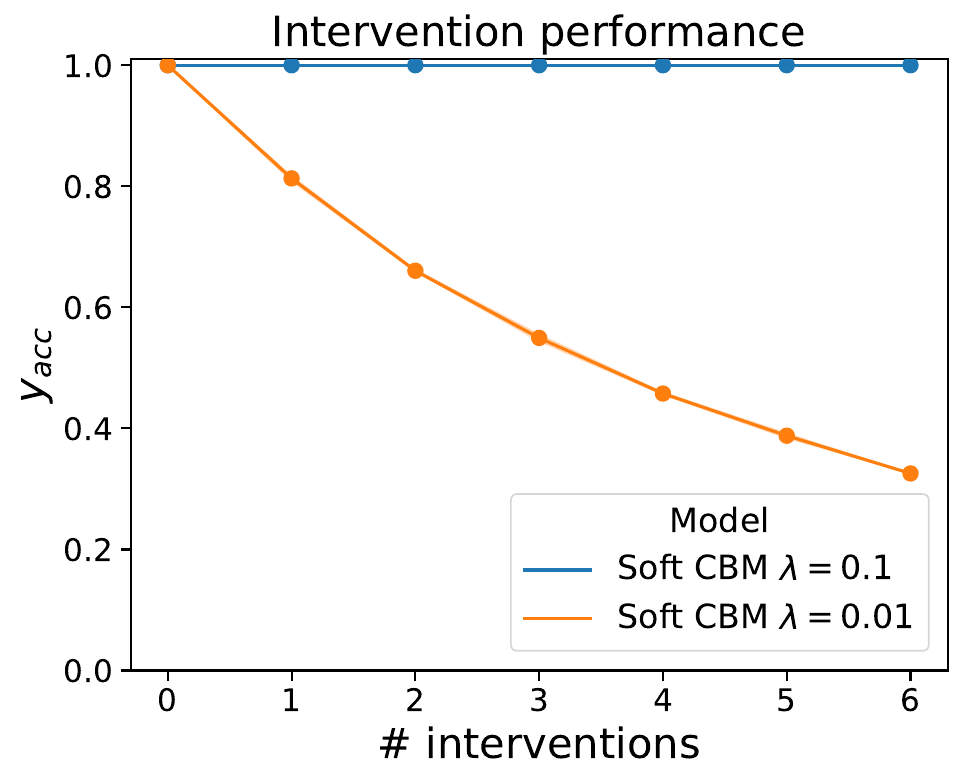} 
\end{minipage}

\begin{minipage}[c]{0.32\textwidth} 
\centering
\includegraphics[width=0.8\textwidth]{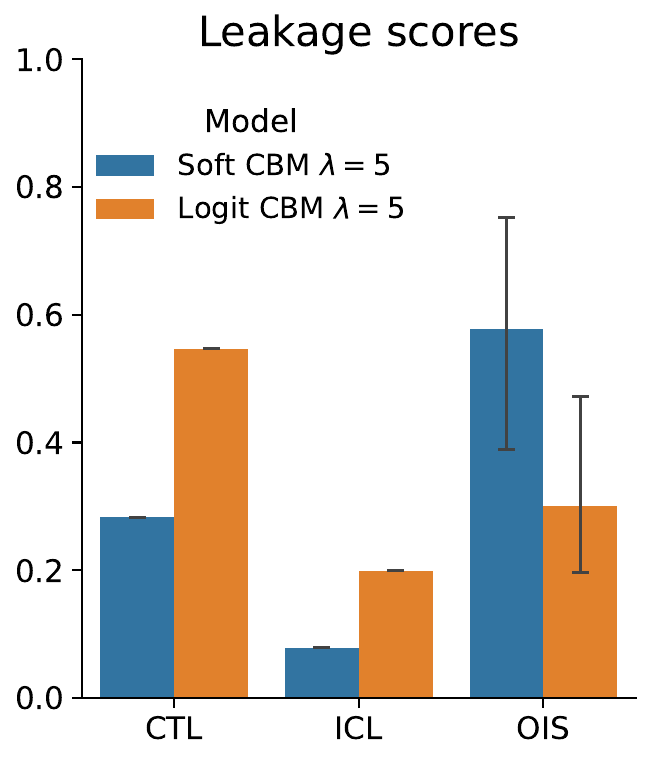} 
\end{minipage}
\begin{minipage}[c]{0.32\textwidth}
\centering
    \includegraphics[width=0.8\textwidth]{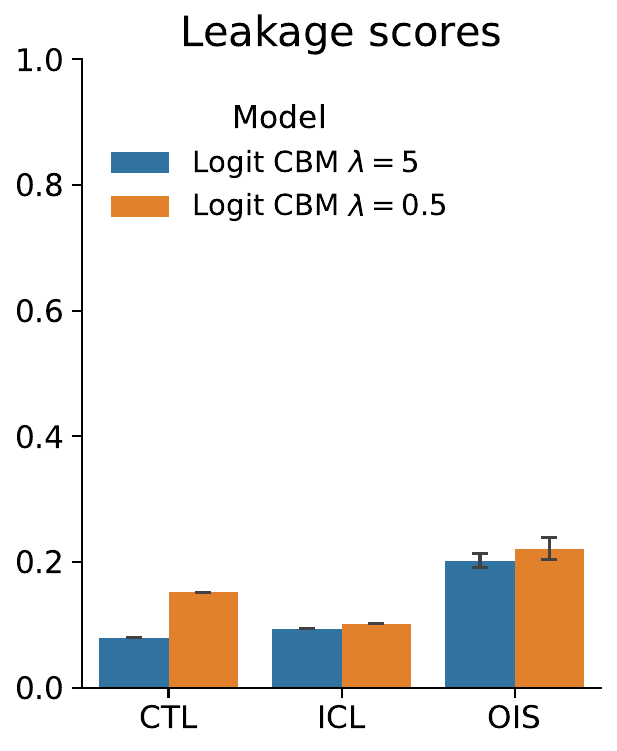} 
\end{minipage}
\begin{minipage}[c]{0.32\textwidth} 
\centering
    \includegraphics[width=0.8\textwidth]{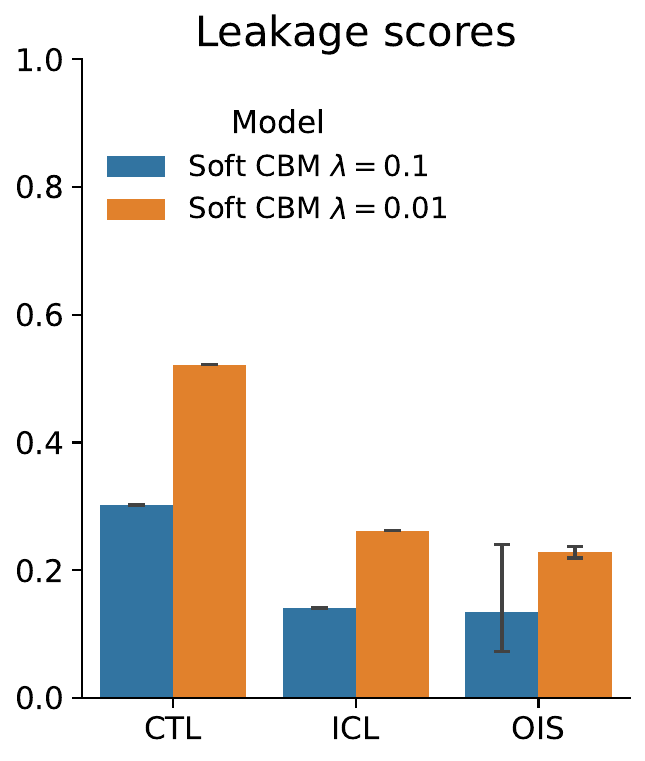} 
\end{minipage}
    \caption{Intervention performance and leakage scores computed for pairs of soft and logit CBMs with different amounts of leakage but similar evaluation scores (see Table \ref{table_evaluation_scores_single_models}) trained on TabularToy(0.25), dSprites(0) and 3dshapes(0). Metrics are evaluated on a 5-fold basis for each model.  
    }
    \label{figure_leakage_scores_single_models}

\begin{minipage}[c]{\textwidth} \small
\vspace{0.6cm}
\centering
    \begin{tabular}{ll|ccc|ccc|c}
          &  & $c_{acc}$ &  $c_{\text{F1}}$ &  $c_{\text{AUC}}$ &  $y_{acc}$ &  $y_{\text{F1}}$ & $y_{\text{AUC}}$ & $\textrm{S}_{int}  (\downarrow)$\\
         \midrule
         \multirow{2}{2.5cm}{\centering TabularToy($0.25$)} & Soft CBM $\lambda = 5$  &  0.993 &  0.992 &  0.992 &  0.990 & 0.990 & 0.990 & 0.000 \\
         & Logit CBM $\lambda = 5$ &  0.995 &  0.995 & 0.995 &  0.991 & 0.991 & 0.991 & 0.301
         \vspace{0.2cm}\\
         \multirow{2}{2.cm}{\centering dSprites($0$)} & Logit CBM $\lambda = 5$  &  0.993 &  0.993 &  0.993 &  0.965 & 0.976 & 0.698 & 0.000  \\
         & Logit CBM $\lambda = 0.5$ &  0.992 &  0.992 & 0.992 &  0.965 & 0.975 & 0.689 & 0.139 
         \vspace{0.2cm}\\
         \multirow{2}{2.cm}{\centering 3dshapes($0$)} &  Soft CBM $\lambda = 0.1$  &  1.000 &  1.000 &  1.000 &  1.000 & 1.000 & 0.201 & 0.000  \\
         & Soft CBM $\lambda = 0.01$ &  1.000 &  1.000 &  1.000 &  1.000 & 1.000 & 0.207 & 0.674 \\
 
    \end{tabular}
    \captionof{table}{Concept and task evaluation scores, and intervention scores for pairs of soft and logit CBMs with different levels of leakage (see Figure \ref{figure_leakage_scores_single_models}).}
    \label{table_evaluation_scores_single_models}
\end{minipage}
\end{figure}

We now demonstrate the robustness of the CTL and ICL scores and how they overcome the issues of the existing measures of leakage. 

\paragraph{Datasets.} For our experiments we considered three synthetic datasets, TabularToy($\delta$),  dSprites($\gamma$) and 3dshapes($\gamma$) (presented in \cite{Espinosa_Zarlenga_2023}), as well as two real-world datasets, the Caltech-UCSD Birds-200-2011 dataset \citep[CUB,][]{CUB} and HAM10K \citep{Tschandl2018}. TabularToy($\delta$) is a binary-class tabular dataset based on the dataset from \cite{Mahinpei2021PromisesAP}, while dSprites($\gamma$) and 3dshapes($\gamma$) are multi-class image-based datasets building upon dSprites \citep{dsprites17} and 3dshapes \citep{3dshapes18} respectively. The CUB dataset contains bird images annotated with 112 binary attributes capturing phenotypic traits, and the task is to classify each image into one of 200 species. HAM10K consists of dermatoscopic lesion images labelled with 18 morphological concepts, with the task of classifying each lesion as benign or malignant.
The use of synthetic and real-world datasets enables us to (i) modify the functional dependence $y=f(c)$ and hence tune the ground-truth concepts-task MI, and (ii) tune the ground-truth interconcept MI via the parameters $\delta \in (0,1)$ and $\gamma$ ($\in \{0, \dots, 4\}$ in dSprites and $\in \{0, \dots, 5\}$ in 3dshapes), an essential aspect to assess interpretability \citep{ConceptCorrelation}. See Appendix \ref{App_experiments} for more details of these experiments.

\paragraph{CTL and ICL are highly sensitive to leakage.} We consider pairs of models encoding different levels of leakage as assessed by intervention performance (Figure \ref{figure_leakage_scores_single_models}). The models in each pair were chosen to have essentially identical concepts and task evaluation scores (Table \ref{table_evaluation_scores_single_models}), evidencing the limitations of these metrics in capturing leakage.
In all cases the CTL and ICL scores correctly detect the different amounts of leakage according to the Leakage Criterion, with minimal uncertainty. In contrast, the OIS has large confidence intervals and so does not allow statistically significant conclusion on leakage to be drawn. Appendix \ref{App_concept_wise_leakage_scores} presents the concept-wise leakage scores for these models and illustrates how they provide more fine-grained information at the level of individual concepts.

\begin{figure}[t]
    \centering
    \includegraphics[width=0.6\textwidth]{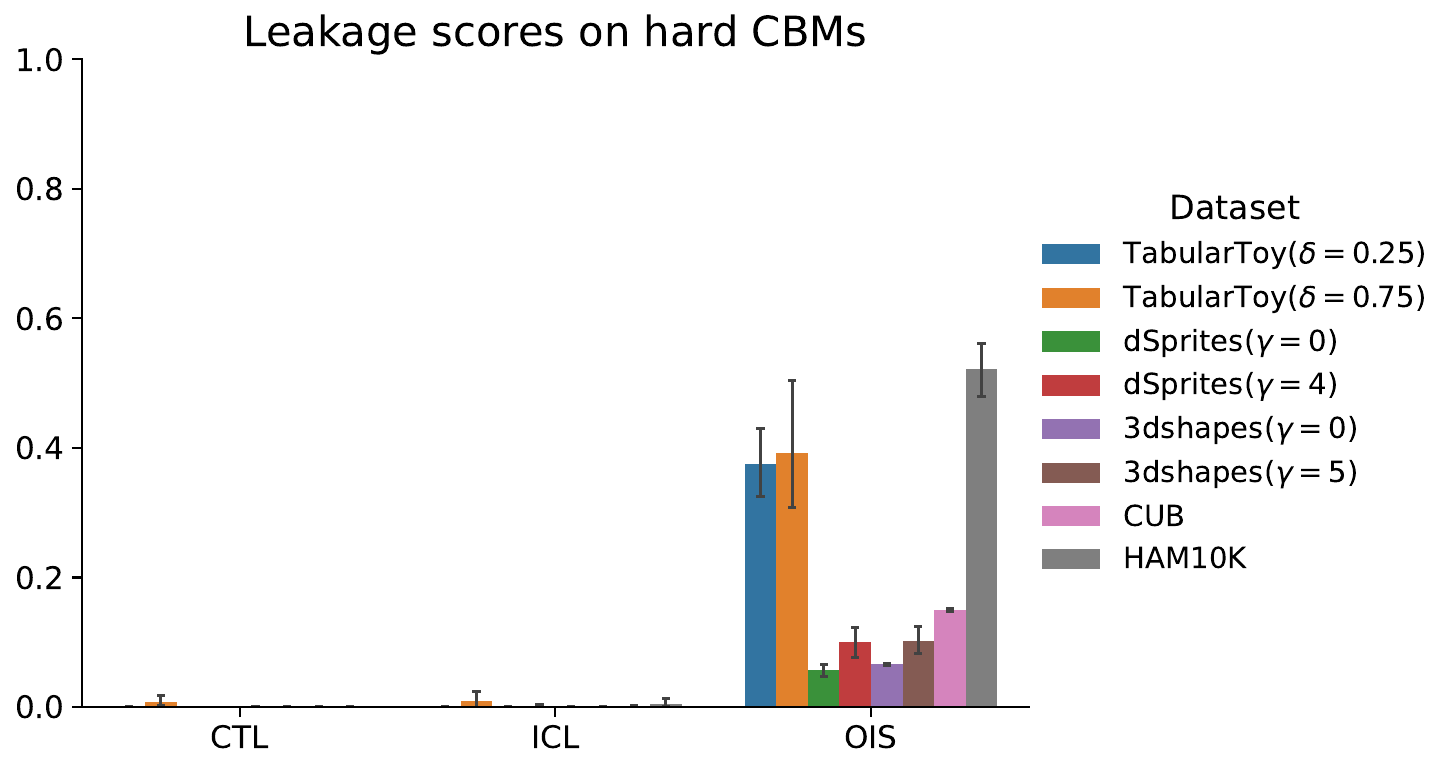}
    \caption{Leakage scores evaluated on hard CBMs on datasets with high and low ground-truth interconcept MI. The 95\% confidence intervals are obtained from 5-fold training on each dataset. }
    \label{figure_leakage_scores_on_hard_CBMs}
\end{figure}

\paragraph{CTL and ICL are vanishing for hard CBMs.} Figure \ref{figure_leakage_scores_on_hard_CBMs} shows the leakage scores for hard models trained on a number of datasets in both low and high regimes of ground-truth interconcept MI. By construction hard CBMs have vanishing leakage \citep{koh20a, Kazhdan2021IsDA, margeloiu2021}. In all cases, the CTL and ICL scores do not deviate significantly from zero, regardless of the interconcept MI. In contrast, the OIS is non-vanishing in all cases, indicating that it does not meet a fundamental requirement as a measure of leakage.
We further note that for each dataset the estimated OIS for hard CBMs are comparable to those of the models shown in Figure \ref{figure_leakage_scores_single_models}, which exhibit a significant amount of leakage. Thus, in these cases OIS does not distinguish interpretable from uninterpretable concept-based models.

\begin{table}[t] \small
    \centering
    \begin{tabular}{c|cc|cc|cc}
 & \multicolumn{2}{c|}{CTL}  &  \multicolumn{2}{c|}{ICL} &  \multicolumn{2}{c}{OIS} \\
 &  Pearson $r$ & $p$-value &   Pearson $r$ & $p$-value  &   Pearson $r$ & $p$-value\\
 \midrule
TT($0.25$) &  \textbf{0.75} & $\bm{3.4 \times 10^{-3}}$
& 0.69  &  $1.4 \times 10^{-2}$
& 0.49 & $1.2 \times 10^{-1}$ \\ 
dS($0$)      & \textbf{0.83} & $\bm{1.1 \times 10^{-3}}$ 
&  0.68 & $1.5 \times 10^{-2}$  
&  0.70 & $1.1 \times 10^{-2}$   \\
3ds($0$)      &  0.84 & $4.2 \times 10^{-5}$  
& \textbf{0.86} & $\bm{2.4 \times 10^{-5}}$ 
&  0.82 & $\!\,\,\,2.0 \times 10^{-4}$  \vspace{0.2cm}\\

TT($0.75$) & \textbf{0.54} & $\bm{4.9 \times 10^{-2}}$ 
& 0.51 & \textbf{$1.1 \times 10^{-1}$} 
& 0.27  & $4.1 \times 10^{-1}$   \\
dS($4$)      & \textbf{0.67} &  $\bm{1.9 \times 10^{-2}}$ 
& 0.48 & $1.2 \times 10^{-1}$
& 0.55 &  $7.4 \times 10^{-2}$ \\
3ds($5$)      & \textbf{0.79} & $\bm{3.9 \times 10^{-4}}$ 
& 0.69 & $3.5 \times 10^{-2}$ 
& 0.70 & $6.5 \times 10^{-3}$ \vspace{0.2cm}\\

CUB & \textbf{0.72} & $\bm{9.0 \times 10^{-3}}$ 
& 0.67 & \textbf{$2.1 \times 10^{-2}$} 
& 0.44  & $1.1 \times 10^{-1}$   \\
HAM10K & \textbf{0.49} & $\bm{1.8 \times 10^{-2}}$ 
& 0.41 & \textbf{$5.4 \times 10^{-2}$} 
& $-0.06$  & $7.7 \times 10^{-1}$   \\
    \end{tabular}
    \caption{Pearson $r$ coefficients and corresponding $p$-values measuring the correlations of CTL, ICL and OIS metrics against $\textrm{S}_{int} $.}
    \label{table_correlation_leakage_scores_intervention}
\end{table}

\paragraph{CTL and ICL strongly correlate with intervention performance.} We trained a number of soft and logit CBMs on each dataset in both high and low regimes of ground-truth interconcept MI and with different levels of concept supervision (see Appendix \ref{App_correlation_leakage_scores_int} for details on this experiment). The evaluation of the OIS, ICL and CTL scores was repeated 5 times for each individual model to provide estimates of uncertainty. 
We then took 10,000 Monte-Carlo (MC) samples from the resulting distributions of each score (assuming normality), and for each sample we computed the Pearson $r$ coefficient between each score and the intervention score $\textrm{S}_{int}$. 
Pooling using Rubin's rule \citep[see][]{rubin2004multiple, BarnardRubin99}, we obtain a single best estimate for the Pearson $r$ and its associated $p$-value for each leakage score and dataset, which are reported in Table \ref{table_correlation_leakage_scores_intervention}.

The proposed CTL and ICL scores strongly correlate with intervention performance, they systematically outperform OIS across all datasets and correspond robustly to ground-truth interconcept MI. This analysis also quantifies the importance of concepts-task and interconcept leakage in each case. CTL generally appears as the prominent form of leakage over ICL. Its correlation with intervention performance is typically stronger than ICL, and is, moreover, significant in all datasets. This behaviour suggests that optimization favours the storage of additional task-relevant information in the learnt concepts rather than information about the other concepts. Nevertheless, these results also provide evidence that interconcept leakage systematically arises during model training; indeed, correlations of comparable magnitude to concept-task leakage were observed in the experiments on 4 out of 8 datasets.

\section{Causes of leakage}
\label{sec_causes_of_leakage}

The previous results indicate that our proposed information-theoretic leakage scores are robust and sensitive and can therefore be used to identify the most common causes of leakage and assess their impact on interpretability. 

A first effect that directly impacts the interpretability of a concept-based model is the concept encoder being misspecified with respect to the target concept distribution. When such a misspecification is significant, the model is essentially not capable of learning concepts and their relations. This may result in additional issues, such as non-locality \citep{raman2023do,Huang24}. To address this issue, concept-based architectures that enforce accurate learning of interconcept relations have been recently proposed \citep{ProbabilisticCBMs, vandenhirtz2024stochastic, xu2024energybased, Kim25}. We recognise however that this problem is strongly use-case specific and it is usually sufficient to opt for a concept encoder with an improved architecture to solve it. In the following experiments we will thus consider concept encoders that are sufficiently well-specified, as relevant for real-world high-risk scenarios where the aim is both high concept performance and the absence of leakage.

\begin{figure}[t]
\begin{minipage}[c]{0.325\textwidth} 
\centering
\includegraphics[width=\textwidth]{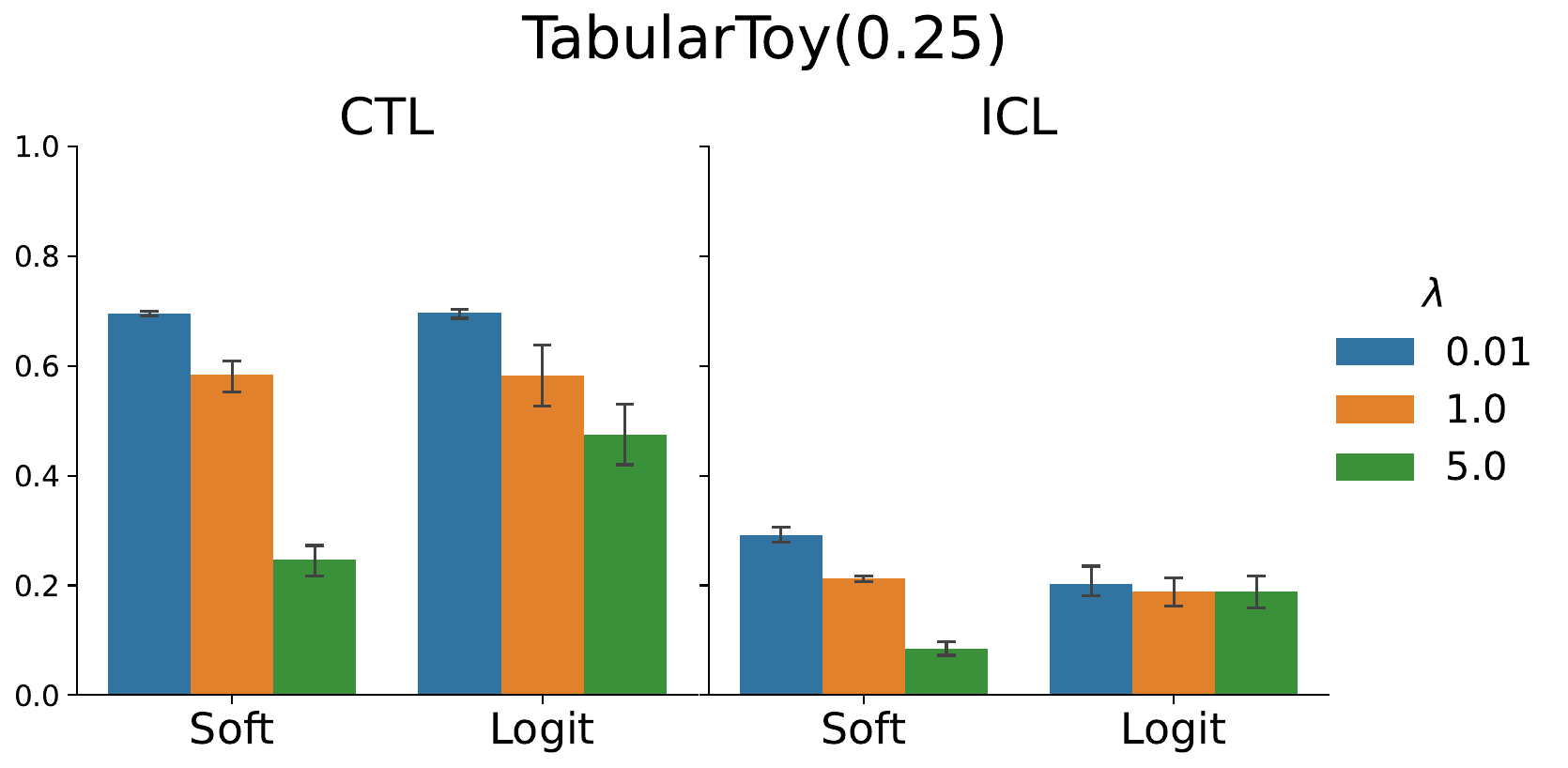} 
\end{minipage}
\begin{minipage}[c]{0.325\textwidth} 
\centering
\includegraphics[width=\textwidth]{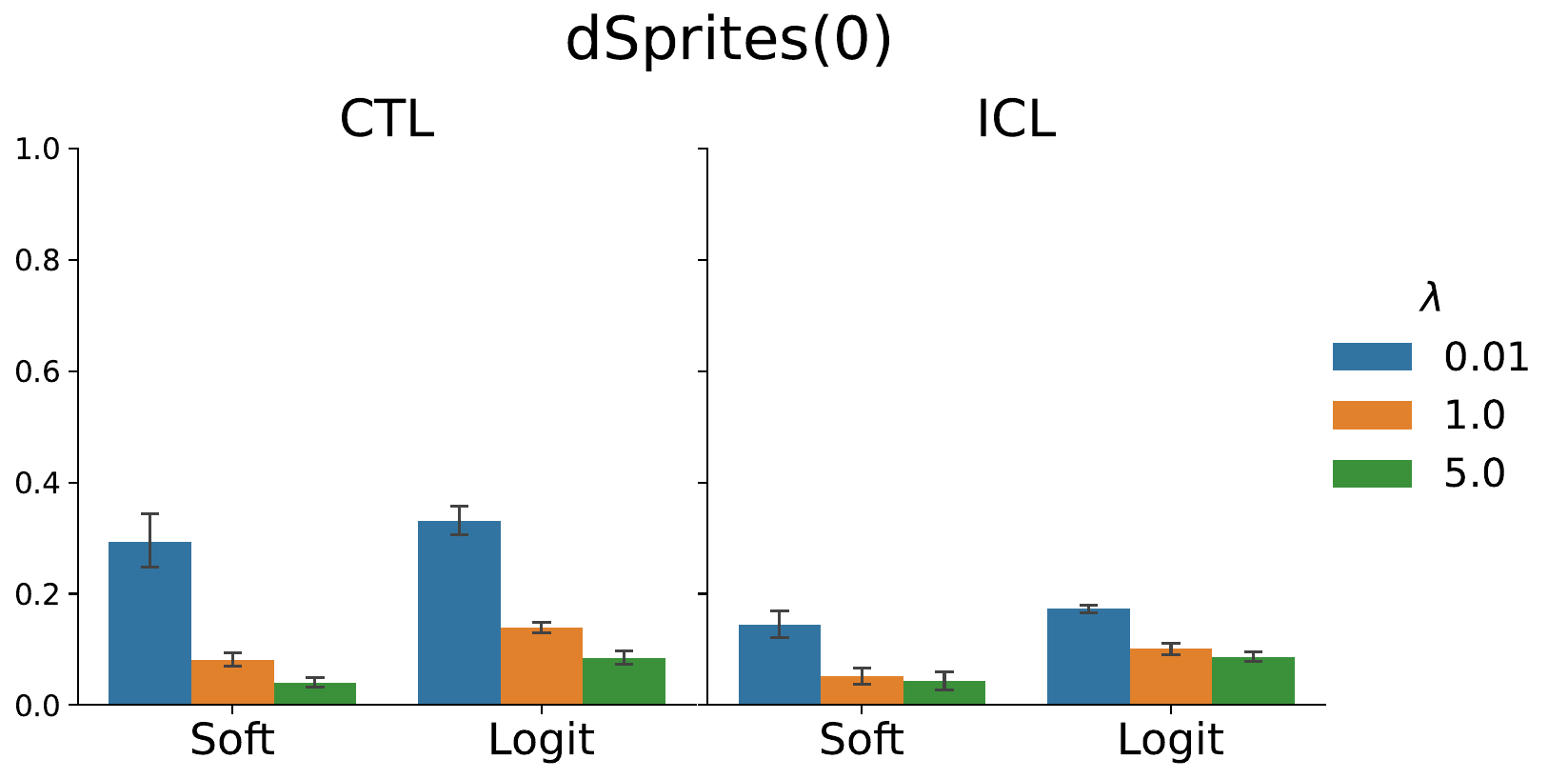} 
\end{minipage}
\begin{minipage}[c]{0.325\textwidth} 
\centering
\includegraphics[width=\textwidth]{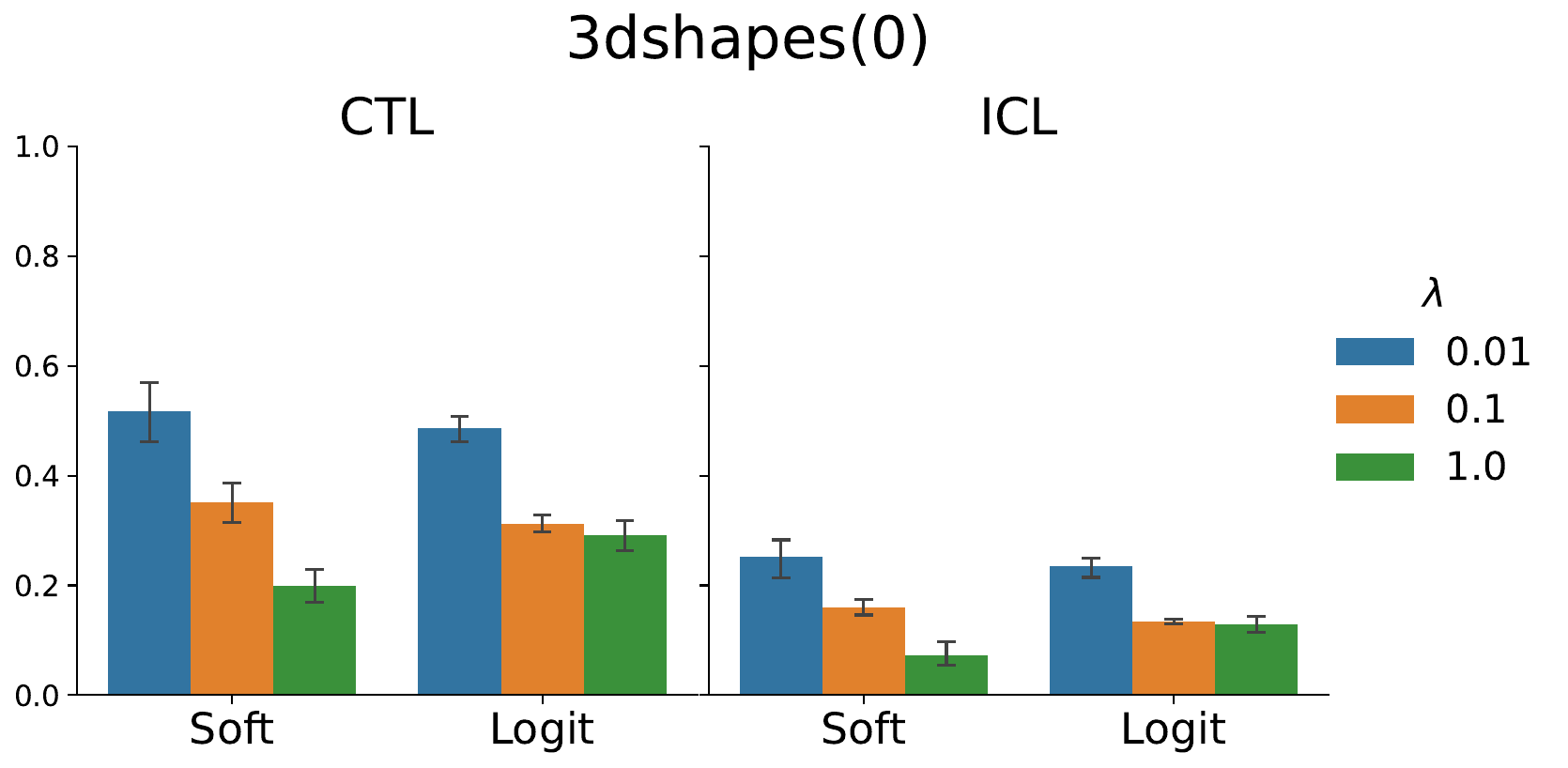} 
\end{minipage}

\begin{center}
    \begin{minipage}[c]{0.325\textwidth} 
\centering
\includegraphics[width=\textwidth]{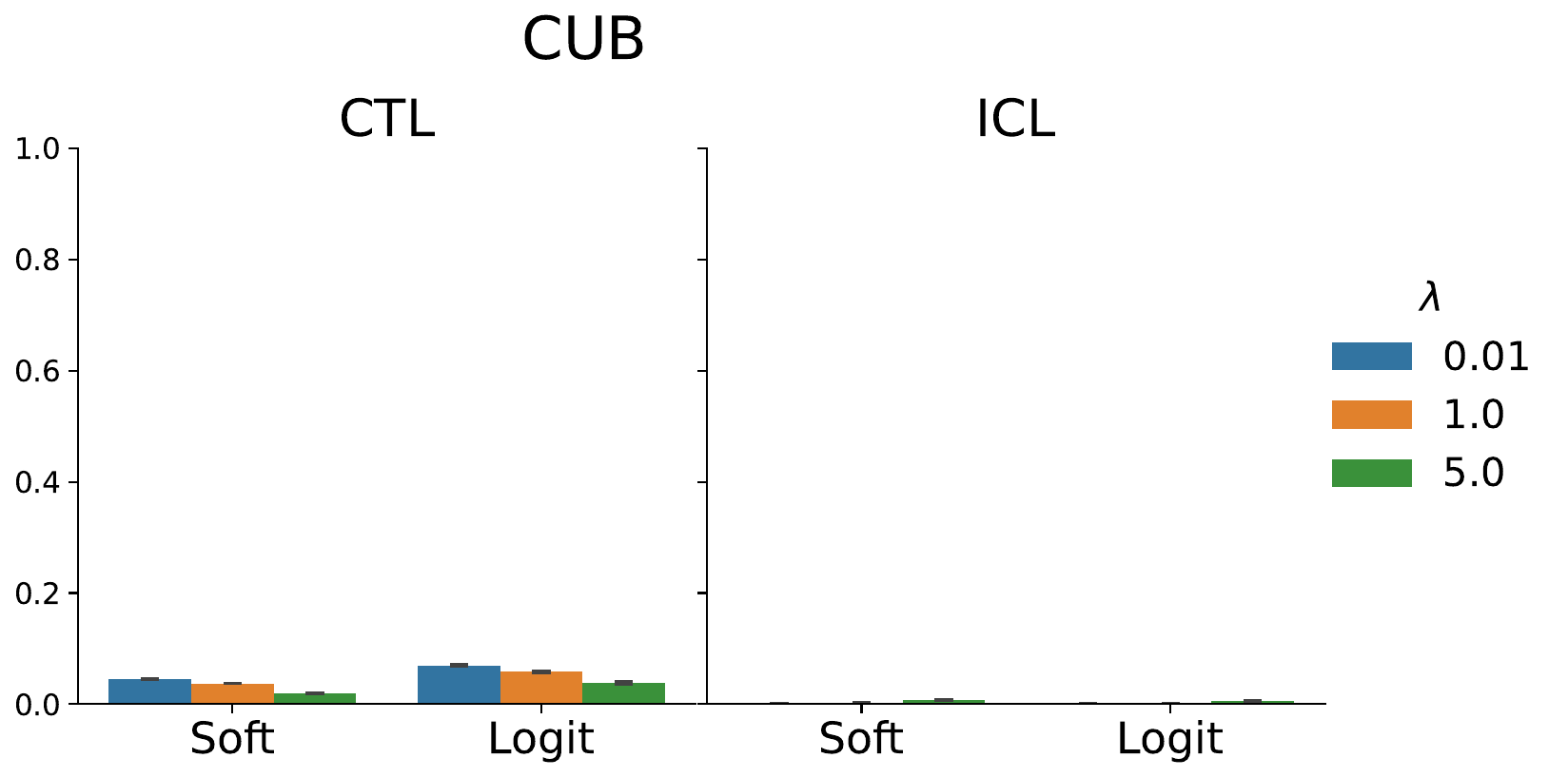} 
\end{minipage}
\begin{minipage}[c]{0.325\textwidth} 
\centering
\includegraphics[width=\textwidth]{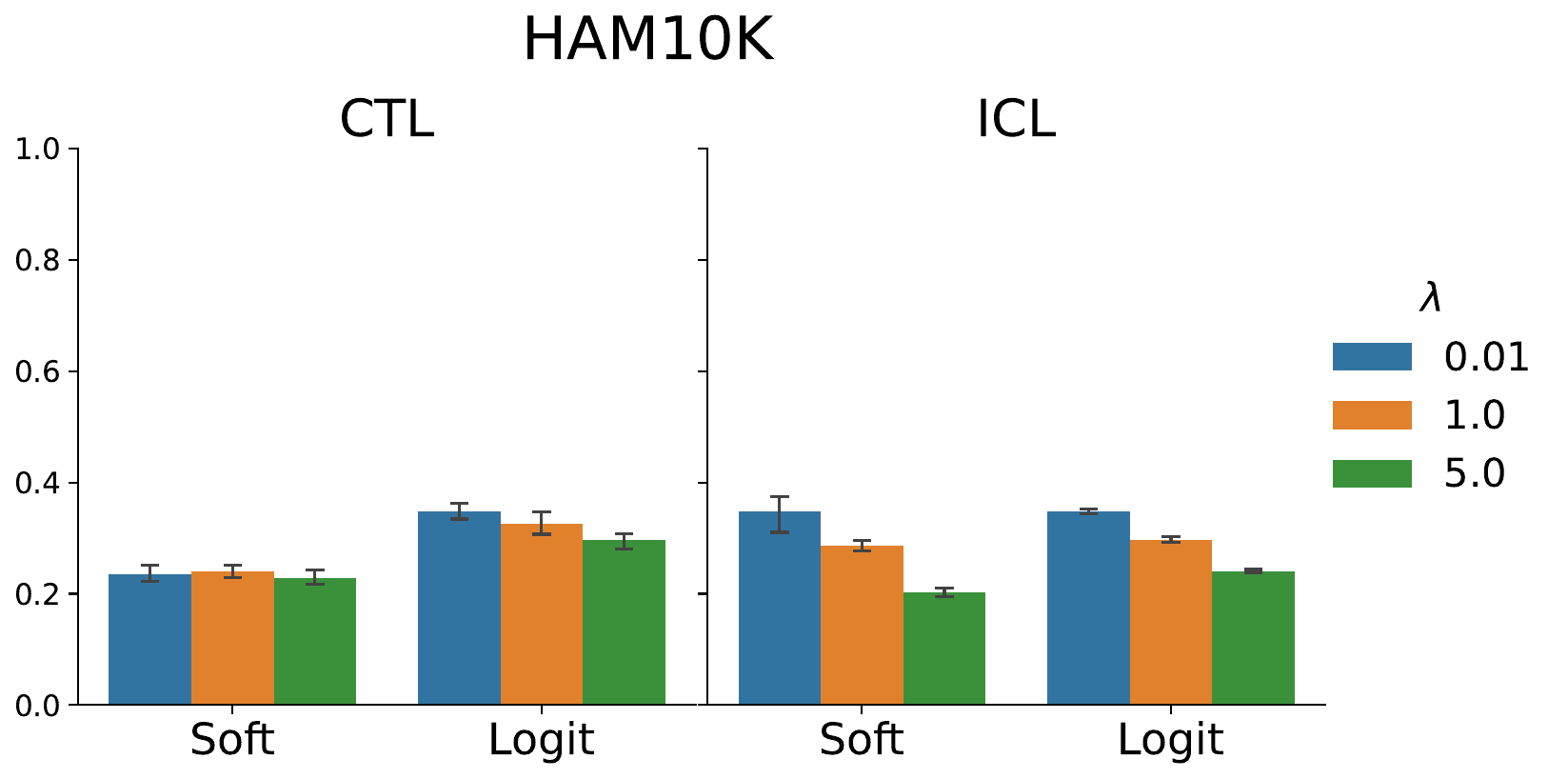} 
\end{minipage}
\end{center}
\caption{
Leakage scores evaluated for soft and logit models with different levels of concept supervision across datasets.
    }
    \label{figure_causes_soft_vs_logit}
\end{figure}

\paragraph{Insufficient concept supervision.} When $\lambda$ is set to be small in the loss \eqref{eq_loss_joint_CBMs} during training, the model will achieve better task performance at the cost of poor concept learning and high leakage. This issue was originally observed in \cite{koh20a}, where poor intervention performance was found when training models with low $\lambda$.
To precisely quantify this effect, we trained soft and logit CBMs with low, intermediate and high values of $\lambda$ on datasets with both low and high interconcept MI. The resulting leakage scores, displayed in Figures \ref{figure_causes_soft_vs_logit} and \ref{figure_causes_soft_vs_logit_highCorr}, confirm that leakage decreases as concept supervision is increased. 

However, as detailed in Appendix \ref{App_high_lambda}, leakage does not fall to zero as $\lambda$ is increased -- instead, the leakage scores reach a non-vanishing minimal value, specific to each model class. 
At higher values (here, $\lambda \gtrsim 10$), models focus excessively on concept learning, and the task objective may be poorly learnt. This results in a drop in task performance or an increase in leakage. Each model class thus has an associated optimal range of $\lambda \in (\lambda_{min}, \lambda_{max})$ where the minimal amount of leakage is attained, below which models exhibit proportionally higher leakage and above which training often fails. As evidenced by our results, such an optimal interval can be identified using the CTL and ICL scores.

\paragraph{Over-expressive concept encoding.} When the chosen concept representation is significantly more expressive than the annotated concept representation, the model will generally misuse the surplus expressivity to store additional information in the concepts to aid task prediction. This phenomenon is illustrated in Figures \ref{figure_causes_soft_vs_logit} and \ref{figure_causes_soft_vs_logit_highCorr} where annotated concepts are binary, while learnt concepts are soft probabilities and logits, and as such over-expressive for the setup. We note that raising $\lambda$ is less effective at decreasing leakage when concepts representations are more expressive, as in the logit models, and furthermore the minimal attainable leakage is significantly higher for such models over soft CBMs. CEMs vector representations with binary ground-truth concepts exhibit a similar behaviour, as we discuss in Section \ref{sec_interpretability_CEMs}. Finally, we observe that models with low $\lambda$ generally have a comparable level of high leakage regardless of the concept encoding.

\paragraph{Incomplete set of concepts.} When the set of annotated concepts is incomplete for the task -- meaning that it is not sufficiently predictive -- models tend to encode information about the missing concepts (or, more generally, any additional information that may be required) into the learnt concepts to nonetheless achieve a high task performance. This typically translates into sizeable leakage and a loss of interpretability. This issue has been discussed in previous work \citep{Kazhdan2021IsDA, margeloiu2021, Mahinpei2021PromisesAP, Addressing_leakage, GlanceNet}.

\begin{figure}[t]
\begin{minipage}[c]{0.325\textwidth} 
\centering
\includegraphics[width=0.9\textwidth]{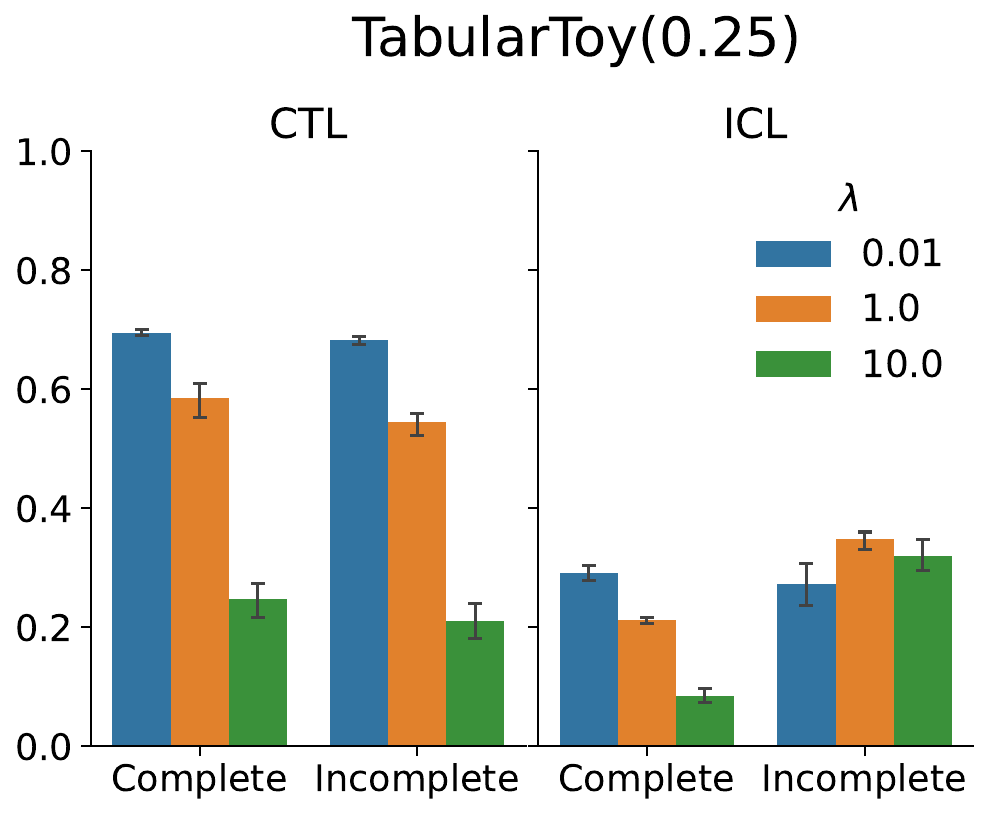} 
\end{minipage}
\begin{minipage}[c]{0.325\textwidth} 
\centering
\includegraphics[width=0.9\textwidth]{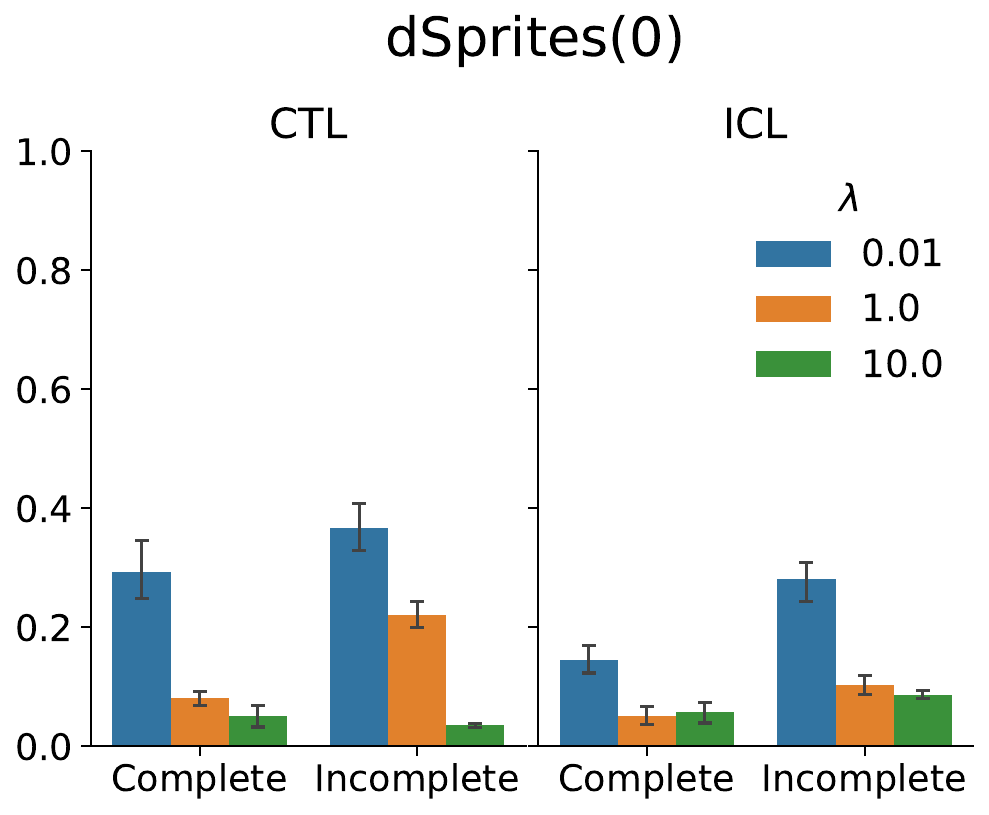} 
\end{minipage}
\begin{minipage}[c]{0.325\textwidth} 
\centering
\includegraphics[width=0.9\textwidth]{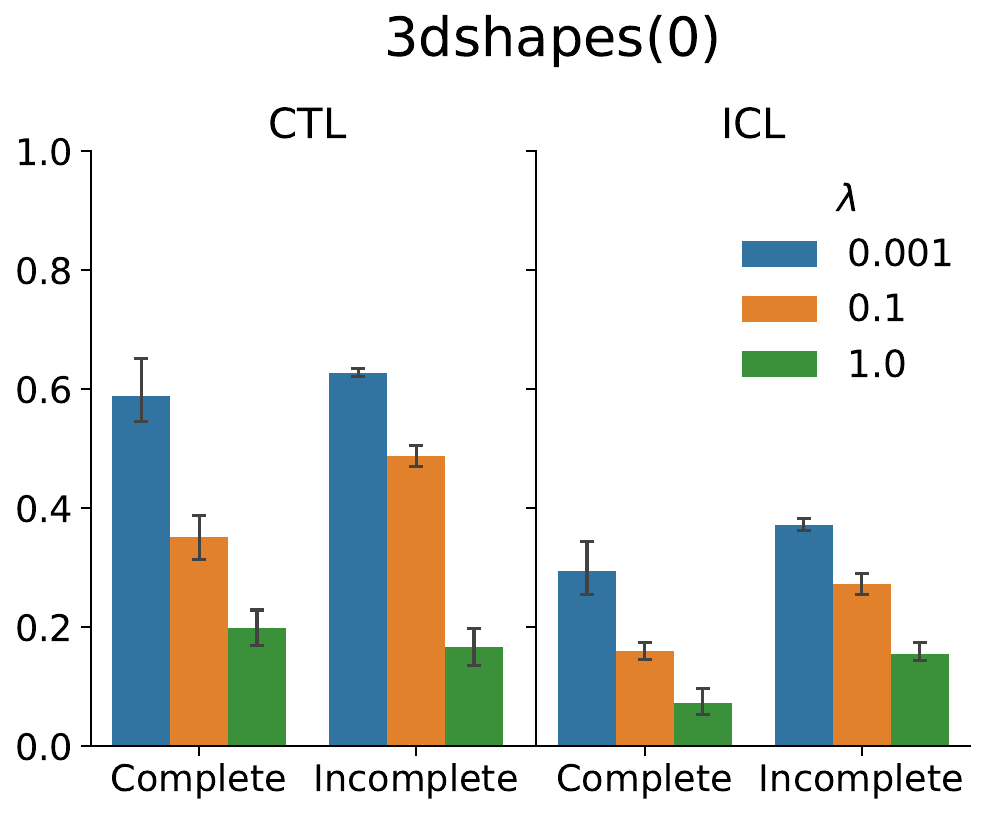} 
\end{minipage}

\begin{center}
    \begin{minipage}[c]{0.325\textwidth} 
\centering
\includegraphics[width=0.9\textwidth]{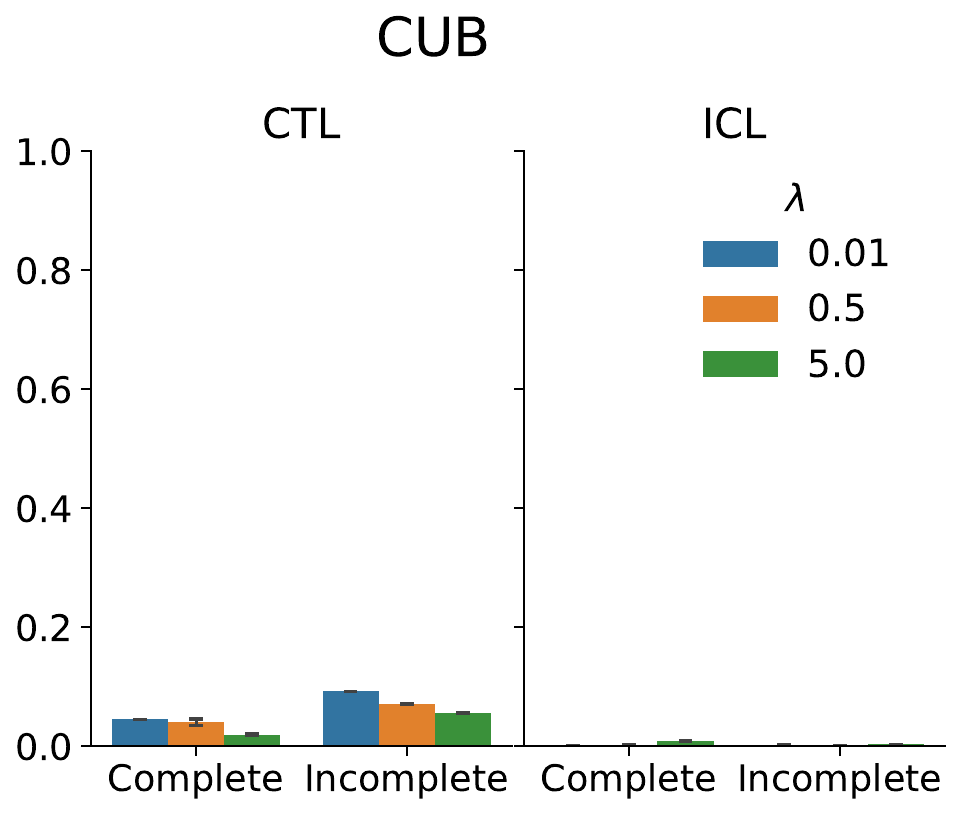} 
\end{minipage}
\begin{minipage}[c]{0.325\textwidth} 
\centering
\includegraphics[width=0.93\textwidth]{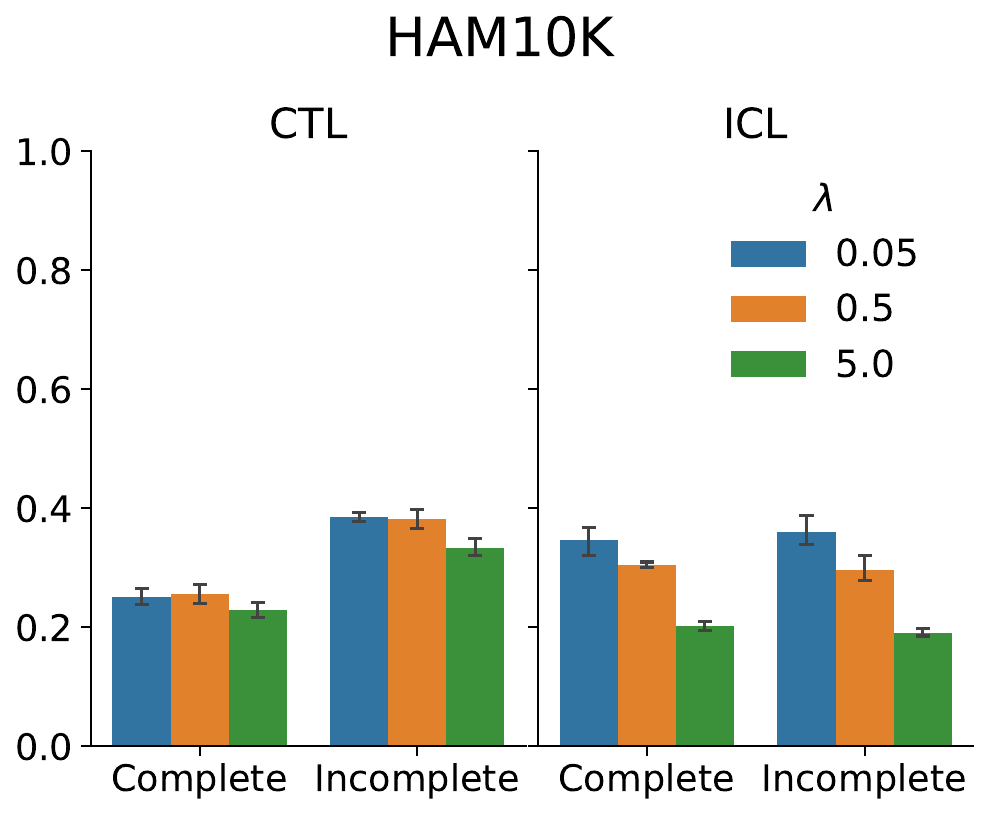} 
\end{minipage}
\end{center}
\caption{
Leakage scores evaluated for soft CBMs at different levels of concept supervision, on datasets with complete and incomplete sets of concepts.
    }
    \label{figure_causes_incomplete_concept_set}
\end{figure}

To accurately assess this phenomenon using our information-theoretic scores, we trained soft CBMs with low, intermediate and high levels of concept supervision on a range of datasets with a significant fraction of concepts removed (see Appendix \ref{App_experiments_Datasets} for more details on which concepts are removed for each dataset). 
This resulted in a sizeable decrease of the reference task accuracy $y_{acc}^{(k)}$ of the final head trained on the ground-truth concepts introduced in equation \eqref{eq_def_S_int} (Table \ref{table_c2y_task_performance_incomplete_concepts_misspecified}, left). The leakage scores of the trained models are displayed in Figure \ref{figure_causes_incomplete_concept_set}, along with those of models with complete sets of concepts for comparison. The relevant concepts and task accuracies are presented in Appendix \ref{App_accuracies_incomplete_concepts_misspecified_head}. 
 
According to the Leakage Criterion and as expected, we detected more leakage in 14/15 model classes trained on incomplete sets of concepts. At low and intermediate values of $\lambda$, there is significantly more leakage in models trained on an incomplete set.

\begin{table}[t] \small
    \centering
    \begin{tabular}{c|cc|cc}
 $y_{acc}^{(k)}$ &  complete & incomplete &  well-specified & misspecified\\
 \midrule
TabularToy(0.25) & 1.000 $\pm$ 0.000 & 0.786 $\pm$ 0.000 & 1.000 $\pm$ 0.000 & 0.687 $\pm$ 0.000 \\
dSprites(0)      & 1.000 $\pm$ 0.000 & 0.749 $\pm$ 0.004 & 1.000 $\pm$ 0.000 & 0.807 $\pm$ 0.000 \\
3dshapes(0)      & 1.000 $\pm$ 0.000 & 0.655 $\pm$ 0.002 & 1.000 $\pm$ 0.000 & 0.859 $\pm$ 0.008 \\
CUB      & 1.000 $\pm$ 0.000 & 0.836 $\pm$ 0.012 & 1.000 $\pm$ 0.000 & -- \\
HAM10K      & 0.999 $\pm$ 0.001 & 0.840 $\pm$ 0.009 & 0.999 $\pm$ 0.001 & -- \\
\end{tabular}
    \caption{First two columns: baseline $y_{acc}^{(k)}$ task accuracies of final heads trained on datasets with complete and incomplete sets of concepts. 
    Second two columns: baseline $y_{acc}^{(k)}$ task accuracies of a linear head on the linear (\textit{well-specified}) and non-linear (\textit{misspecified}) ground-truth task labels. We report 95\% confidence intervals over 5-fold training.}
    \label{table_c2y_task_performance_incomplete_concepts_misspecified}
\end{table}

\paragraph{Misspecified final head.} If the final head is not flexible enough to learn the ground-truth $y=f(c)$ dependence, during training then the model tends to store the necessary dependences into the learnt concepts, which are thereby deformed to encode additional information about the task and the other concepts. In many practical applications a common choice for the final head of a CBM is a linear layer which in principle allows for full model explainability. In reality, many tasks are best described by a non-linear function of the concepts, and a linear layer can be misspecified in such cases.

To explore the general effects of final head misspecification we considered soft CBMs with a linear head trained on datasets where the ground-truth task is either a linear or non-linear function of the concepts. To do so we modified the functional dependences in TabularToy(0.25), dSprites(0) and 3dshapes(0) by adding non-linear terms to the original tasks (see Appendix \ref{App_details_c2y_misspecification} for details). As a baseline measuring the final head misspecification, for each dataset we trained a separate linear head on the ground-truth concepts. Table  \ref{table_c2y_task_performance_incomplete_concepts_misspecified} shows the resulting decrease in task performance in terms of the reference task accuracy $y^{(k)}_{acc}$. 
The corresponding leakage scores are displayed in Figure \ref{figure_causes_c2y_misspecification} and their concepts and task accuracies are presented in Appendix \ref{App_accuracies_incomplete_concepts_misspecified_head}. According to the Leakage Criterion, a misspecified final head causes a higher leakage in 7/9 model classes. Moreover, high concept supervision is generally not sufficient to compensate for the additional leakage induced by the final head misspecification.

\begin{figure}[t]
\begin{minipage}[c]{0.32\textwidth} 
\centering
\includegraphics[width=\textwidth]{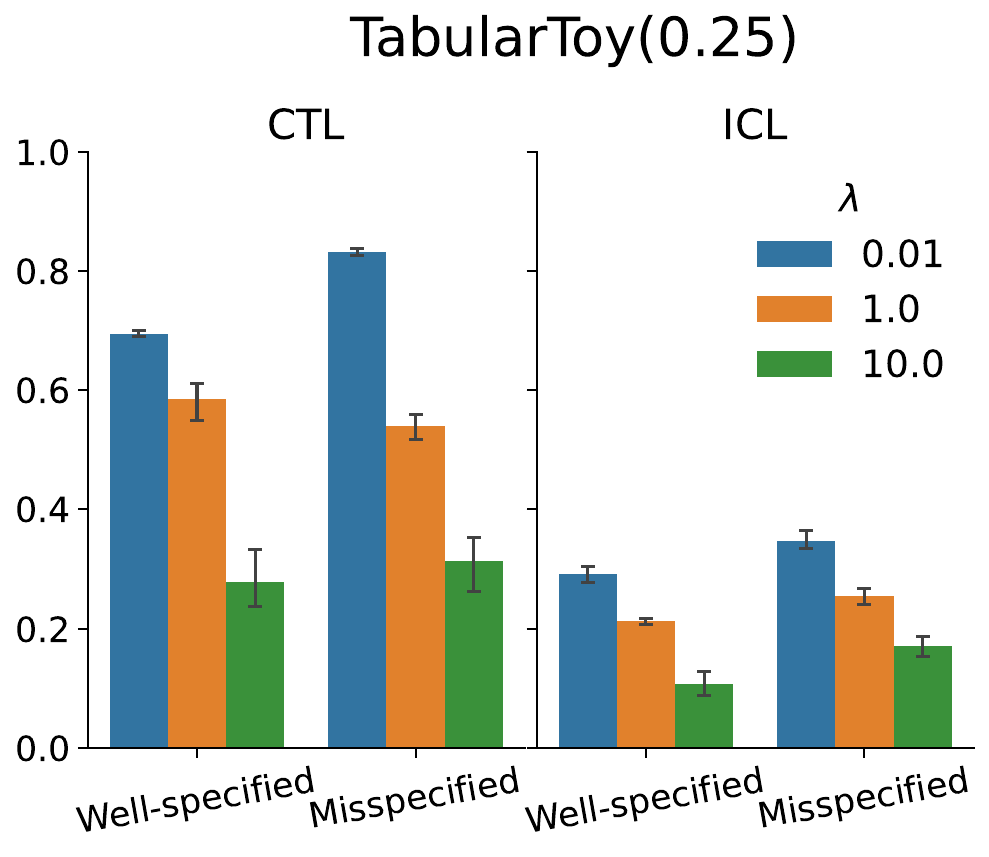} 
\end{minipage}
\begin{minipage}[c]{0.32\textwidth} 
\centering
\includegraphics[width=\textwidth]{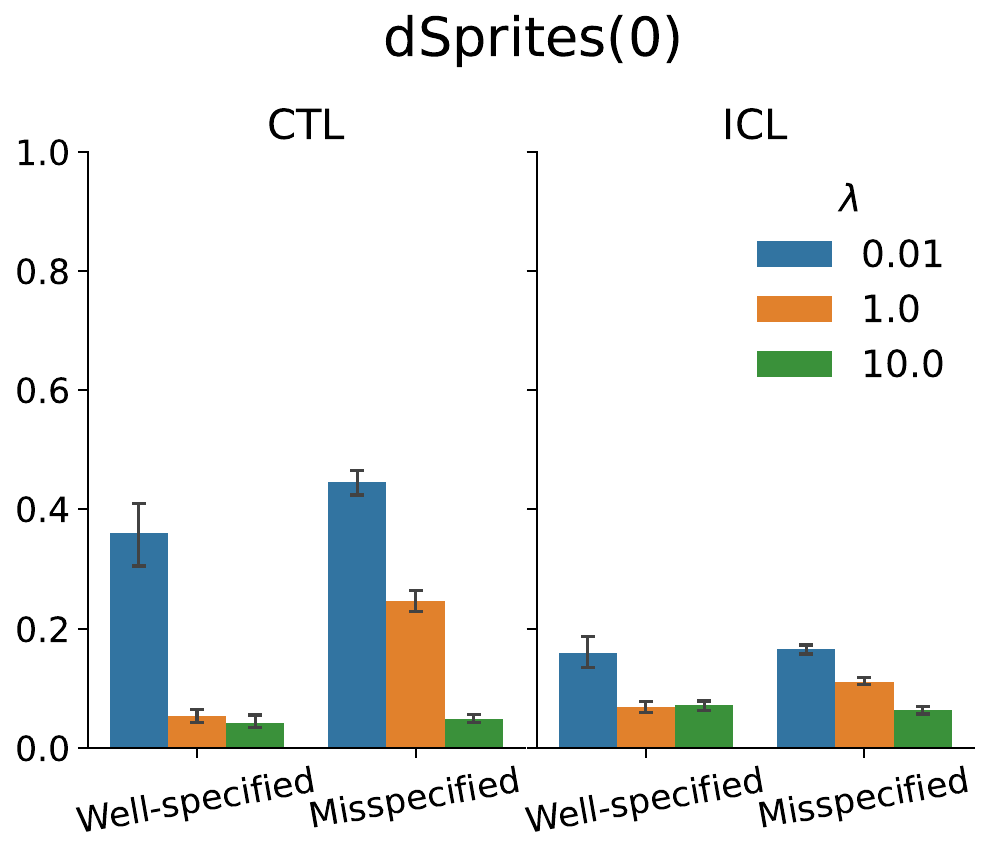} 
\end{minipage}
\begin{minipage}[c]{0.325\textwidth} 
\centering
\includegraphics[width=\textwidth]{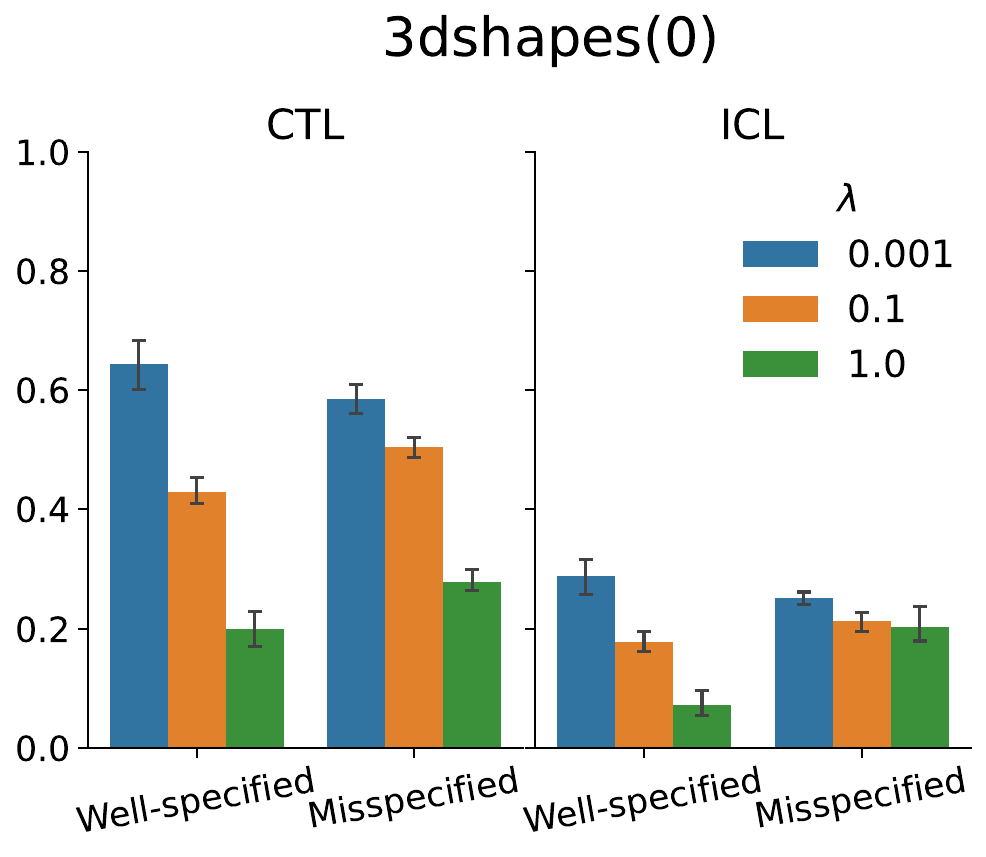} 
\end{minipage}
\caption{
Leakage scores evaluated for soft CBMs with a linear final head trained on datasets where the task is either a linear (\textit{Well-specified}) or a non-linear (\textit{Misspecified}) function of the concepts, at low, intermediate and high levels of concept supervision.
}
    \label{figure_causes_c2y_misspecification}
\end{figure}

\section{Leakage in CEMs}
\label{sec_interpretability_CEMs}

CEMs \citep{CEMs} are widely recognized for achieving higher task accuracy than CBMs, particularly when the concept set is incomplete, and can often approach the performance of end-to-end models without concept supervision. CEMs reasoning is based on data point-dependent concept vector representations, and not directly on concept activations as in CBMs. As noted in \cite{CEMs, IntCEMs, zarlenga2025avoiding}, this allows them to encode additional task-related information into concept vectors boosting task performance. In the following, we show that our information-theoretic framework confirms this picture, and that CEM's design results in concepts-task leakage for any value of the hyperparameters. Moreover, we find that CEMs exhibit significant interconcept leakage which, in contrast to CBMs, typically increases with higher concept supervision. Finally, we identify \emph{alignment leakage}, a novel subtype of concepts-task leakage that can arise in models exposed to interventions during training such as CEMs.
Taken together, these observations indicate that CEMs may be ill-suited for high-risk applications due to their limited interpretability.

\paragraph{Information-theoretic measures of CEM leakage.}
CEMs incorporate several concept representations -- the concept probabilities $\hat{c}_i$ just as CBMs, as well as the vector encodings $\bm{\hat{c}^+}_i, \bm{\hat{c}^-}_i$ and $\bm{\hat{c}^{w}}_i$. Reasoning happens at the level of the weighted vectors $\bm{\hat{c}^{w}}_i$, making them critical objects where leakage can hinder interpretability. 
Entropy and MI estimators are generally affected by biases depending on the dimension of variables, as a consequence of the curse of dimensionality \citep{Geometric_knn_18, Carrara19, czy2023beyond, gowri2024approximating}. Thus, one cannot compare quantities estimated for concept representations of different dimensions as required by the definitions of the CTL and ICL scores in equations \eqref{eq_CTL_i}-\eqref{eq_ICL_ij}, since this would involve subtracting normalised MIs estimated on the ground-truth concepts $c_i$ from those on the vectors $\bm{\hat{c}^{w}}_i$. To avoid this bias we will therefore analyse the behaviour of the normalised MIs on $\bm{\hat{c}^{w}}_i$ across different models without referring to ground-truth normalised MIs.

We define the concepts-task normalised MI between each vector $\bm{\hat{c}^{w}}_i$ and the task label, averaged across the $k$ concepts,
\begin{equation} \label{eq_I_CT_CEM}
    \widetilde{I}^{(CT)}\left(\bm{\hat{c}^{w}}, y\right) = 
    \frac{1}{k} \sum_{i = 1}^k \frac{I\left(\bm{\hat{c}^{w}}_i, y\right)}{H(y)}.
\end{equation}
This score captures how informative the weighted vectors $\bm{\hat{c}^{w}}_i$ are, on average, of the task. This score can be considered as a simpler version of the CTL score we introduced for CBMs, that does not use the ground-truth concepts-task MI as a reference. 

In a similar spirit to the ICL scores, we can also define a simplified metric capturing the amount of information that a concept representation $\bm{\hat{c}^{w}}_i$ encodes about the other concepts. Averaging over the $k(k-1)/2$ non-trivial concept pairs, we get:
\begin{equation} \label{eq_I_IC_CEM}
    \widetilde{I}^{(IC)} (\bm{\hat{c}^{w}}, c) = 
    \frac{2}{k(k-1)} \sum_{i = 1}^k \sum_{j < i}
    \frac{I(\bm{\hat{c}^{w}}_i, c_j)}{H(c_j)}.
\end{equation}
Note that in contrast to the ICL scores, $\widetilde{I}^{(IC)}$ does not measure how much a learnt concept representation is predictive of another. This approach, based on annotated concepts instead, has the benefit of limiting the overall dimensionality of the space used to estimate MIs (in this case $d+1$ instead of 2$d$), mitigating the effects of dimension-associated bias.

An associated quantity that is relevant whenever higher-dimensional representations contain non-trivial information is the normalised MI of a concept representation $\bm{\hat{c}^{w}}_i$ relative to its ground-truth concept $c_i$. Its average over the $k$ concepts is:
\begin{equation} \label{eq_I_self_CEM}
    \widetilde{I}^{\text{(self)}} (\bm{\hat{c}^{w}}, c) = 
    \frac{1}{k} \sum_{i = 1}^k
    \frac{I(\bm{\hat{c}^{w}}_i, c_i)}{H(c_i)},
\end{equation}
which provides a measure of how predictive each concept representation is of its own concept, averaged over concepts.

\paragraph{Our framework detects CEM's concepts-task leakage.}
By design, the concept representations $\bm{\hat{c}^+}_i$, $\bm{\hat{c}^-}_i$ and $\bm{\hat{c}^{w}}_i$ in CEMs are encouraged to encode additional task-relevant information besides the concept state itself, with the aim of improving task performance. This is a consequence of the fact that no direct supervision is performed on the inputs $\bm{\hat{c}^{w}}_i$ to the final head, but only on the predicted concept activations $\hat{c}_i$ via the concept reconstruction loss. This means that in the forward pass, an arbitrary amount of information is allowed to flow from the input $x$ to the task prediction through the concept vectors, as long as the embeddings $\bm{\hat{c}^+}_i$ and $\bm{\hat{c}^-}_i$ generated from $x$ are predictive of the scalar $c_i$.

\begin{figure}[t]
\begin{minipage}[c]{0.23\textwidth} 
\centering \small \sffamily
$\;\;\;\;\;$ TabularToy(0.25)
\vspace{0.2cm}
\end{minipage}
\hfill
\begin{minipage}[c]{0.23\textwidth} 
\centering \small \sffamily
dSprites(0) \hspace{0.67cm}
\vspace{0.2cm}
\end{minipage}
\begin{minipage}[c]{0.23\textwidth}
\centering \small \sffamily
 3dshapes(0) \hspace{0.35cm}
\vspace{0.2cm}
\end{minipage}
\begin{minipage}[c]{0.23\textwidth}
\centering \small \sffamily
 \hspace{0.35cm}HAM10K
\vspace{0.2cm}
\end{minipage}

\begin{minipage}[c]{0.24\textwidth} 
\centering
\includegraphics[width=0.83\textwidth]{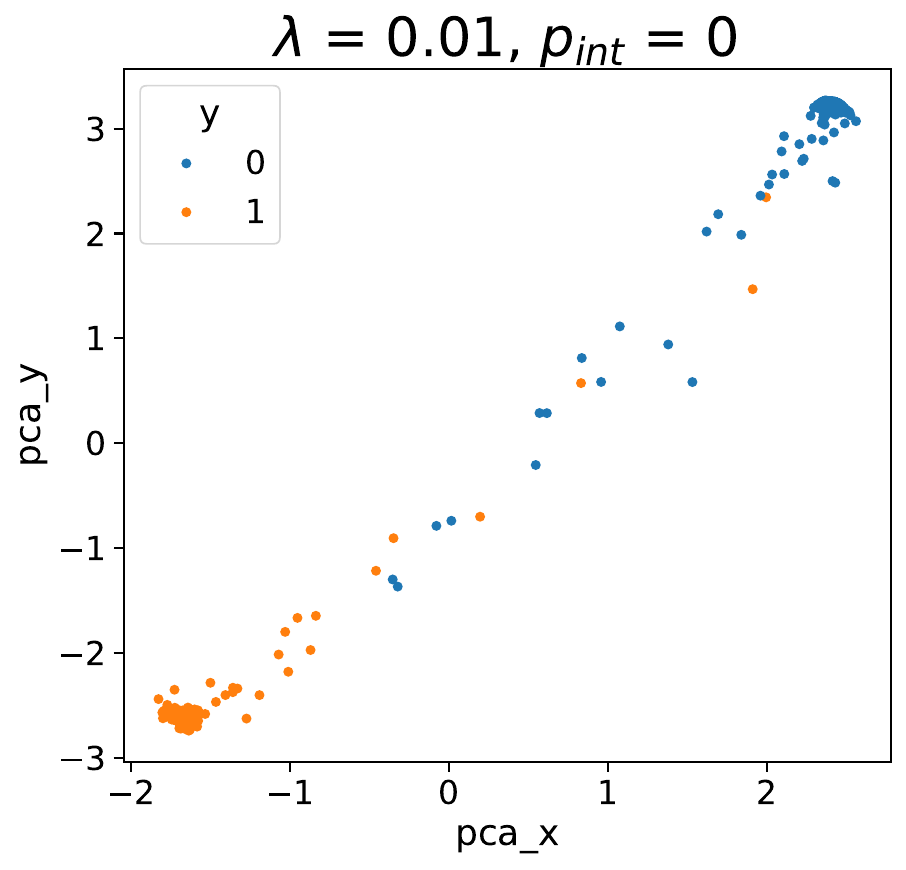} 
\end{minipage}
\hfill
\begin{minipage}[c]{0.24\textwidth} 
\centering
    \includegraphics[width=0.96\textwidth]{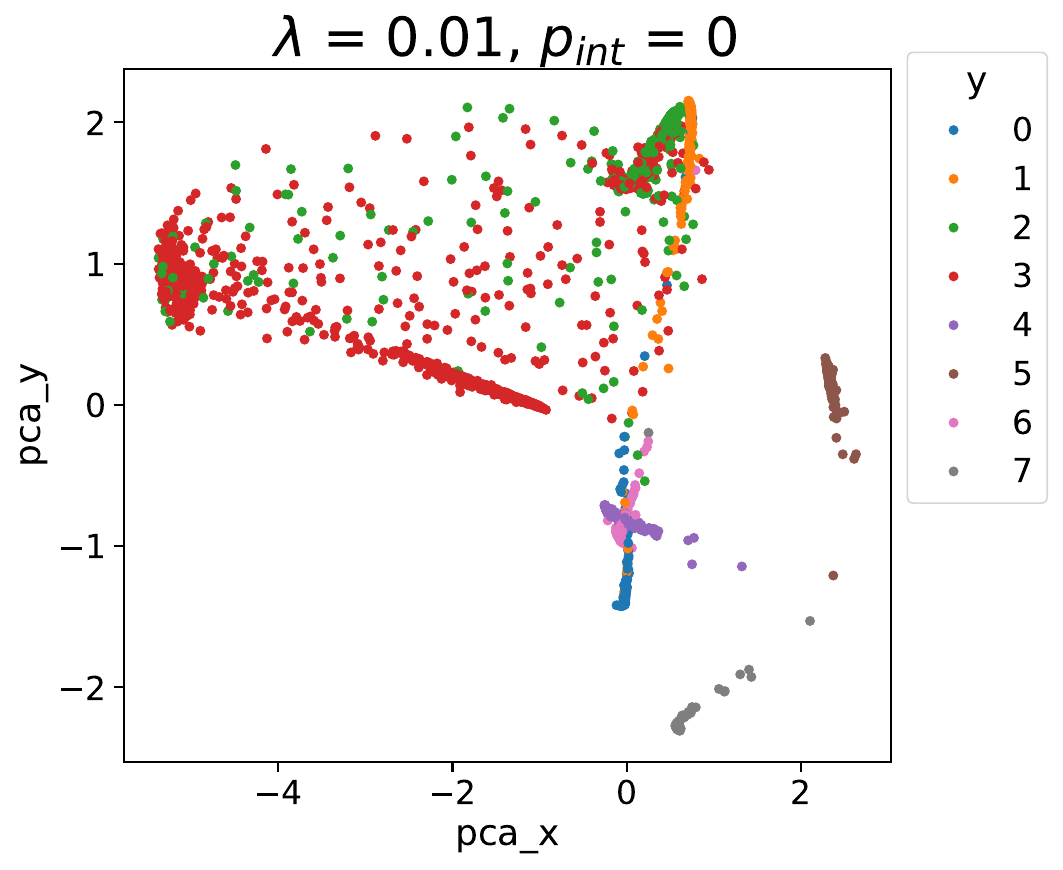} 
\end{minipage}
\begin{minipage}[c]{0.24\textwidth} 
\centering
    \includegraphics[width=1\textwidth]{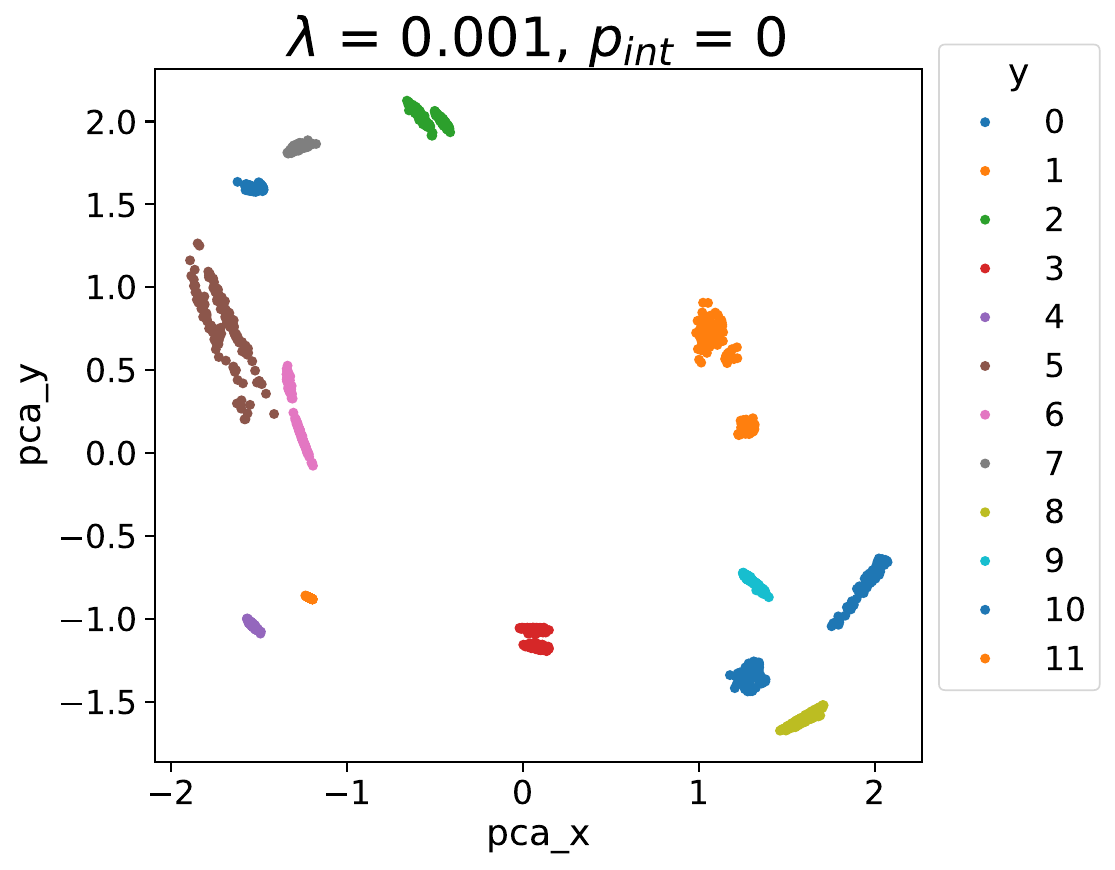} 
\end{minipage}
\begin{minipage}[c]{0.24\textwidth} 
\centering
    \includegraphics[width=0.84\textwidth]{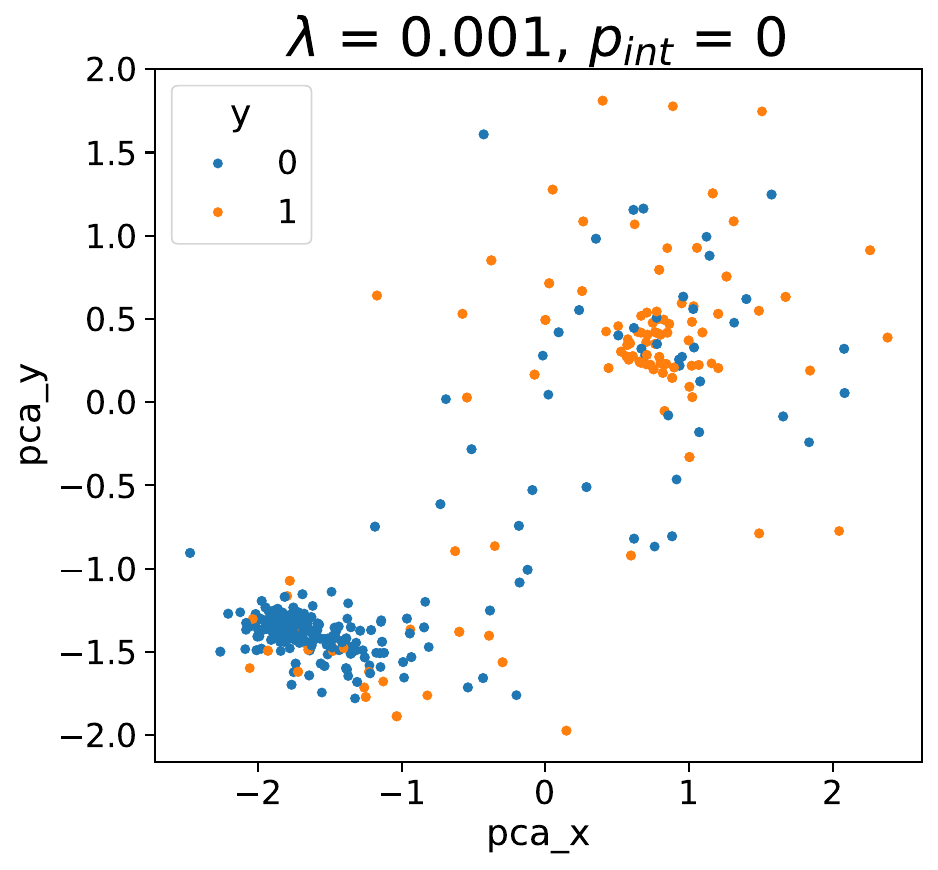} 
\end{minipage}

\begin{minipage}[c]{0.24\textwidth} 
\centering
\includegraphics[width=0.83\textwidth]{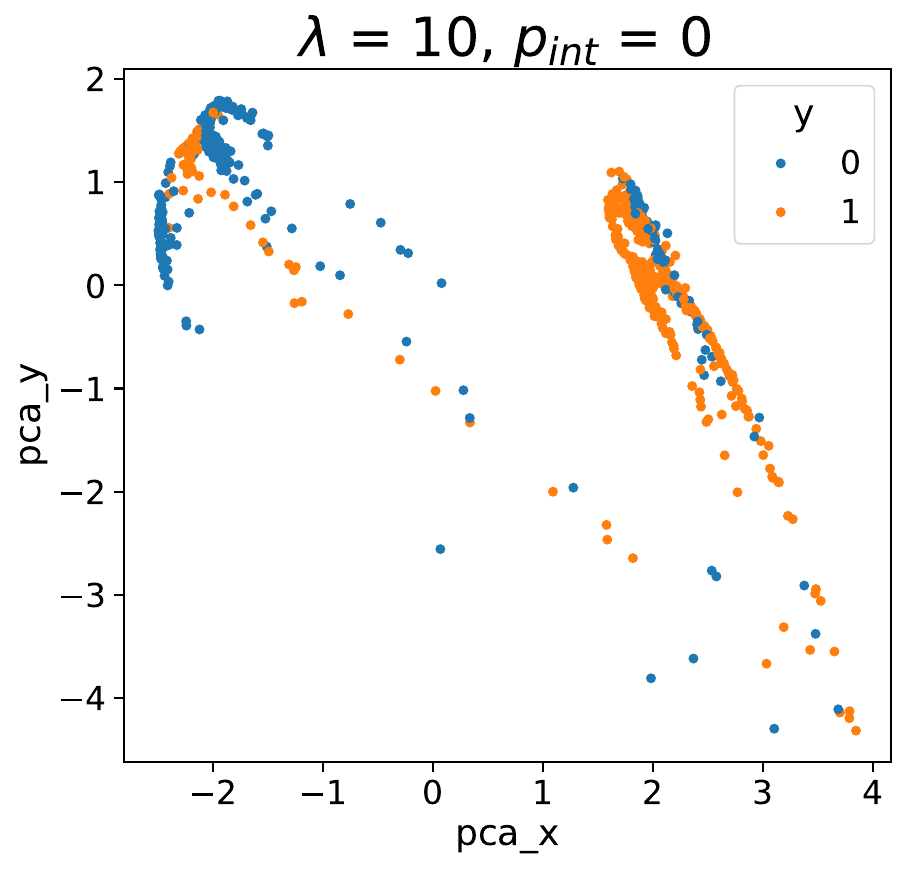} 
\end{minipage}
\hfill
\begin{minipage}[c]{0.24\textwidth} 
\centering
    $\;$\includegraphics[width=0.96\textwidth]{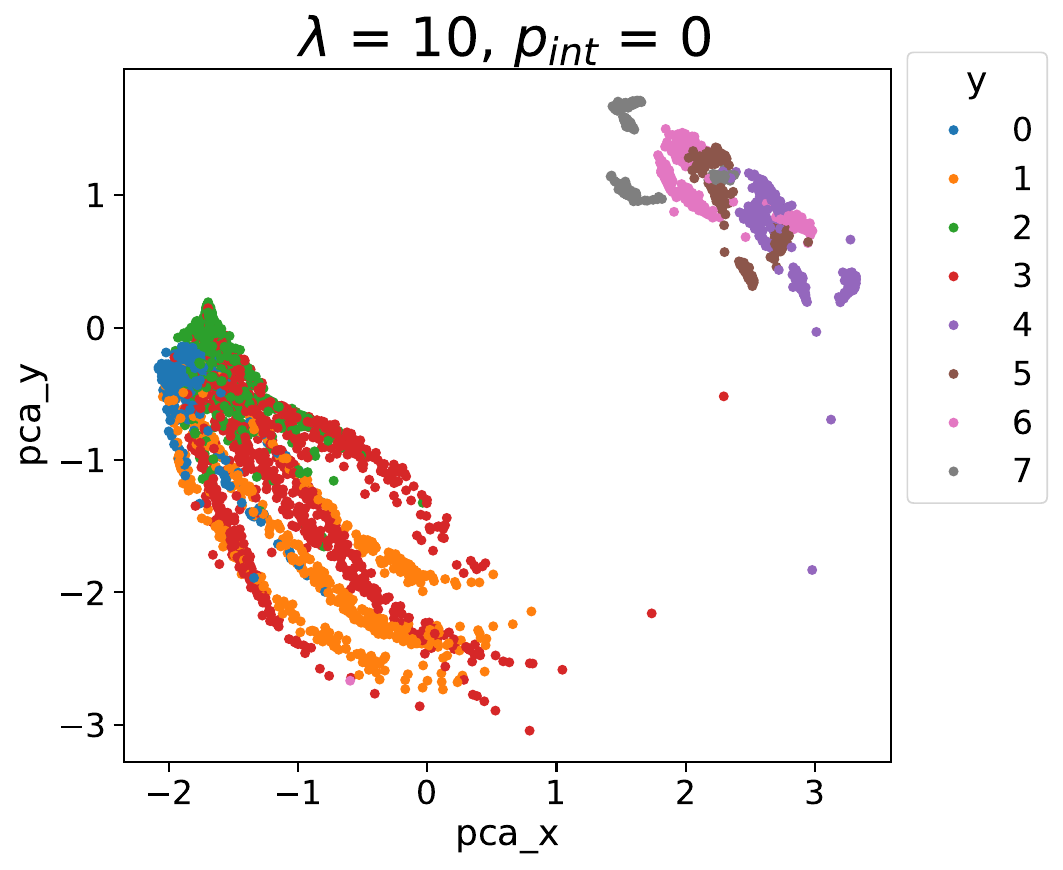} 
\end{minipage}
\begin{minipage}[c]{0.24\textwidth} 
\centering
    \includegraphics[width=0.99\textwidth]{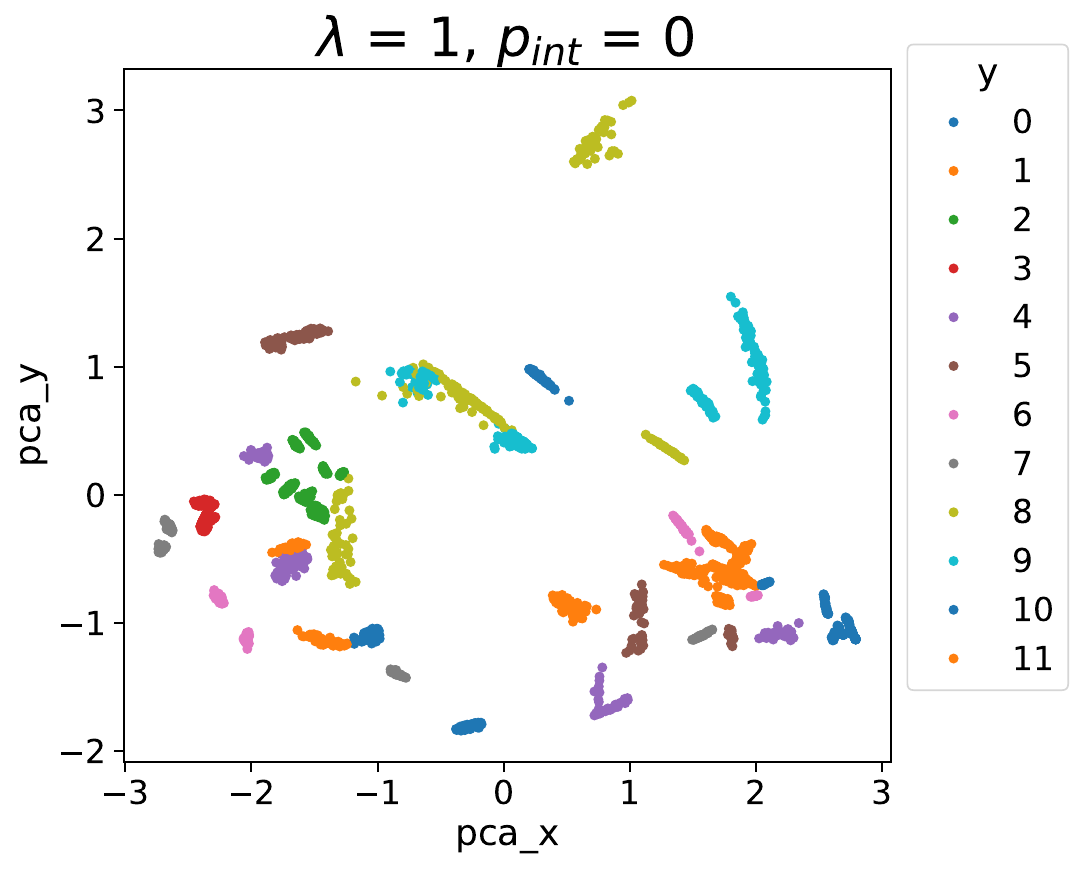} 
\end{minipage}
\begin{minipage}[c]{0.24\textwidth} 
\centering
    \includegraphics[width=0.84\textwidth]{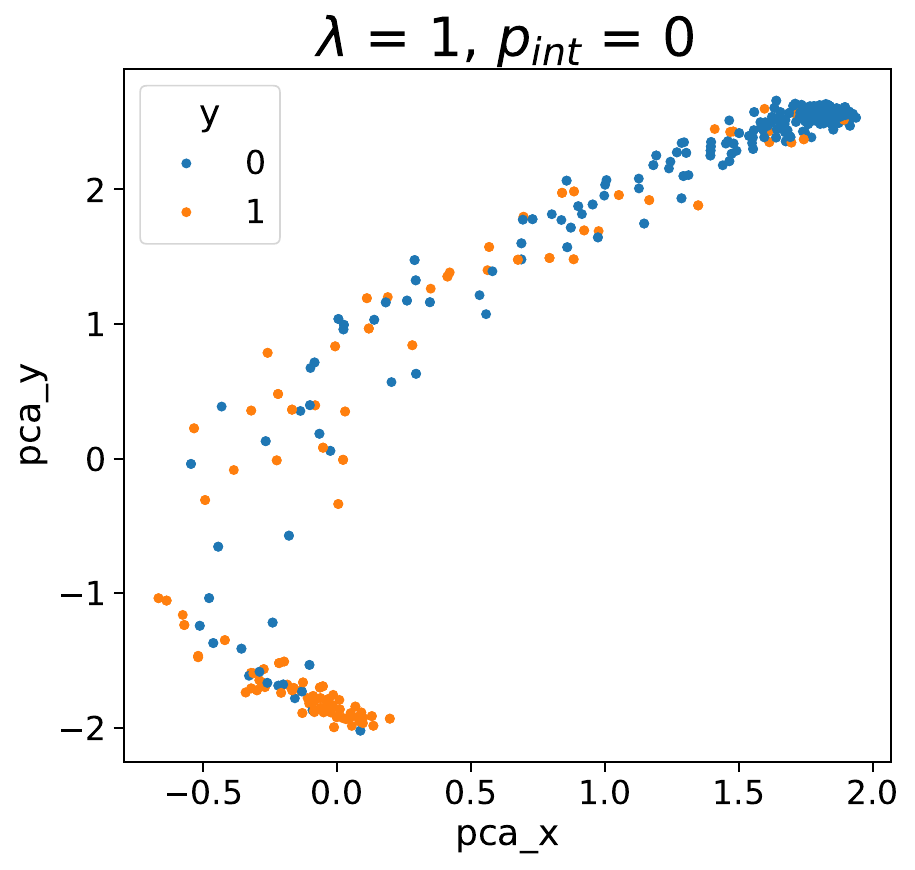} 
\end{minipage}

    \caption{
    2-dimensional PCA projections of the weighted embeddings $\bm{\hat{c}^{w}}_1$ for the first concept across datasets and for different values of $\lambda$ at $p_{int}=0$. The colouring indicates the value of the ground-truth task label. See \cite{CEMs} for similar representations in CUB.
    }
    \label{figure_PCAs_y_main}
\end{figure}

An effect of this task-relevant information is the clear clustering of the embeddings based on the value of the task label, as shown in Figure \ref{figure_PCAs_y_main} (see also  Figure \ref{figure_PCAs_y_app} for similar results at non-vanishing $p_{int}$, as well as \cite{CEMs, Santis2025VCEMBP}). This parallels the structure that appears in the concept distributions of CBMs affected by leakage (in this case, accounting for higher dimensional concept representations).

\begin{figure}[t]
\begin{minipage}[c]{0.195\textwidth} 
\centering
\includegraphics[width=1\textwidth]{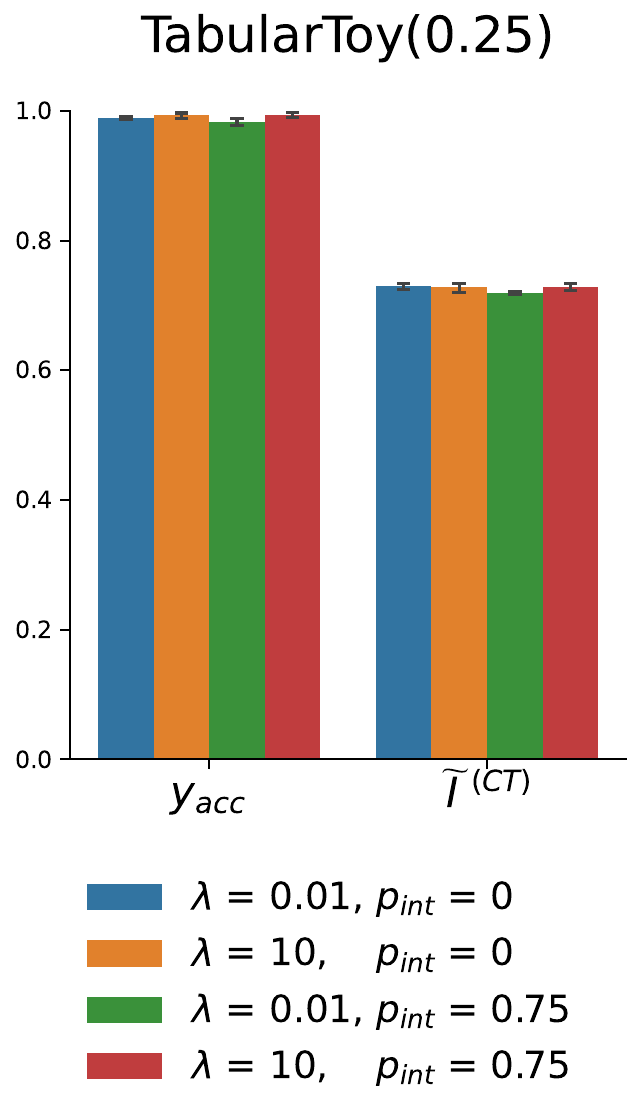} 
\end{minipage}
\begin{minipage}[c]{0.195\textwidth} 
\centering
\includegraphics[width=1\textwidth]{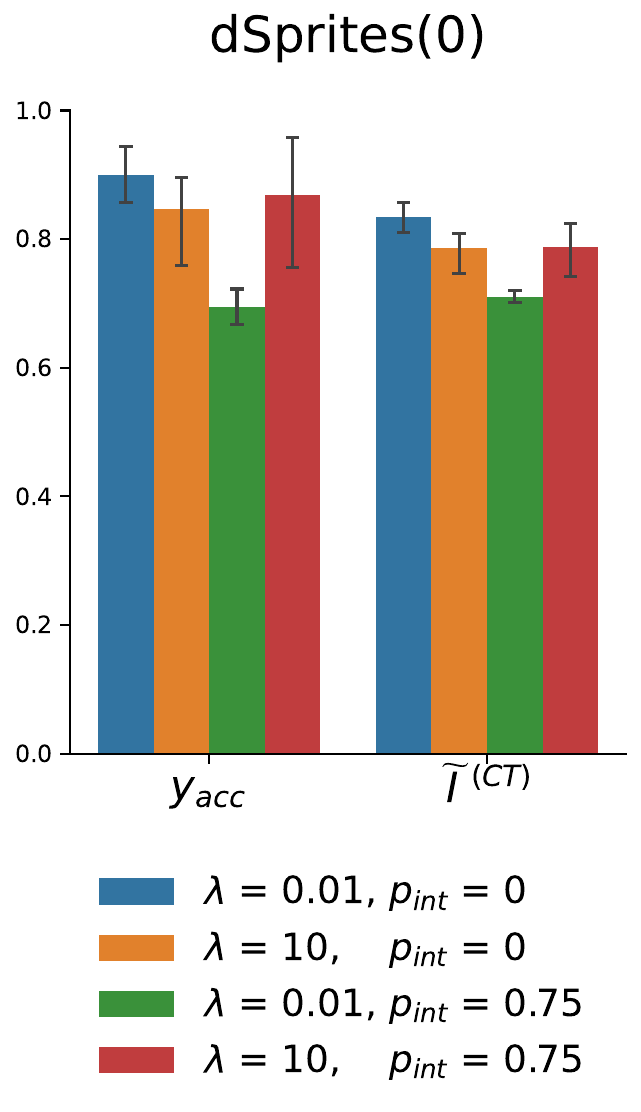} 
\end{minipage}
\begin{minipage}[c]{0.195\textwidth} 
\centering
\includegraphics[width=1\textwidth]{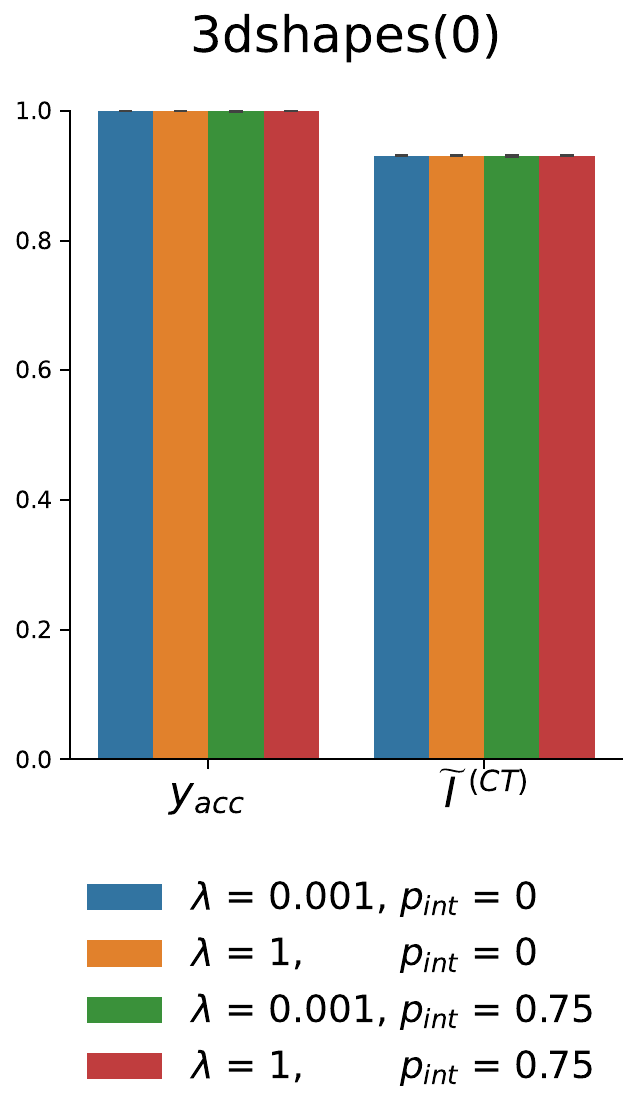} 
\end{minipage}
\begin{minipage}[c]{0.195\textwidth} 
\centering
\includegraphics[width=1\textwidth]{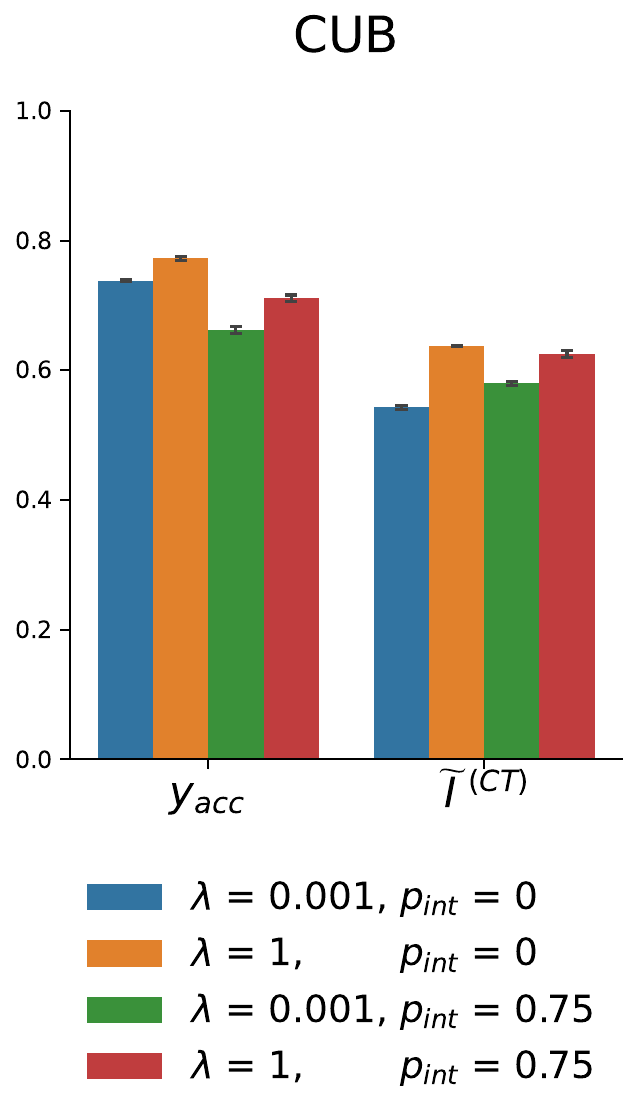} 
\end{minipage}
\begin{minipage}[c]{0.195\textwidth} 
\centering
\includegraphics[width=1\textwidth]{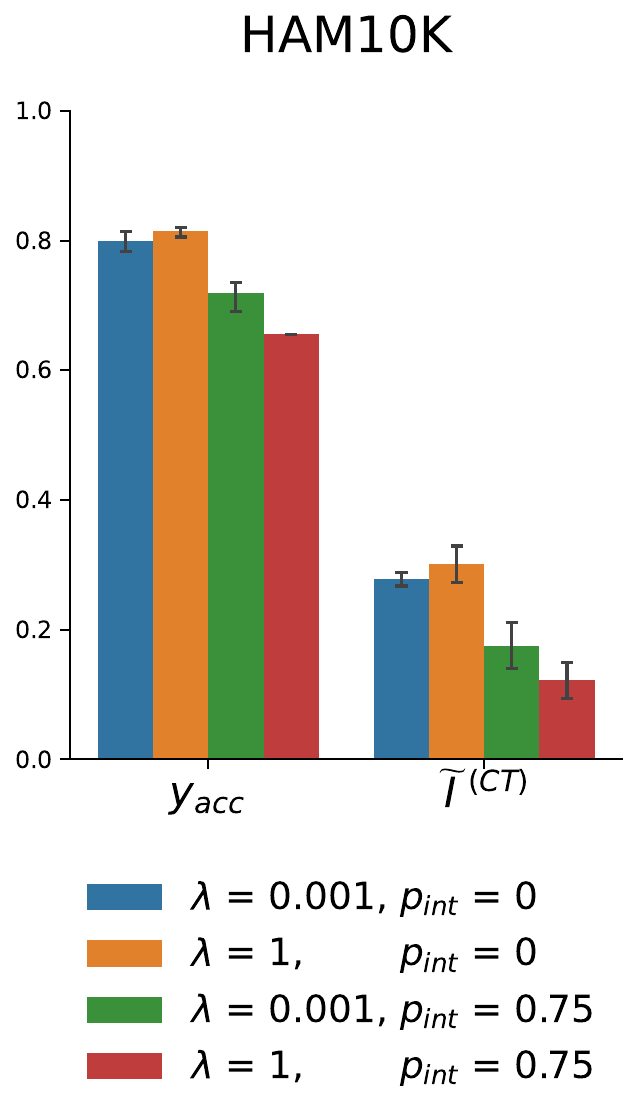} 
\end{minipage}
\caption{
Task accuracy and $\widetilde{I}^{(CT)}$ score on a range of CEMs and datasets. 
}
    \label{figure_CEM_CTL}
\end{figure}

Information theory offers additional insight into this phenomenon, in particular via the $\widetilde{I}^{(CT)}\left(\bm{\hat{c}^{w}}, y\right)$ score defined in \eqref{eq_I_CT_CEM}. 
As shown in Figure \ref{figure_CEM_CTL} this score is representative of the degree of structure in the learnt embeddings given in Figures \ref{figure_PCAs_y_main} and \ref{figure_PCAs_y_app}, and is closely associated with task accuracy. On the other hand, concept accuracy does not correlate with task accuracy or $\widetilde{I}^{(CT)}$. This not only indicates the presence of concepts-task leakage, but also suggests that CEMs rely heavily on concepts-task leakage for their success: regardless of concepts, task accuracy is higher in models where the vectors $\bm{\hat{c}^{w}}_i$ encode more information about $y$. This picture agrees with the previous evidence that task accuracy is insensitive to the number of used concepts (see \cite{CEMs}, Appendix 8). In particular, when training CEMs on the CUB dataset with up to 90\% of the annotated concepts removed, the task accuracy remains essentially the same as that of models trained on all the concepts.

\begin{figure}[t]
\begin{minipage}[c]{0.23\textwidth} 
\centering \small \sffamily
$\;\;\;\;\;$ TabularToy(0.25)
\vspace{0.2cm}
\end{minipage}
\hfill
\begin{minipage}[c]{0.23\textwidth} 
\centering \small \sffamily
dSprites(0) \hspace{0.67cm}
\vspace{0.2cm}
\end{minipage}
\begin{minipage}[c]{0.23\textwidth}
\centering \small \sffamily
 3dshapes(0) \hspace{0.35cm}
\vspace{0.2cm}
\end{minipage}
\begin{minipage}[c]{0.23\textwidth}
\centering \small \sffamily
 HAM10K \hspace{0.18cm}
\vspace{0.2cm}
\end{minipage}

\begin{minipage}[c]{0.24\textwidth} 
\centering
\includegraphics[width=0.99\textwidth]{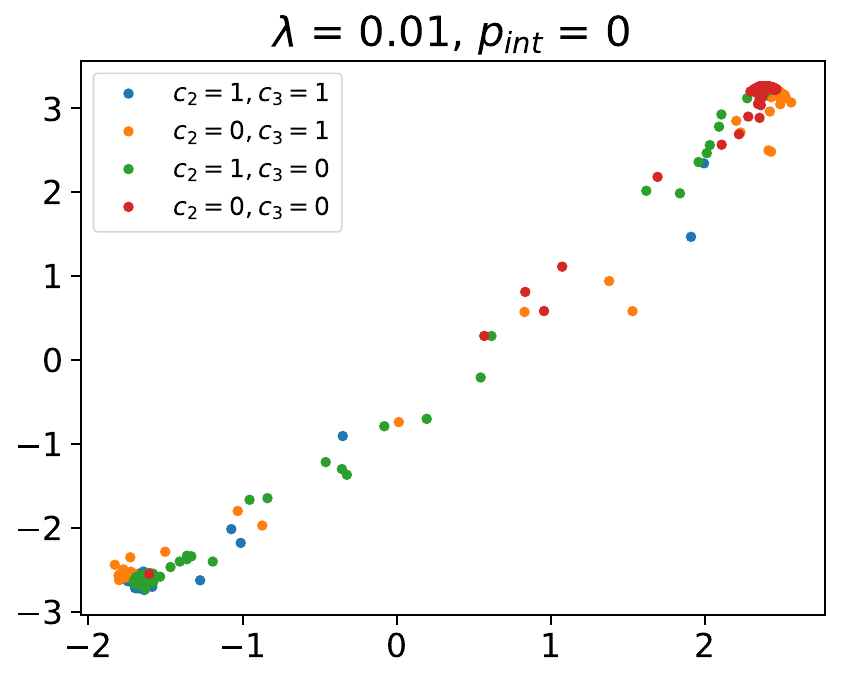} 
\end{minipage}
\begin{minipage}[c]{0.24\textwidth} 
\centering
    \includegraphics[width=1\textwidth]{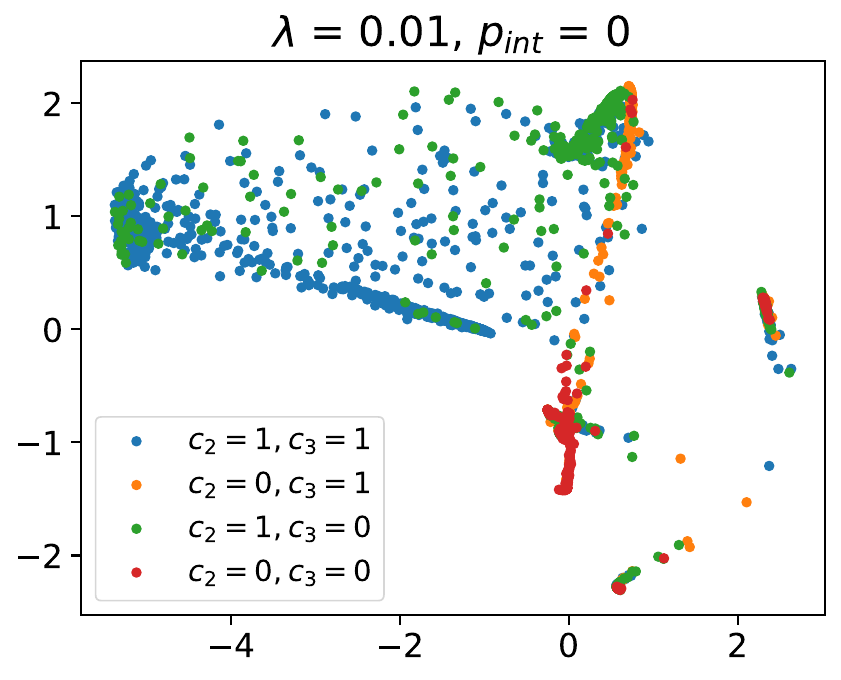} 
\end{minipage}
\begin{minipage}[c]{0.24\textwidth} 
\centering
    \includegraphics[width=1\textwidth]{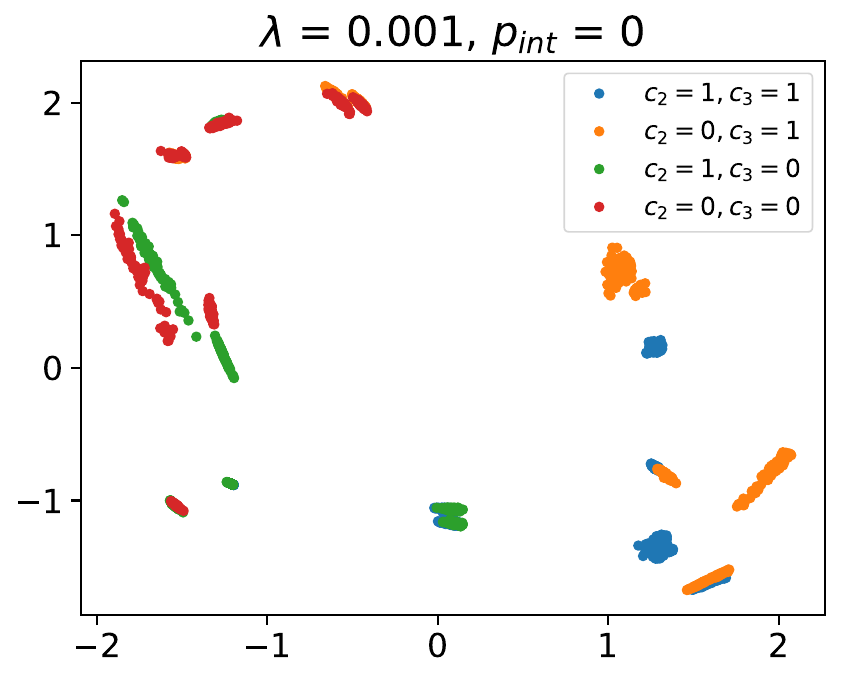} 
\end{minipage}
\begin{minipage}[c]{0.24\textwidth} 
\centering
\includegraphics[width=0.99\textwidth]{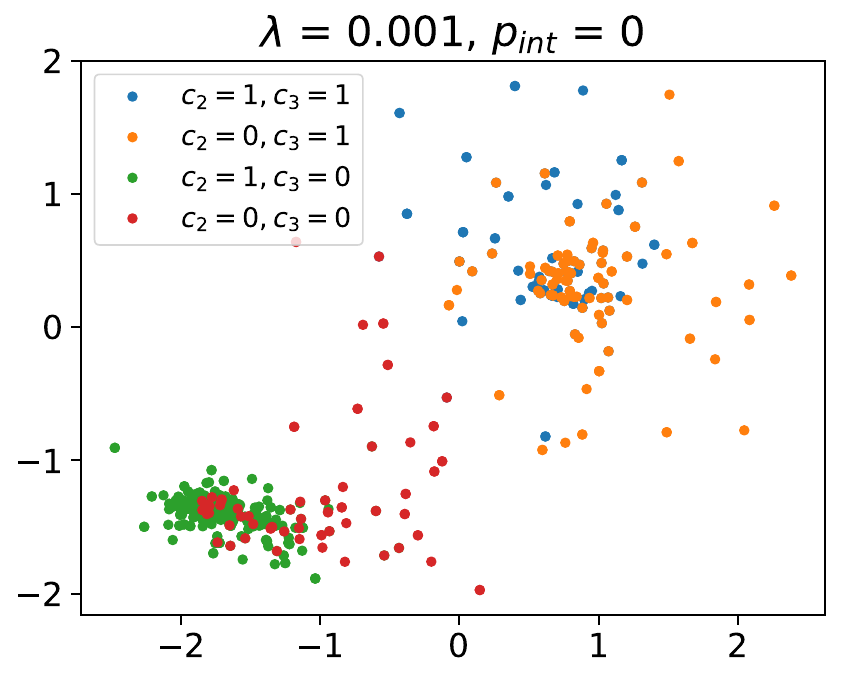} 
\end{minipage}

\begin{minipage}[c]{0.24\textwidth} 
\centering
\includegraphics[width=0.98\textwidth]{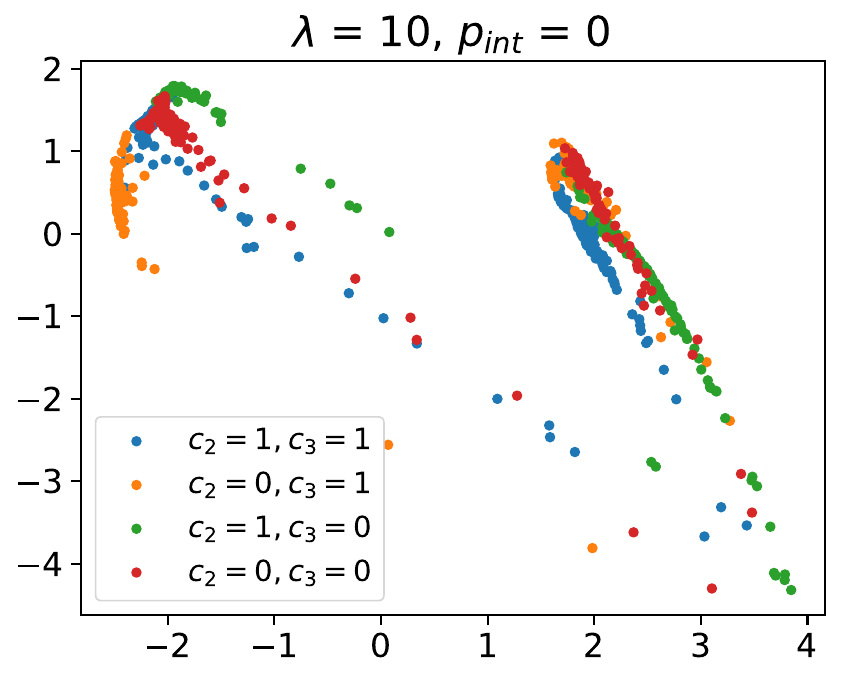} 
\end{minipage}
\begin{minipage}[c]{0.24\textwidth} 
\centering
    $\;$\includegraphics[width=0.99\textwidth]{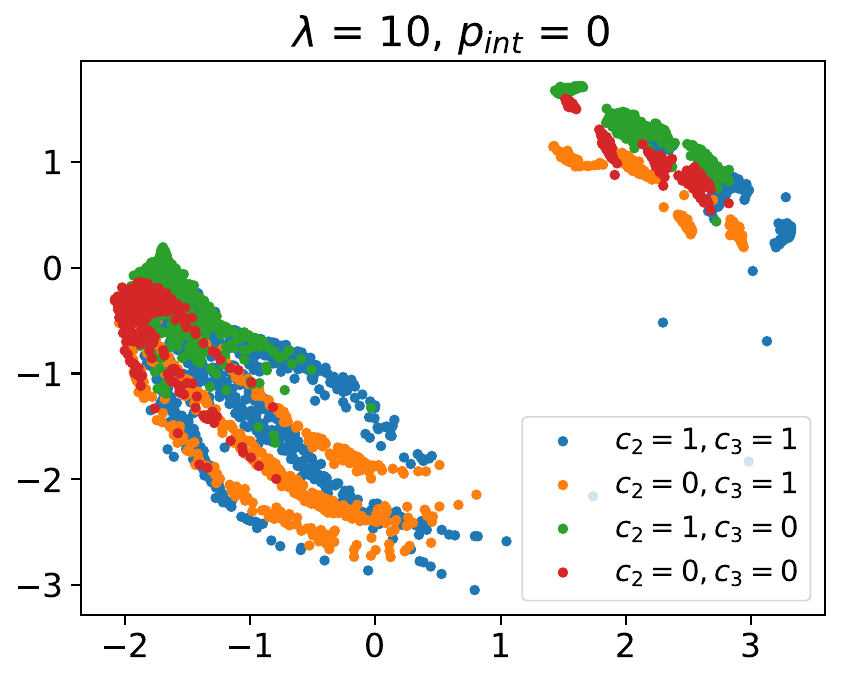} 
\end{minipage}
\begin{minipage}[c]{0.24\textwidth} 
\centering
    \includegraphics[width=0.99\textwidth]{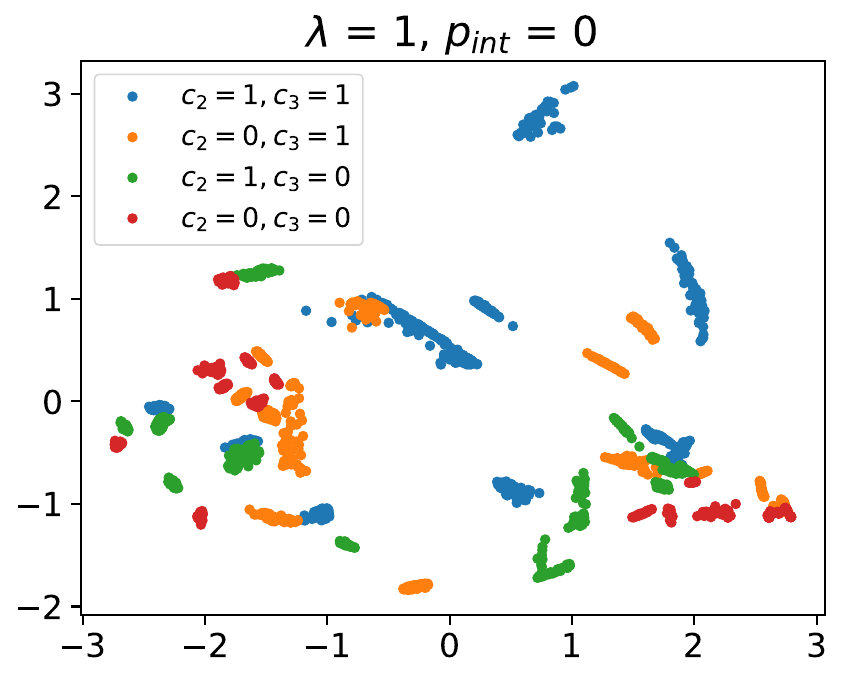} 
\end{minipage}
\begin{minipage}[c]{0.24\textwidth} 
\centering
\includegraphics[width=1\textwidth]{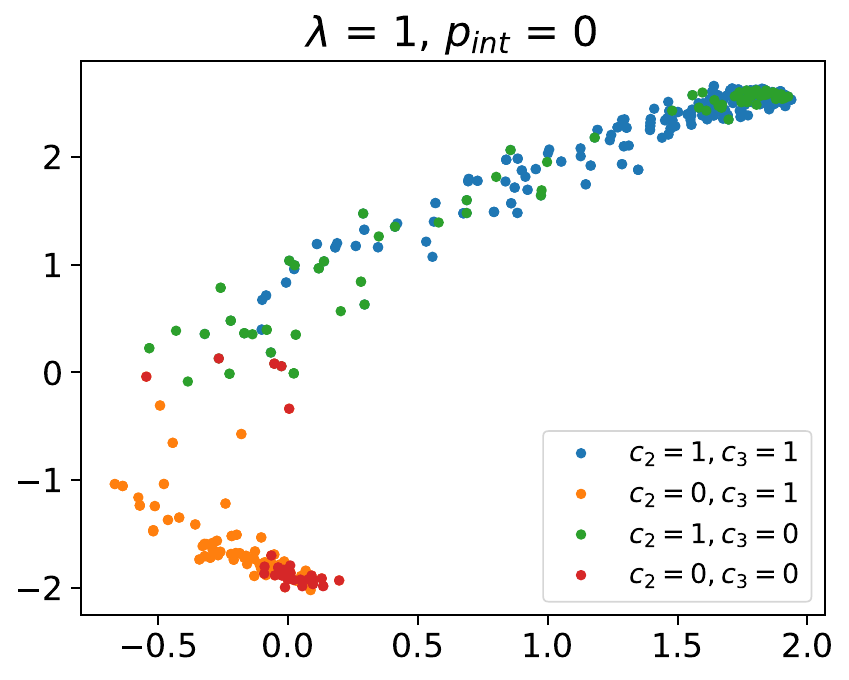} 
\end{minipage}
    \caption{
    2-dimensional PCA projections of the weighted embeddings $\bm{\hat{c}^{w}}_1$ for the first concept across datasets and for different values of $\lambda$ at $p_{int}=0$. The colouring indicates the ground-truth value of concepts 2 and 3.
    }
    \label{figure_PCAs_c_main}
\end{figure}

\begin{figure}[t]
\begin{minipage}[c]{0.195\textwidth} 
\centering
\includegraphics[width=1\textwidth]{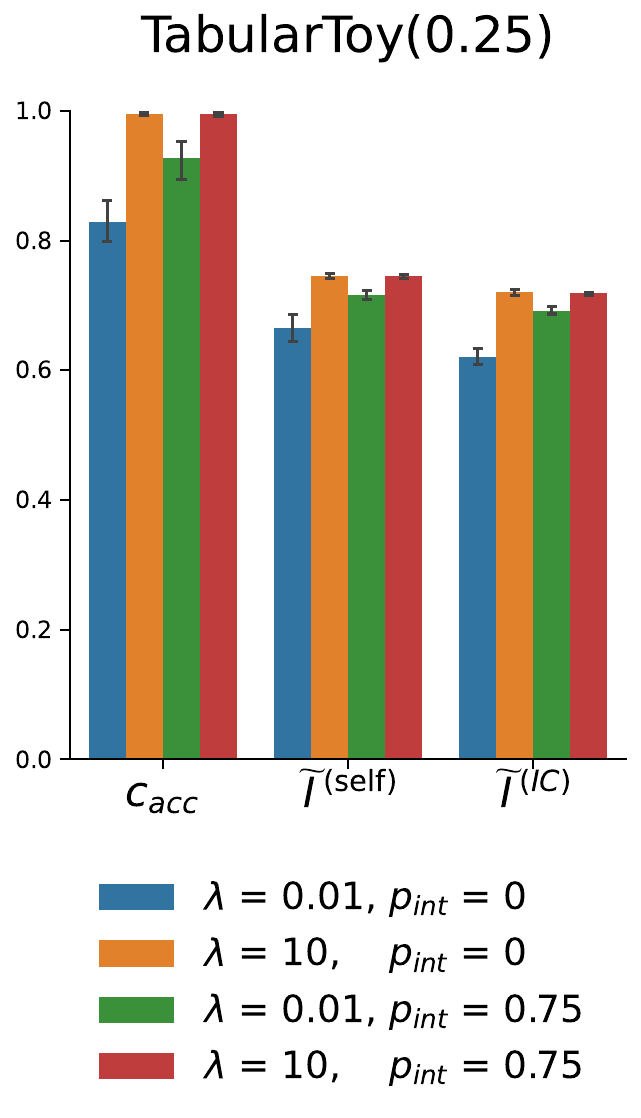} 
\end{minipage}
\begin{minipage}[c]{0.195\textwidth} 
\centering
\includegraphics[width=1\textwidth]{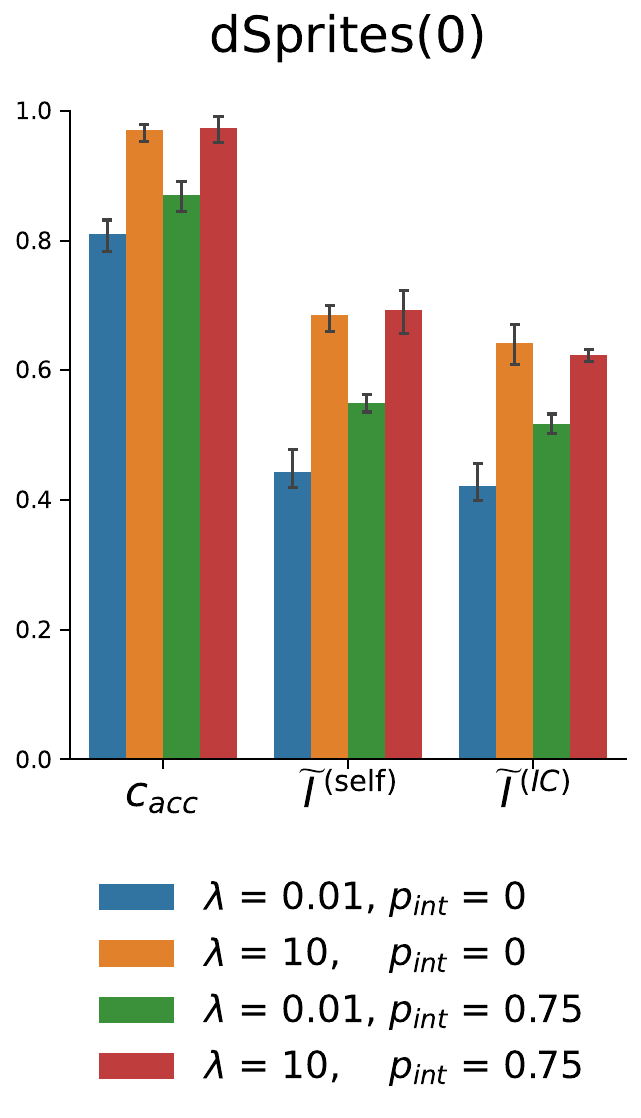} 
\end{minipage}
\begin{minipage}[c]{0.195\textwidth} 
\centering
\includegraphics[width=1\textwidth]{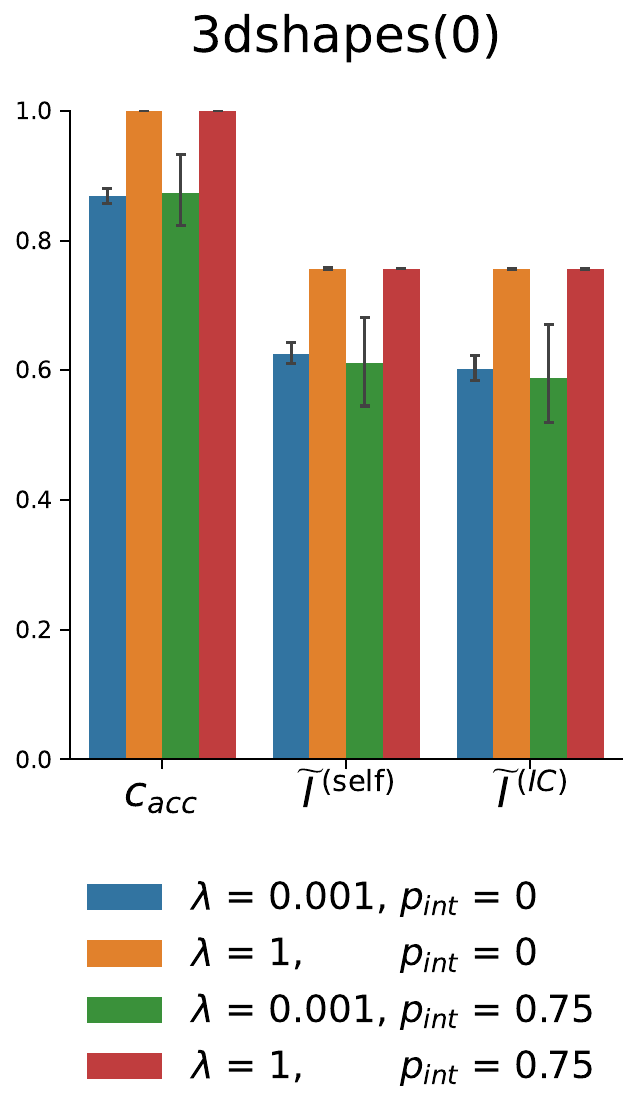} 
\end{minipage}
\begin{minipage}[c]{0.195\textwidth} 
\centering
\includegraphics[width=1\textwidth]{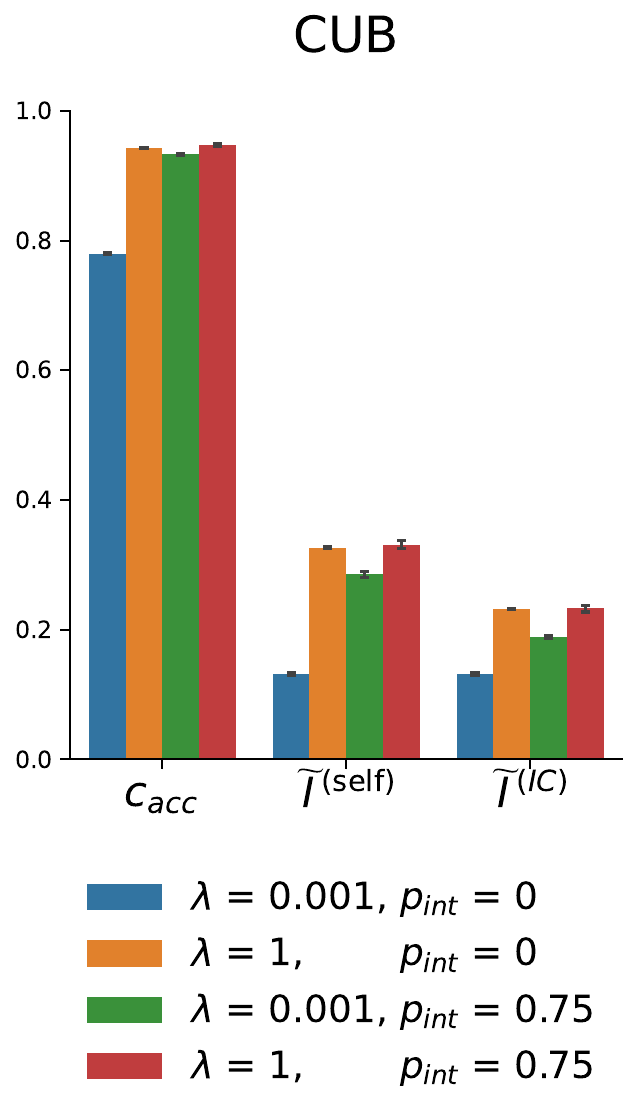} 
\end{minipage}
\begin{minipage}[c]{0.195\textwidth} 
\centering
\includegraphics[width=1\textwidth]{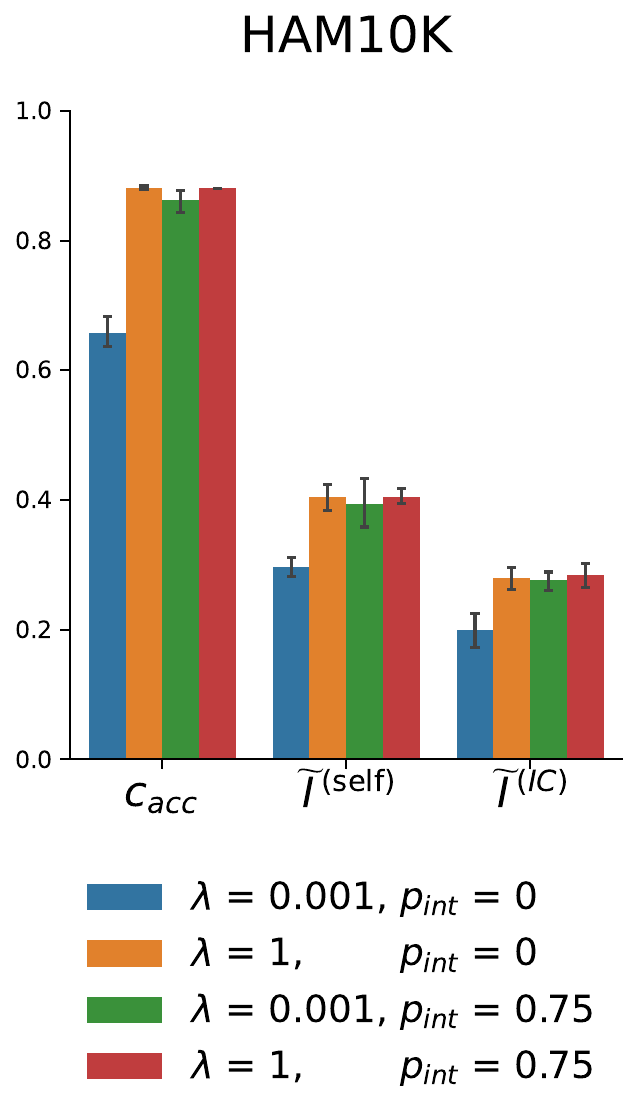} 
\end{minipage}
\caption{Concept accuracy, $\widetilde{I}^{(IC)}$ and $\widetilde{I}^{\text{(self)}}$ scores for CEMs with low and high $\lambda$ and $p_{int}$.
}
    \label{figure_CEM_ICL}  
\end{figure}

\paragraph{Interconcept leakage increases with concept supervision.}
On closer inspection of the structure of the learned embeddings $\bm{\hat{c}^{w}}_i$, strong clustering is also observed based on the ground-truth values of the other concepts $j \neq i$. For example, Figure \ref{figure_PCAs_c_main} shows PCA projections of $\bm{\hat{c}^{w}}_1$ in terms of the values of $c_2$ and $c_3$ (see Figure \ref{figure_PCAs_c_app} for similar plots for models with non-vanishing $p_{int}$). This phenomenon is particularly apparent in the TabularToy(0.25) data, where the task is binary and there are three concepts: in this case, the fine structure present in Figures \ref{figure_PCAs_y_main} and \ref{figure_PCAs_y_app} which is not explained by the $y$ labels is clearly associated with the value of the other concepts.

This effect can be quantified using the information-theoretic measures $\widetilde{I}^{(IC)} (\bm{\hat{c}^{w}}, c)$ and $\widetilde{I}^{\text{(self)}} (\bm{\hat{c}^{w}}, c)$ defined in equations \eqref{eq_I_IC_CEM}-\eqref{eq_I_self_CEM}. Each vector $\bm{\hat{c}^{w}}_i$ becomes more and more predictive of the corresponding $c_i$ as one increases $\lambda$ and $p_{int}$, as indicated in Figure \ref{figure_CEM_ICL} by higher $c_{acc}$ and $\widetilde{I}^{\text{(self)}}$. This is the expected and desired behaviour as it corresponds to increasing concept supervision. However, $\widetilde{I}^{(IC)}$ undergoes a very similar growth, indicating that the vectors $\bm{\hat{c}^{w}}_i$ contain increasing information about the other concepts, which, by definition, is interconcept leakage.
Thus, and in contrast to CBMs, there is no choice of the hyperparameters $\lambda$ and $p_{int}$ that improves concept learning while decreasing interconcept leakage.

\begin{figure}[t]
\begin{minipage}[c]{0.32\textwidth} 
\centering
\includegraphics[width=1\textwidth]{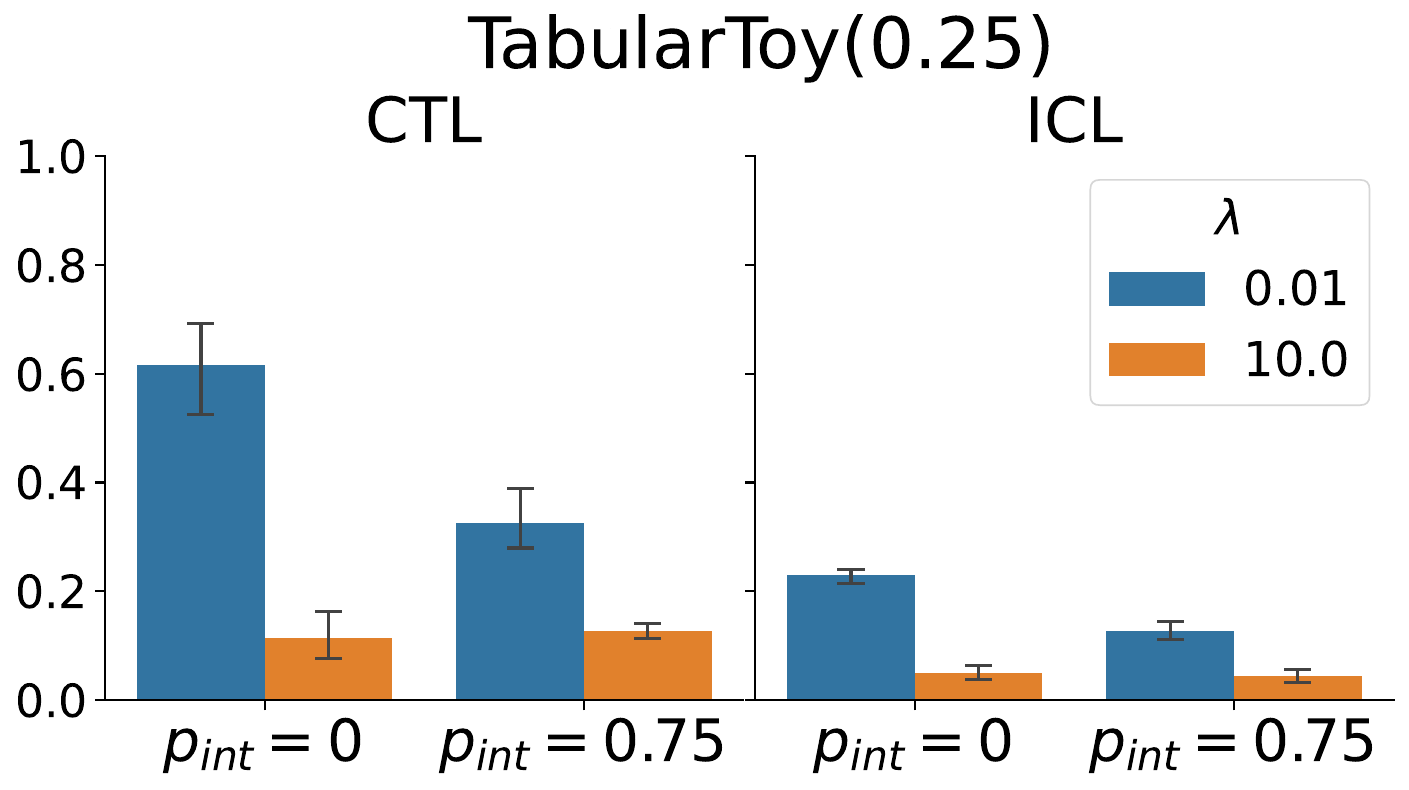} 
\end{minipage}
\hfill
\begin{minipage}[c]{0.32\textwidth}
\centering
    \includegraphics[width=1\textwidth]{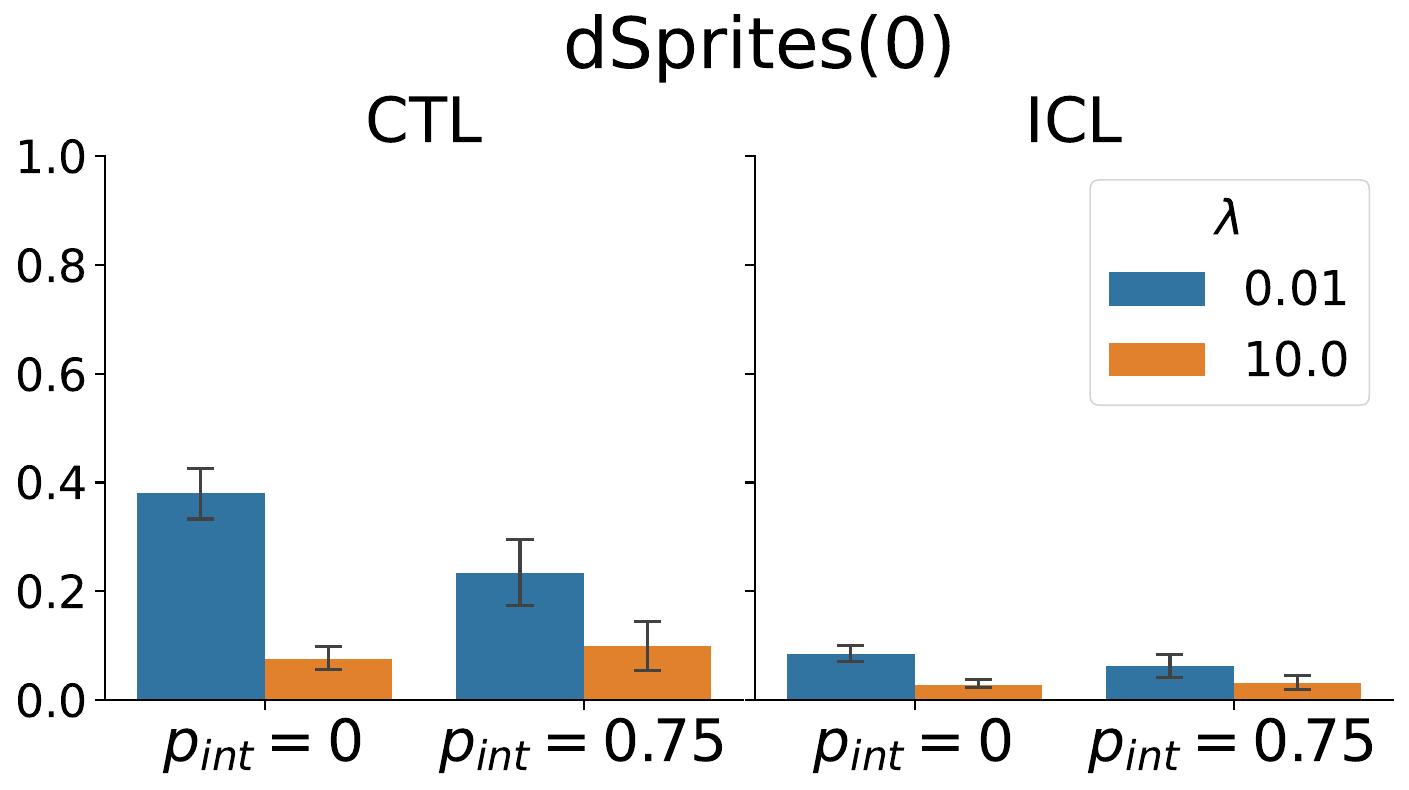} 
\end{minipage}
\begin{minipage}[c]{0.32\textwidth}
\centering
    \includegraphics[width=1\textwidth]{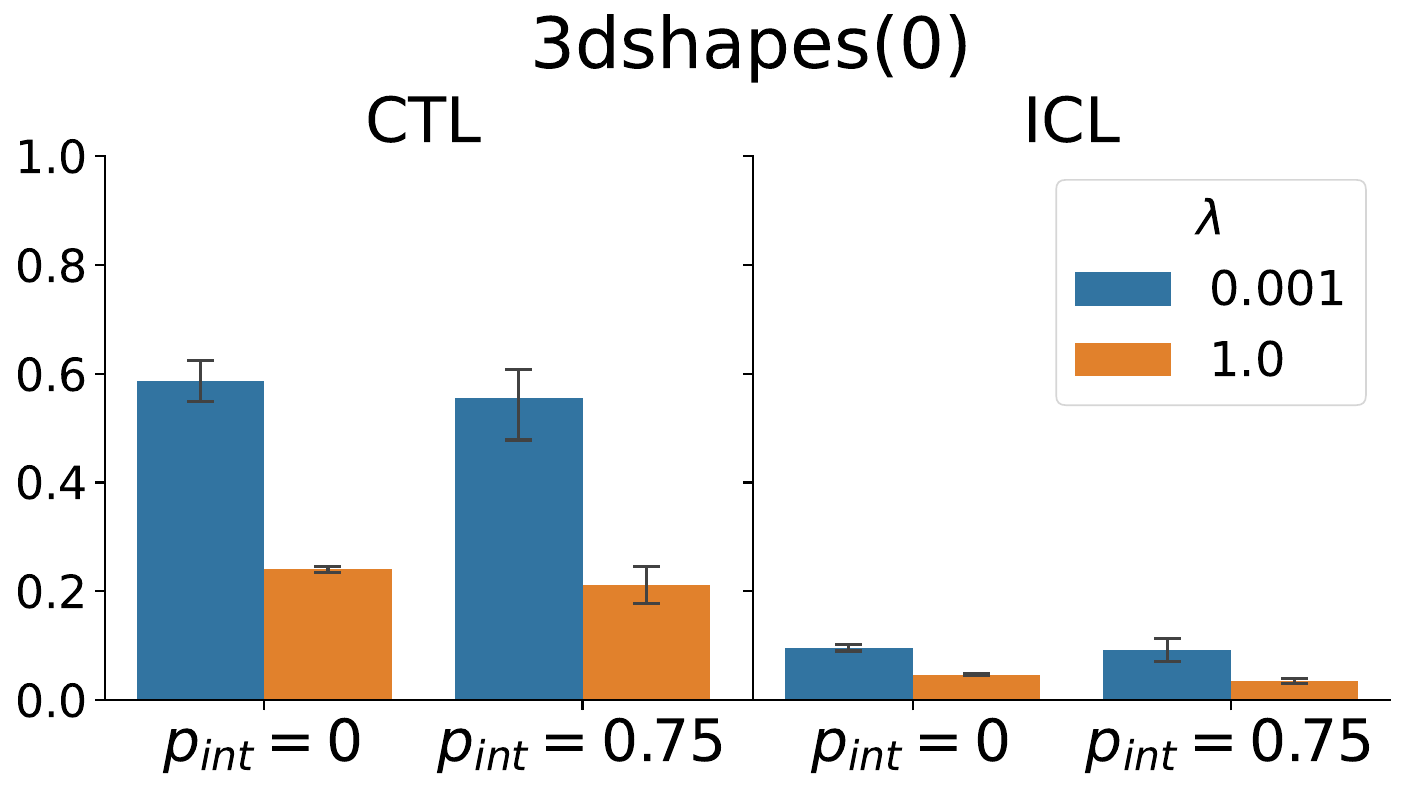} 
\end{minipage}
\begin{center}
    \begin{minipage}[c]{0.32\textwidth}
\centering
    \includegraphics[width=0.98\textwidth]{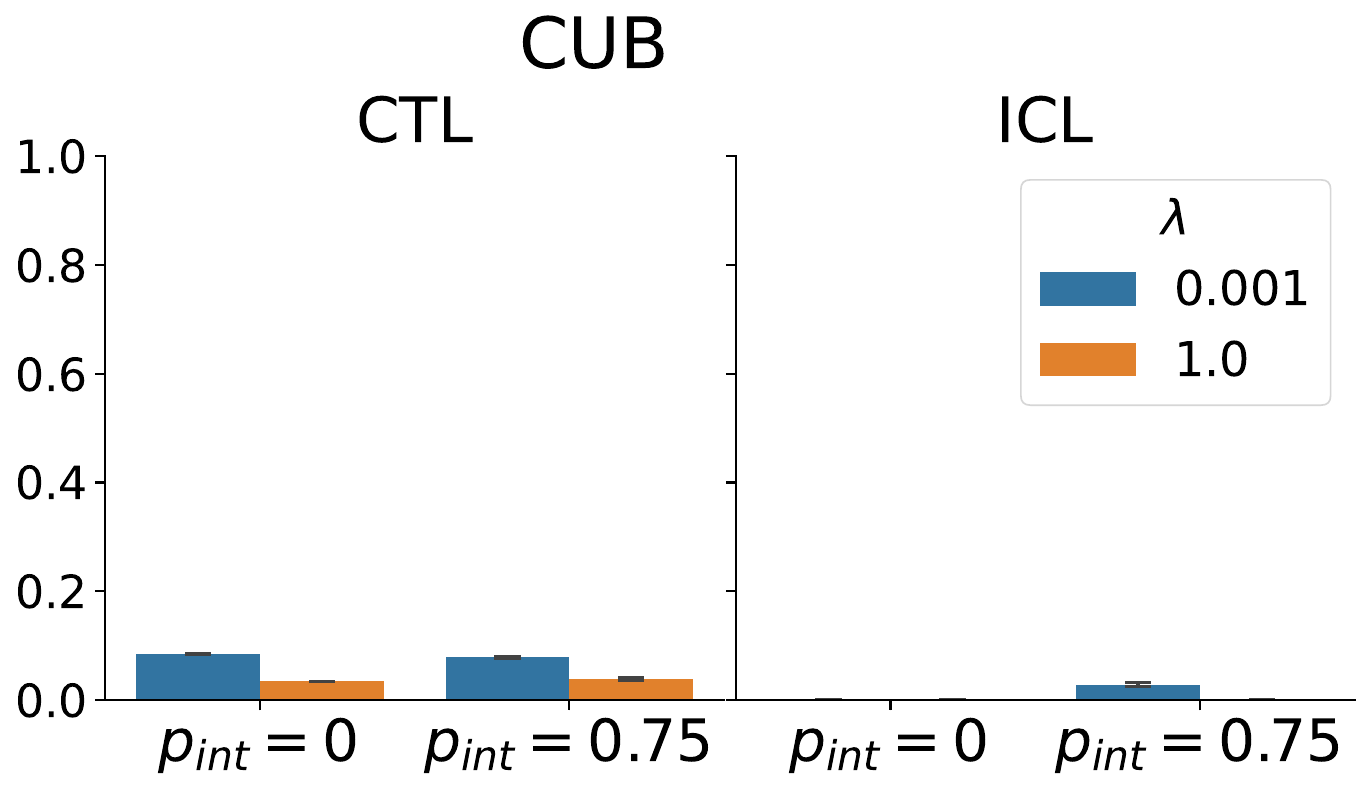} 
\end{minipage}
\begin{minipage}[c]{0.32\textwidth}
\centering
    \includegraphics[width=1\textwidth]{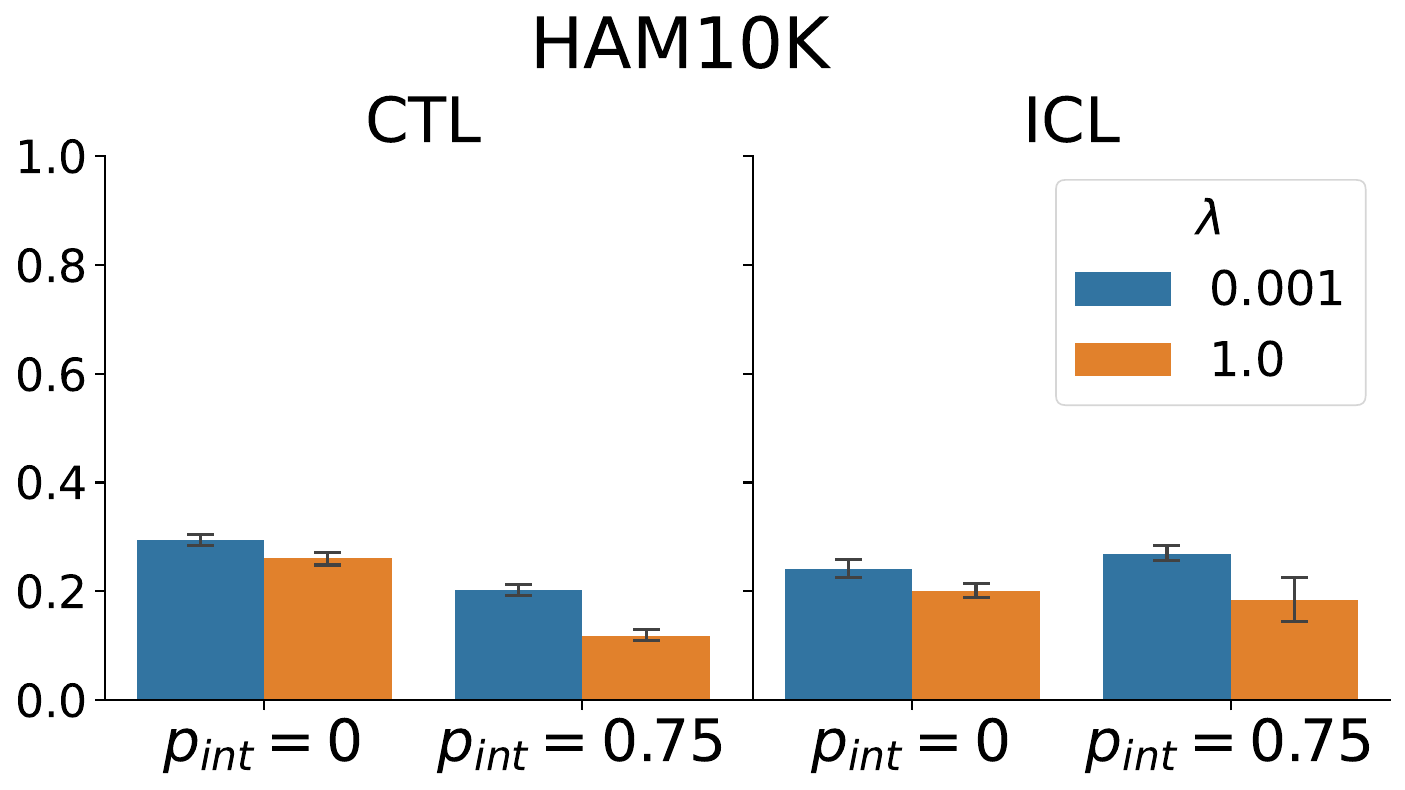} 
\end{minipage}
\end{center}
\caption{
CTL and ICL scores evaluated on the predicted concept probabilities $\hat{c}_i$ in CEMs with different values of $\lambda$ and $p_{int}$.
    }
    \label{figure_CEM_CTL_ICL}
\end{figure}

To gain deeper insight into interconcept leakage in CEMs, we evaluated the CTL and ICL scores on the predicted concept probabilities $\hat{c}_i$. Although not indicative of model interpretability -- since reasoning occurs at the level of the weighted vectors for CEMs -- it is worth noting that they decrease in a manner similar to CBMs as concept supervision increases. The picture that emerges is thus that as one raises $\lambda$ and $p_{int}$, the leaked information vanishes from the concept probabilities, and is effectively transferred into the vector representations as interconcept leakage.
The vector representations that are used in CEMs hence provide a mechanism for interconcept leakage to persist even at high concept supervision.

\paragraph{Alignment leakage in CEMs.} A subtle leakage effect may in principle arise when exposing a model like CEMs to interventions during training. For each data point $n \in \{1, \dots, N\}$ with ground-truth value $c_i^{(n)}$ for concept $i$, we label $\hat{c}^{+(n)}_i$ as aligned if $c_i^{(n)} = 1$, and unaligned otherwise, while we label $\hat{c}^{-(n)}_i$ as aligned if $c_i^{(n)} = 0$, and unaligned otherwise.
The meaning of this distinction is that after intervening on the $i$-th concept at training or test time, the weighted vector $\hat{c}^{w(n)}_i$ is equal to the aligned vector. 
A model exposed to interventions may thus learn to improve performance upon interventions by generating aligned vectors that are more predictive of the task than the unaligned ones. This form of leakage results in embeddings that have inhomogeneous reliability depending on the alignment with ground-truth concepts, leading to an overall reduction in model interpretability. 

This effect, which we label \emph{alignment leakage}, is a specific manifestation of concepts-task leakage. To quantify it,  we define the following information-theoretic score, which indicates the extent to which aligned vectors are more predictive of the task than unaligned vectors across both positive and negative embeddings, as an excess in the concepts-task normalised MIs defined in \eqref{eq_I_CT_CEM}. Specifically, we define:
\begin{align} \label{eq_I_align_CEM}
    \widetilde{I}^{(\text{align})}\left(\bm{\hat{c}^{+}}, \bm{\hat{c}^{-}}, c, y\right) &= 
    \widetilde{I}^{(CT)}\left(\hat{c}^{+\text{(aligned)}}, y\right) -
    \widetilde{I}^{(CT)}\left(\hat{c}^{+\text{(unaligned)}}, y\right) \\
    & \,+ 
    \widetilde{I}^{(CT)}\left(\hat{c}^{-\text{(aligned)}}, y\right) -
    \widetilde{I}^{(CT)}\left(\hat{c}^{-\text{(unaligned)}}, y\right)
    , \nonumber
\end{align}
where $\hat{c}^{\pm\text{(aligned)}}$ denote the set of positive/negative vectors aligned with the corresponding ground-truth concept, and analogously for $\hat{c}^{\pm\text{(unaligned)}}$. $\widetilde{I}^{(\text{align})}$ takes values in (-2,2) in units of normalised MI. Positive values indicate that the aligned vectors are more predictive of the task than the unaligned ones, while a value of zero corresponds to no alignment leakage.

\begin{figure}[t]
\centering
\begin{minipage}[c]{0.3\textwidth} 
\centering
\includegraphics[width=0.98\textwidth]{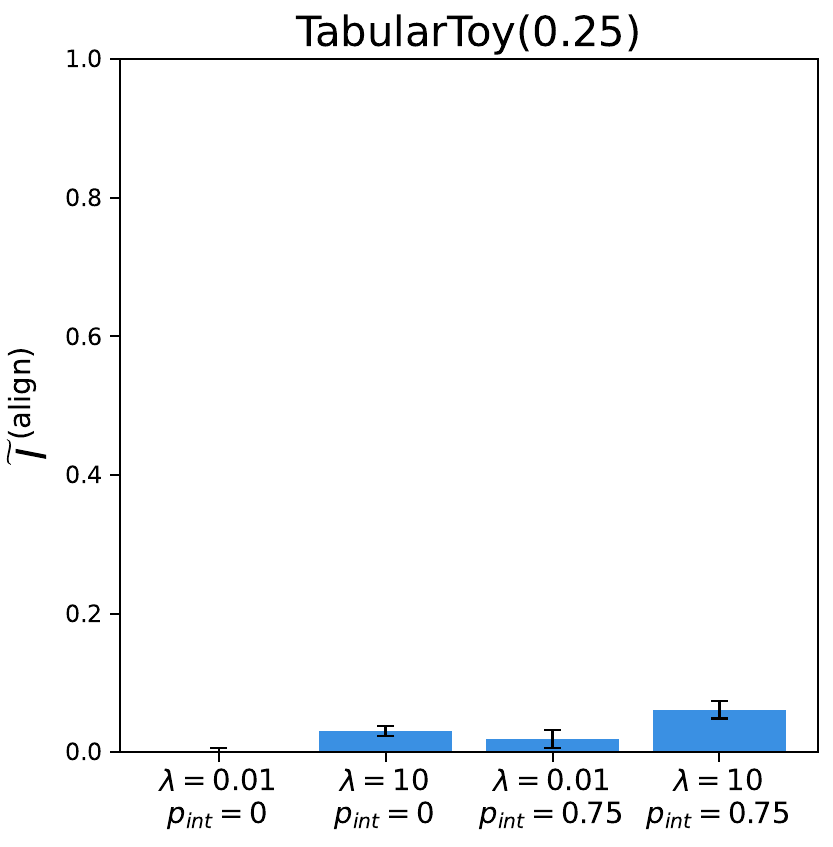} 
\end{minipage}
\begin{minipage}[c]{0.3\textwidth} 
\centering
\includegraphics[width=1\textwidth]{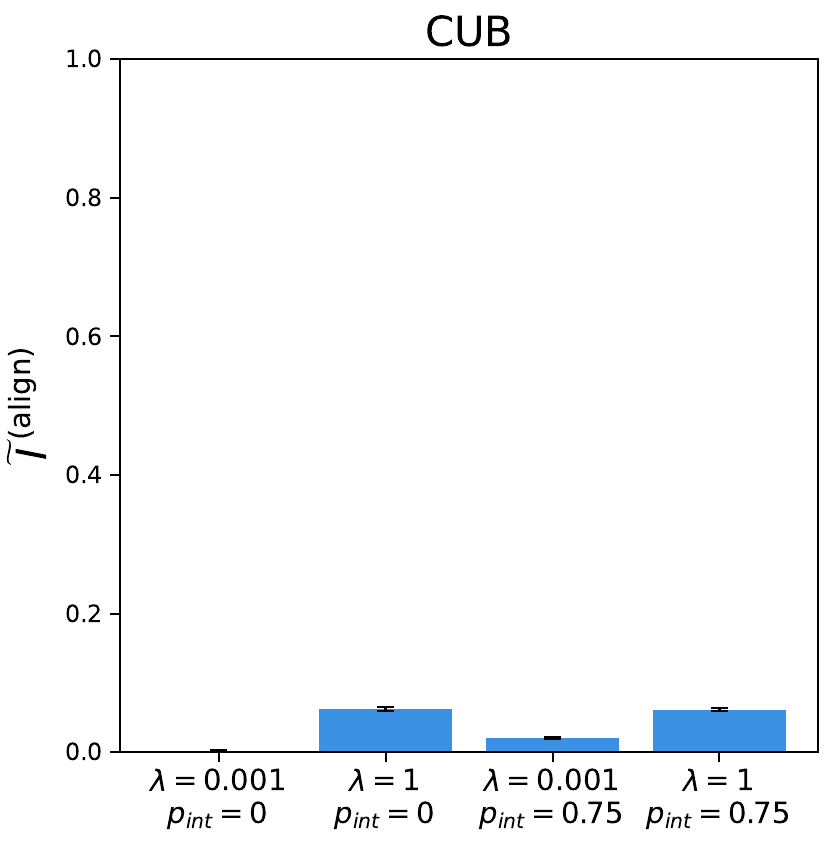} 
\end{minipage}
\caption{
Alignment leakage score for CEMs trained on the TabularToy(0.25) and CUB datasets with low and high values of $\lambda$ and $p_{int}$.
}
    \label{figure_TT_alignment_leakage}  
\end{figure}

We assessed this form of leakage in CEMs trained at different values of $\lambda$ and $p_{int}$. While in models trained on dSprites(0), 3dshapes(0) and HAM10K it does not appear, we find significantly non-zero scores for models trained on TabularToy(0.25) and CUB (Figure \ref{figure_TT_alignment_leakage}). Alignment leakage is stronger in models at high values of $\lambda$ and $p_{int}$, supporting the intuition that this effect is associated with exposure to interventions during training.

Mechanisms analogous to alignment leakage may help explain why CEMs trained with $p_{int}>0$ often perform well under interventions (see Appendix \ref{App_further_CEM_noninterpret} and \cite{CEMs}). This interpretation is also consistent with the observation that CEMs trained with $p_{int}=0$ are typically only mildly responsive to interventions (see Figure 6 and A.8 in \cite{CEMs}). Taken together, these findings suggest that the positive and negative embeddings may encode the concept state only weakly, or at least much less strongly than in CBMs, for example.

\section{Lessons for model design}
\label{sec_lessons}

Our framework and results on leakage enable us to outline the key steps for designing and training a concept-based model to ensure its interpretability.
The first step consists of assessing whether the annotated concepts are sufficiently predictive of the task. To that aim, one should train several classifier heads on the ground-truth concepts, starting from a linear baseline. 
Depending on the use-case, while the choice of an interpretable head (such as a linear classifier) may be desirable, it may also result in severe misspecification. In this case, one should expect the gain in functional interpretability to be compensated to a certain extent by higher leakage, as per the results of Section \ref{sec_causes_of_leakage}.

If a sufficiently high task accuracy is not achieved with arbitrarily complex architectures, this is generally an indication that the set of concepts is not complete for the task. This either causes a drop in task performance in the case of hard architectures, or introduces a significant amount of leakage in the case of soft architectures.
Furthermore, the experiments from Section \ref{sec_causes_of_leakage} suggest that higher concept supervision is not sufficient in general to prevent leakage in such cases. If additional annotated concepts are not available, one should thus conclude that the task is not amenable to a concept-based approach given the dataset at hand.

Possible solutions to concept set incompleteness have been studied. These include models that allow leakage into concept representations to encode the additional required information; CEMs and related architectures are one such example \citep{CEMs}. Another approach consists of adding unsupervised concepts or channels, meant to account for the missing information while in principle preserving the interpretability from the annotated concepts \citep[see][]{Mahinpei2021PromisesAP, Addressing_leakage, Sawada2022ConceptBM, yuksekgonul2023posthoc, Sheth23, ismail2024concept, ismail2025concept}.
However, when adopting this approach, one has to ensure the model does not over-rely on such unsupervised channels, effectively bypassing the concept bottleneck when making a task prediction, and resulting equivalent to an end-to-end model without concept supervision. Especially for high-risk scenarios, a general framework to interpret unknown concepts and safely include them in the concept refinement process is still missing.

Another approach to dealing with an incomplete (or missing) set of annotated concepts that has recently gained popularity is the use of CBMs with label-free concept representations \citep{yuksekgonul2023posthoc, oikarinen2023labelfree, LaBo}. A common modelling strategy in this line of work is to define task-relevant concepts using a large language model and encode them as concept prototypes with a vision-language model such as CLIP \citep{CLIP} The projection of an input image embedding onto these prototypes yields soft concept activations, on top of which a final head is trained sequentially to predict the task label. From the perspective of leakage, these models behave as standard sequential CBMs, which are known to exhibit leakage similar to jointly trained CBMs \citep{Mahinpei2021PromisesAP}. In particular, in these models the structure and geometry of the concept embeddings from the multimodal encoder may carry additional task-relevant information beyond the ground-truth interconcept and concepts-task relationships, which can then leak into the concept activations. Our framework could help clarify these effects using analyses similar to those applied to CEMs in Section \ref{sec_interpretability_CEMs}, addressing questions such as whether task performance is truly driven by the activation of the selected concepts, or rather by task-relevant information encoded in the concept and input representations (using the analogue of the $\widetilde{I}^{(CT)}$ score defined in \eqref{eq_I_CT_CEM}). Since little or no concept annotation is available in these setups, it would also be valuable to accurately benchmark leakage using our framework on datasets where annotated concepts are available and explicitly imposed as the bottleneck of an otherwise label-free model.

It is also worth noting that an incomplete concept set and a misspecified classifier head are in principle degenerate issues, both leading to a drop in task performance and similarly higher leakage scores.
Quantifying the completeness of a concept set for a given task is by itself an information-theoretic problem that has been extensively discussed in the context of feature selection, and some measures have been presented, for example in \cite{Completeness_concept_set, Addressing_leakage}, which may provide indications of ways to avoid such degeneracy.

If such issues do not arise, the next design choice concerns the form of concept encoding. The results of Section \ref{sec_causes_of_leakage} show that this step is critical: the encoding should not be overly expressive relative to the annotated concepts, otherwise leakage may be encouraged.
Logit encoding is over-expressive for binary concepts, while it is arguably a suitable choice for continuous concepts.
Similarly, vector encoding may be generally over-expressive for one-dimensional variables.

For binary annotated concepts, two common setups are soft and hard CBMs. To reduce leakage and achieve accurate concept learning, soft CBMs should be trained at relatively high $\lambda \in (\lambda_{min}, \lambda_{max})$, according to the results of Section \ref{sec_causes_of_leakage}. The CTL and ICL scores enable tuning $\lambda$ to values that minimise leakage. Throughout our experiments they also indicate that leakage is often unavoidable in soft CBMs. This may be acceptable depending on the use case, particularly since at high $\lambda$ the model's reliance on leaked information may be secondary to concept activations.
Note that when designing a concept-based model, it can also be beneficial to train an end-to-end model without concept supervision, in order to establish an upper bound on the overall task performance, and assess the intrinsic suitability of the input data for the task.

More broadly, when developing concept-based architectures beyond CBMs, our metrics can assess how different design choices promote leakage within the concept representations. Interconcept leakage may be significantly reduced by encouraging the concept encoder to learn the ground-truth interconcept correlation structure, for instance by adopting a probabilistic perspective as in Stochastic \citep{vandenhirtz2024stochastic, PosthocStochastic} and Autoregressive CBMs \citep{Mahinpei2021PromisesAP}. However, per se a probabilistic approach does not prevent concepts-task leakage: if the concept encoder $p(c|x)$ is learnt jointly or sequentially with the final head and no hard threshold is imposed on the sampled concepts, $p(c|x)$ and the sampled concepts will generally encode task-relevant information.

In summary we have proposed an information-theoretic framework to assessing the interpretability of concept-based models. The leakage measures we introduce are robust and sensitive, and we propose their use as standard tools for ensuring interpretability when designing concept-based models.

\acks{We thank Aya Abdelsalam Ismail, Anthony Baptista and Gregory Verghese for insightful discussions on related topics. All authors except CH are supported by the Turing-Roche partnership. EP is funded by the Department of Science Innovation and Technology under PharosAI.
CRSB is supported by a King's College London AI+ Fellowship. TC is supported by a principal research fellowship from University College London. CH is an employee of Roche.} 

\appendix
\section{Details on the experimental setup}
\label{App_experiments}

\subsection{Datasets}
\label{App_experiments_Datasets}

For additional details on these datasets, we refer the reader to \cite{Espinosa_Zarlenga_2023}.

\paragraph{TabularToy($\delta$).} This dataset is constructed by sampling a 3-dimensional latent variable $z \sim \mathcal{N}(0, \Sigma(\delta))$, with correlation matrix  $\Sigma(\delta)_{ij} = \delta_{ij} + \delta (1-\delta_{ij})$ with $i, j=1, 2, 3$, where $\delta_{ij}$ is the Kronecker symbol and $\delta \in (-1,1)$ indicating the correlation between components. The 7-dimensional input $x$ is a trigonometric function of the latent variables \citep[see][]{Mahinpei2021PromisesAP}, while the three binary concepts are defined as the sign of the latent variables, $c_i = (z_i > 0)$. The binary task label is the following linear function of concepts,
\begin{equation} \label{eq_TT_original_task}
    y_{\text{TT}}^{(orig)} = (c_1 + c_2 + c_3 \geq 2)\,.
\end{equation}
For our experiments we generate a dataset of size 10,000, with a 0.7/0.2/0.1 train / validation / test ratio.
\textbf{For Figures \ref{figure_TT_leakage_concept_distributions} and \ref{figure_TT_intervention_2conceptsmodels_random}}, we consider a simplified version of TabularToy($\delta$) with only the first two concepts $c_1$ and $c_2$ for each input, and binary task ${y = (c_1 + c_2 \geq 1)}$.

\paragraph{dSprites($\gamma$).} For this family of datasets, each input $x \in \{0,1\}^{64\times64\times1}$ is an image generated deterministically from five latent variables 
$z = (\text{shape} \in \{0,1,2\},$ 
$\text{scale} \in \{0,\dots, 5\},$ 
$\theta \in \{0, \dots, 39\},$ 
$X \in \{0,\dots, 31\},$ $Y \in \{0,\dots, 31\}),$ 
where shape corresponds to either a square, ellipse or a heart, while
$\theta$, $X$ and $Y$ indicate the rotation and the position of the shape in the plane respectively.
Following \cite{Espinosa_Zarlenga_2023}, we define dSprites($0$) as the datasets with no interconcept correlations. Here the five binary concepts are based on the value of the latent variables,
$c = ( z_1 < 2,$ 
$z_2 < 3 ,$ 
$z_3 < 20 ,$ 
$z_4 < 16,$ $z_5 < 16)$,
and the 8-class task label is 
\begin{equation}\label{eq_dS_original_task}
    y_{\text{dS}}^{(orig)} =  (1-c_1) (2 c_4 + c_5 + 4)    
    + c_1   (2 c_2 + c_3)\,.
\end{equation}
To generate dSprites($\gamma$) with increasing interconcept correlations for $\gamma = 1, \dots, 4$, we adopt the procedure detailed in \cite{Espinosa_Zarlenga_2023}. We use datasets with approximatively 25K and 14K samples for dSprites(0) and dSprites(4) respectively, with a 0.7/0.1/0.2 train / validation / test ratio.

\paragraph{3dshapes($\gamma$).} Each input is an image $x \in \{0,1\}^{64\times64\times3}$ generated from six latent variables
$z = (\text{floor\_hue} \in \{0,\dots,9\},$ 
$\text{wall\_hue} \in \{0,\dots, 9\},$ 
$\text{object\_hue} \in \{0,\dots, 9\},$ 
$\text{scale} \in \{0,\dots, 7\},$ 
$\text{shape} \in \{0, 1, 2, 3\},$ 
$\text{orientation} \in \{0,\dots, 14\}),$ where shape denotes either a sphere, a cube, a capsule or a cylinder. Following \cite{Espinosa_Zarlenga_2023}, in 3dshapes($0$) concepts are defined as 
$c = ( z_1 < 5,$ 
$z_2 < 5 ,$ 
$z_3 < 5 ,$ 
$z_4 < 4,$ $z_5 < 2,$ $z_6 < 7),$ and the 12-class labels are obtained as
\begin{equation}\label{eq_3ds_original_task}
    y_{\text{3ds}}^{(orig)} =  (1-c_5) (2 c_1 + c_2)    
    + c_5 (4 c_3 + 2 c_4 + c_6 + 4)
\end{equation}
Increasing interconcept correlations are induced for $\gamma = 1, \dots, 5$
in a similar fashion to dSprites($\gamma$). For our experiments we consider 3dshapes(0) and 3dshapes(5) with size 16K and 14K respectively, and a 0.7/0.1/0.2 train / validation / test ratio.

\paragraph{CUB.} This dataset consists of 11,788 images of birds labelled with 112 binary phenotype attributes, with the objective of predicting the correct species among 200 classes \citep{CUB}. It provides rich fine-grained concept annotations that collectively offer a near-complete description of the target classification task. We preprocessed it following \cite{koh20a}, and consider a balanced 0.4/0.1/0.5 train / validation / test ratio. 

\paragraph{HAM10K.}  In this dataset the objective is to determine whether a skin lesion is benign or malignant, using dermatoscopic images annotated with 18 morphological attributes. We use the annotations from \cite{Chanda2024DermatologistLikeXAI} and the preprocessing from \cite{Patricio2025TwoStepConceptSkin}, yielding 6498 images in total, split with a
0.8/0.1/0.1 train / validation / test ratio.

\begin{figure}[t]
\begin{minipage}[c]{0.325\textwidth} 
\centering
\includegraphics[width=0.8\textwidth]{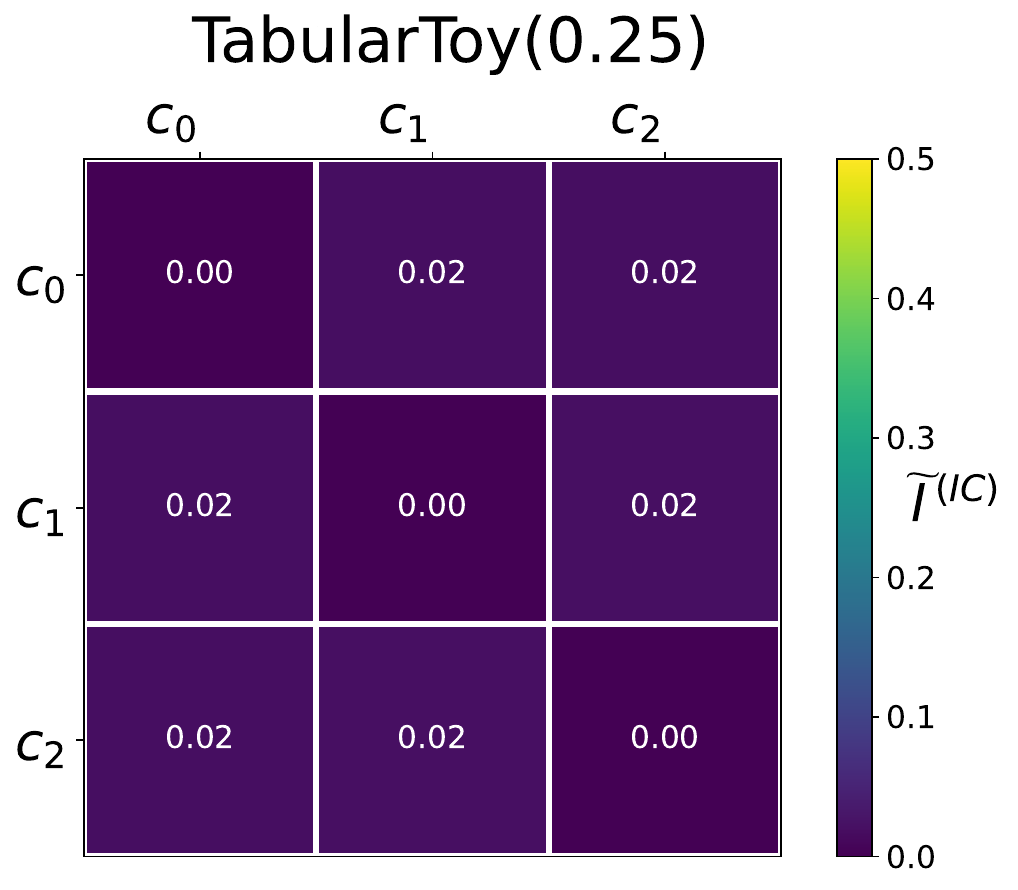} 
\end{minipage}
\begin{minipage}[c]{0.325\textwidth} 
\centering
\includegraphics[width=0.8\textwidth]{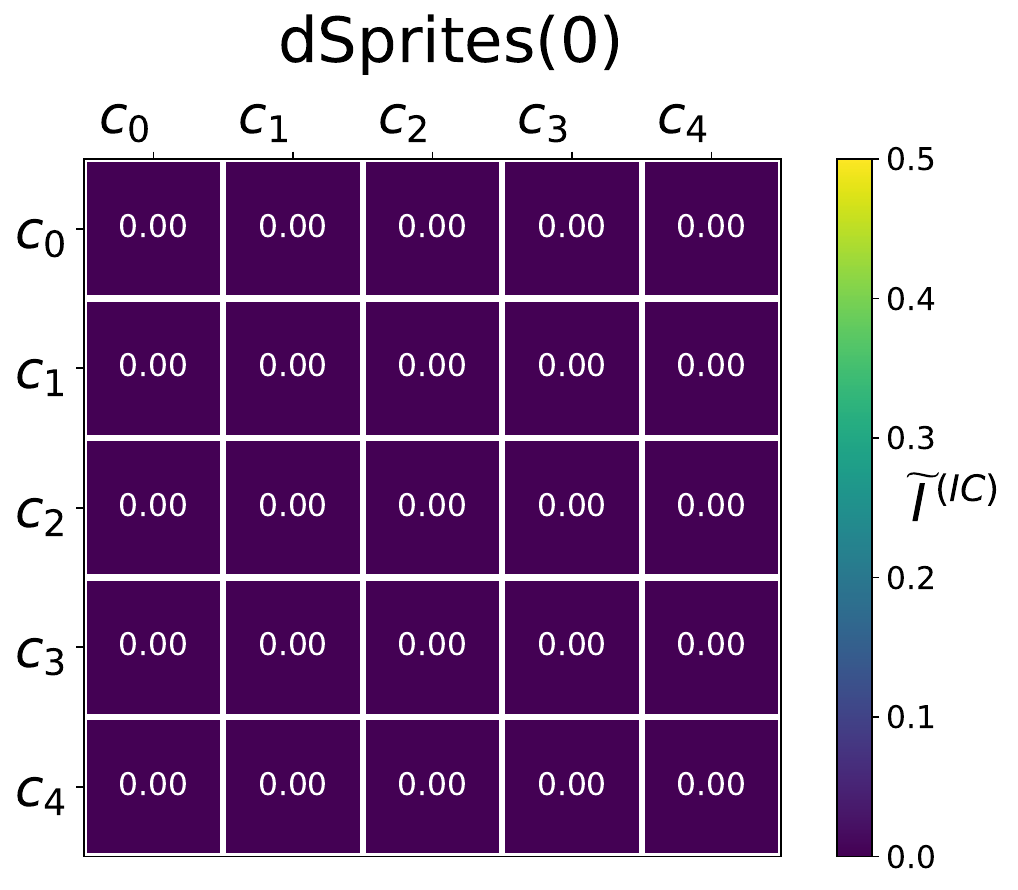} 
\end{minipage}
\begin{minipage}[c]{0.325\textwidth} 
\centering
\includegraphics[width=0.8\textwidth]{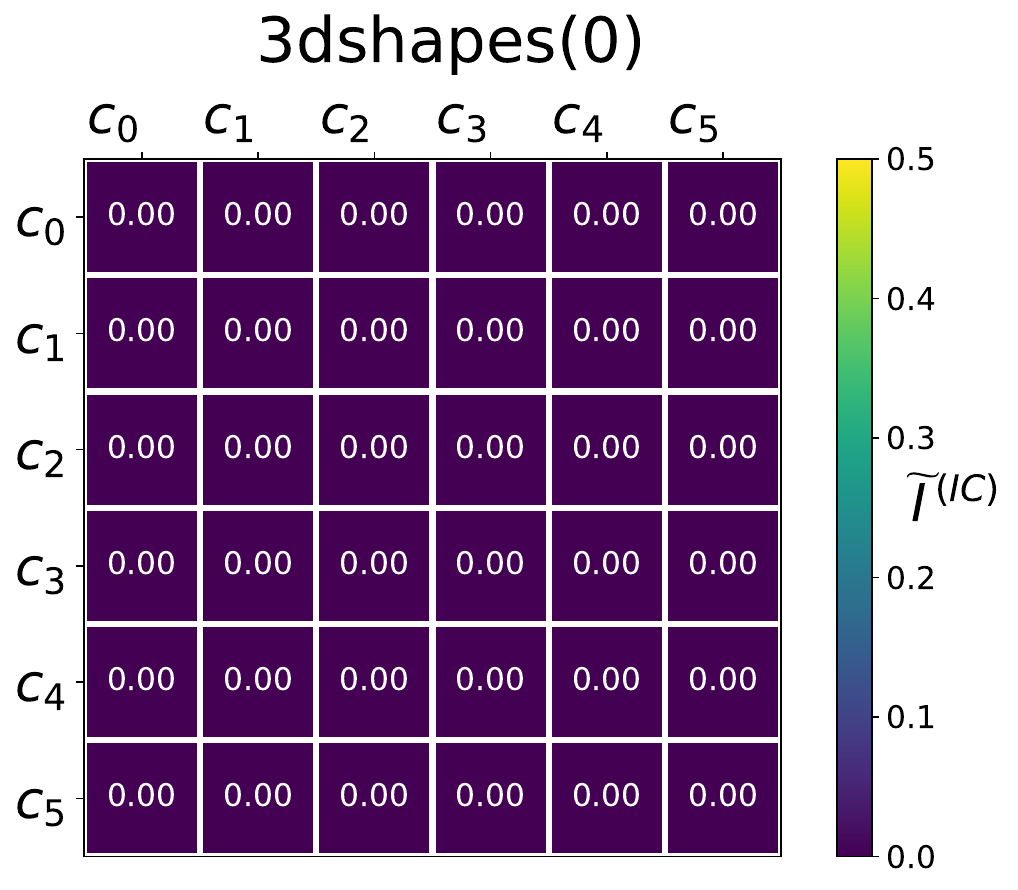} 
\end{minipage}
\begin{minipage}[c]{0.325\textwidth} 
\centering
\includegraphics[width=0.8\textwidth]{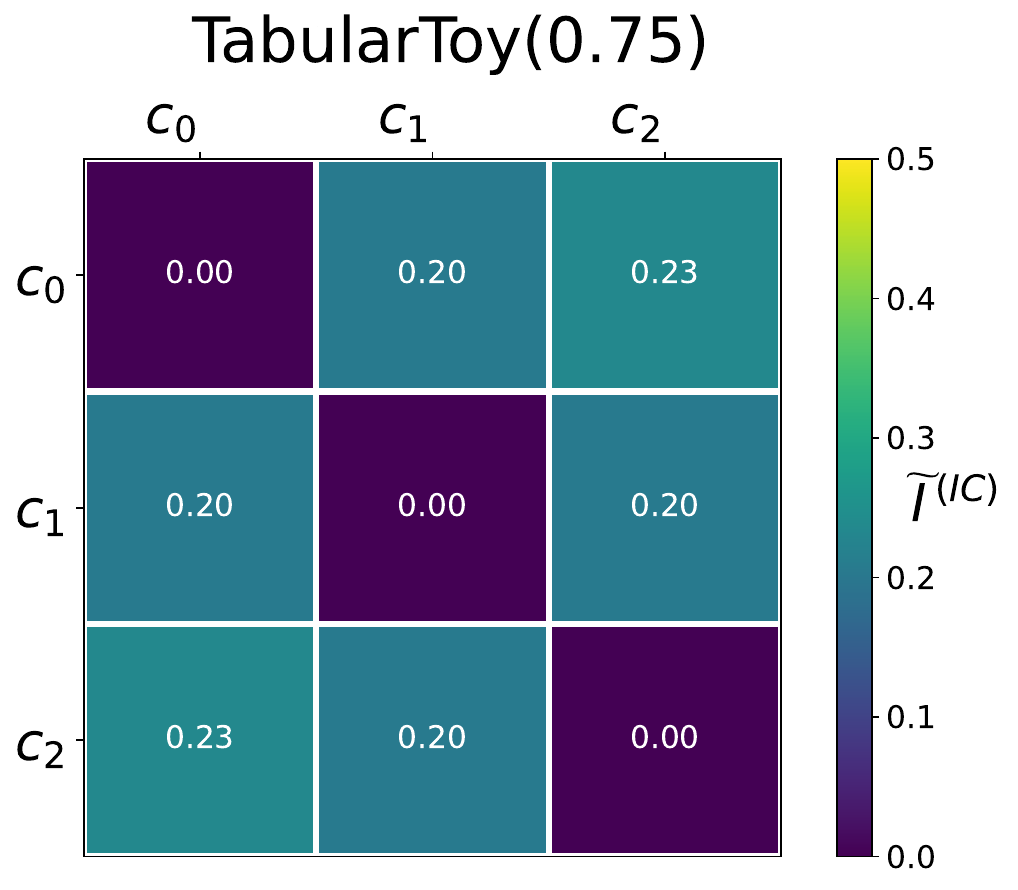} 
\end{minipage}
\begin{minipage}[c]{0.325\textwidth} 
\centering
\hspace{0.01cm} \includegraphics[width=0.8\textwidth]{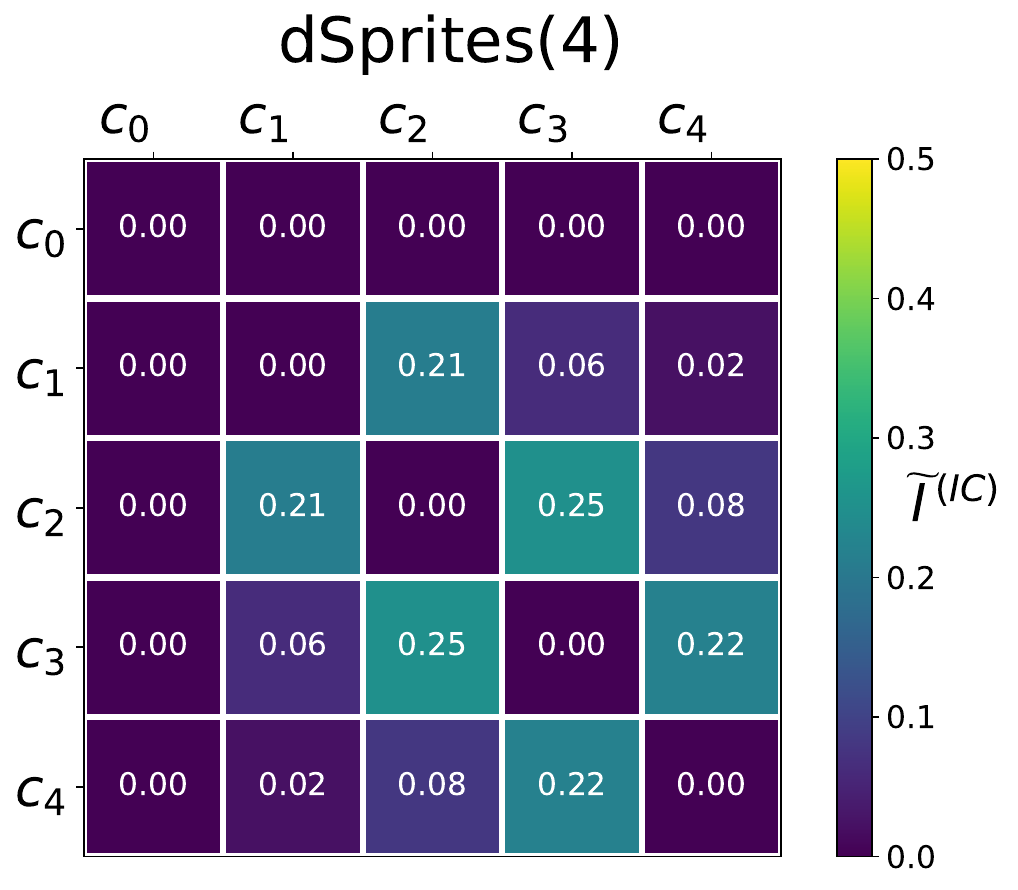} 
\end{minipage}
\begin{minipage}[c]{0.325\textwidth} 
\centering
\hspace{0.1cm} \includegraphics[width=0.8\textwidth]{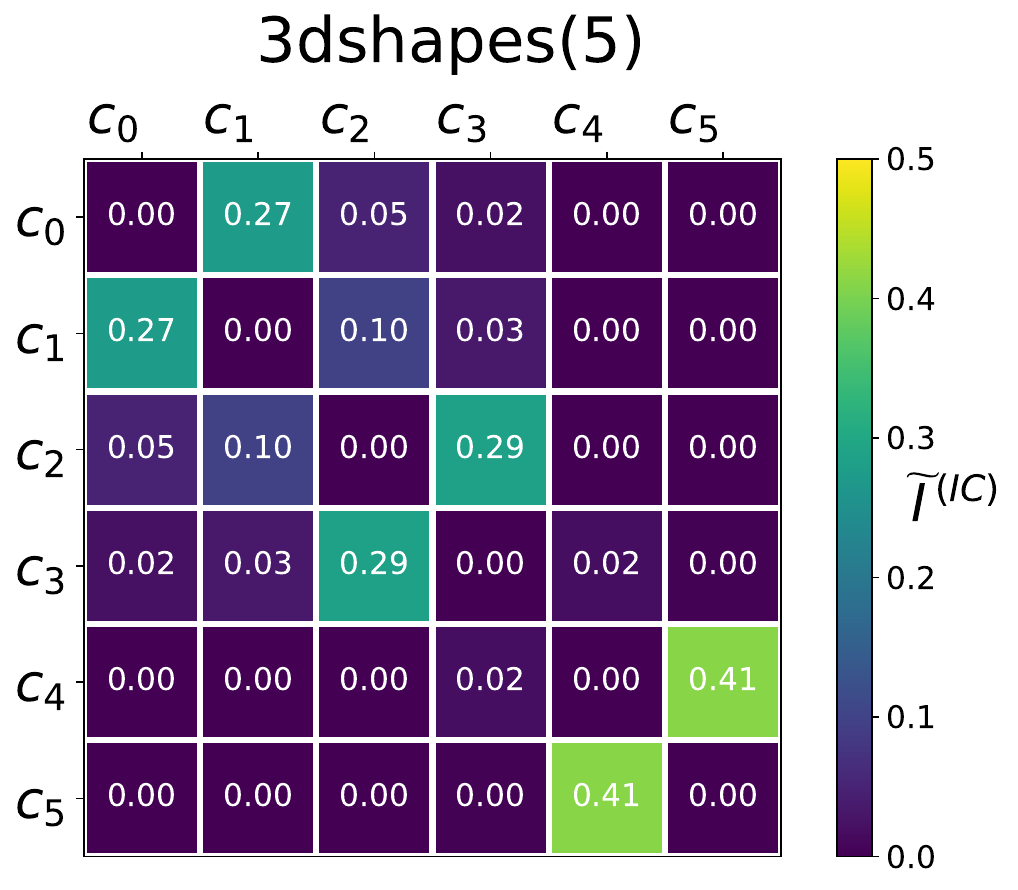} 
\end{minipage}
\caption{
Ground-truth interconcept normalised MI in the synthetic datasets.
}
    \label{figure_groundtruth_interconcept_nMIs}
\end{figure}

\paragraph{Ground-truth interconcept correlations.}
In Figure \ref{figure_groundtruth_interconcept_nMIs} we display as a reference the ground-truth interconcept normalised MI for the synthetic training datasets used in the experiments.

\paragraph{Experiments with an incomplete concept set.}
In TabularToy(0.25), we removed $c_3$ from the three concepts. In dSprites(0), we removed $c_4$ and $c_5$ from the five concepts. In 3dshapes(0), we removed $c_3$ and $c_6$ from the six concepts. For CUB we use its concept-incomplete version defined in \cite{zarlenga2025avoiding} where only 22 out of the 112 annotated concepts are kept. In HAM10K we remove 8 out of the 18 concepts, with the resulting concept set corresponding to the annotated features
\texttt{ ESA, PRL, PES, PIF, OPC, SPC, MVP, PRLC, PLF, PDES}.

\paragraph{Experiments with a misspecified head.}
\label{App_details_c2y_misspecification}
We provide here more details on the experiments to assess the effects of a misspecified final head discussed in Section \ref{sec_causes_of_leakage}. To generate non-linear concepts-task dependences in TabularToy($\delta$), we added a non-linear term to the original linear task in \eqref{eq_TT_original_task}, resulting in the binary classification problem,
$$
    y_{\text{TT}}^{(new)} = (c_1 + c_2 + c_3 - c_1 c_2 \geq 2)\,.
$$
The original tasks in \eqref{eq_dS_original_task} and \eqref{eq_3ds_original_task} for dSprites($\gamma$) and 3dshapes($\gamma$) are non-linear, however a linear classifier is able to reach essentially perfect task accuracy (Table \ref{table_c2y_task_performance_incomplete_concepts_misspecified}). To induce a sizeable misspecification, we thus added further higher-order non-linear terms. The tasks implemented for our experiments on dSprites(0) and 3dshapes(0) are:
\begin{align*}
    y_{\text{dS}}^{(new)} &= y_{\text{dS}}^{(orig)} - (1 - c_1)  c_3  c_4 
    - c_1  ( c_2  c_5 + c_3  c_5 + c_2  c_4)
    \,,\\
    y_{\text{3ds}}^{(new)} &= y_{\text{3ds}}^{(orig)}
    - 3  c_1 c_2  c_3 
    - c_4  c_5  c_6 
    - c_1  c_3  c_5 
    + c_2  c_4  c_6
    \,.
\end{align*}
We cannot run a similar experiment on CUB and HAM10K, since a linear final head already achieves near-perfect task performance.

\subsection{Model architectures and training} 

On TabularToy($\delta$) we used as concept encoder a 4-layer leaky-ReLU MLP with activations $\{ 7, 64, 64, 3\}$, and a linear classifier head. On dSprites($\gamma$), 3dshapes($\gamma$) and CUB we used a ResNet-18 encoder, while for HAM10K a ViT-B/16 encoder pretrained on ImageNet. For these datasets we used a 4-layer ReLU MLP with activations $\{ k, 64, 64, \ell\}$ as a final head. We trained CEMs with 16-dimensional embeddings.

We trained all models for 200 epochs, using the Adam optimiser with momentum 0.9 and learning rate $10^{-3}$ for TabularToy($\delta$), dSprites($\gamma$) and 3dshapes($\gamma$); $10^{-2}$ for CUB; $10^{-4}$ for HAM10K. For hard models, we trained the encoder and the classifier head for 200 and 20 epochs respectively. We trained with batch sizes 512 for TabularToy($\delta$); 32 for dSprites($\gamma$), 3dshapes($\gamma$) and HAM10K; 64 for CUB.

\subsection{Score evaluations}
\label{App_experiments_Scores_evaluation}

Except for Figure \ref{figure_groundtruth_interconcept_nMIs}, all the displayed scores are evaluated on test sets. The reported intervention performances were obtained using a random policy: for each data point, the next concept to intervene on was selected uniformly at random from the set of concepts that had not yet been intervened on.
In Figures \ref{figure_TT_intervention_2conceptsmodels_random}, \ref{figure_leakage_scores_single_models} and \ref{figure_NIS_single_models} we show results for single models, and the displayed means and 95\% confidence intervals refer to a 5-fold evaluation of each score, including intervention performance. In the rest of the paper, we repeated the evaluation of each score 5 times for individual models, and we display means and 95\% confidence intervals representing the distribution of the mean upon 5-fold training for each model class. 
MIs and entropies were estimated via the KSG estimator \citep{KSG03} with 3 nearest neighbours. The OIS and NIS were evaluated using their original implementations and default settings from \cite{Espinosa_Zarlenga_2023}.

\section{Undesired features of the Niche Impurity Score}
\label{App_NIS}

\begin{wrapfigure}{r}{0.35\textwidth}
\includegraphics[width=0.9\linewidth]{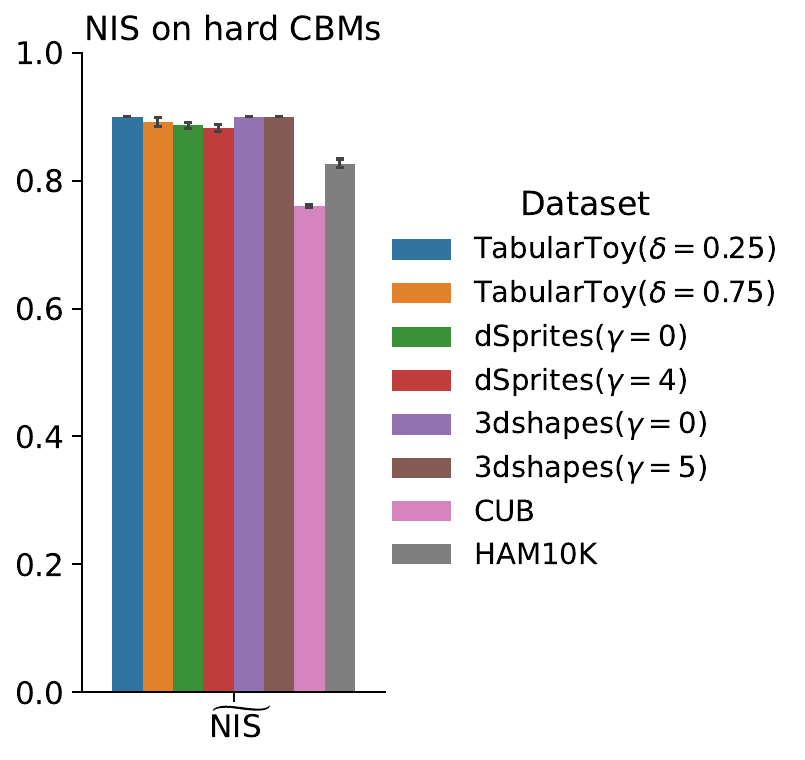} 
\caption{$\widetilde{\text{NIS}}$ evaluated on hard CBMs across datasets.}
\label{figure_NIS_on_hard_CBMs}
\end{wrapfigure}

Because it is computed as an AUC, NIS = 1/2 indicates that concept representation subsets are not predictive of the value of concepts, while NIS = 1 indicates maximal higher-order interconcept leakage. For a fair comparison of our scores with NIS, we plotted $\widetilde{\text{NIS}} = 2(\text{NIS}-1/2)$, so that scores of 0 and 1 correspond to no and maximal leakage respectively.

\begin{figure}[t]
\begin{minipage}[c]{0.32\textwidth} 
\centering \small \sffamily
$\quad \;$ TabularToy($\delta = 0.25$)
\vspace{0.2cm}
\end{minipage}
\hfill
\begin{minipage}[c]{0.32\textwidth} 
\centering \small \sffamily
$\quad$ dSprites($\gamma = 0$)
\vspace{0.2cm}
\end{minipage}
\begin{minipage}[c]{0.32\textwidth}
\centering \small \sffamily
$\quad \;$ 3dshapes($\gamma = 0$)
\vspace{0.2cm}
\end{minipage}
\begin{minipage}[c]{0.32\textwidth} 
\centering
\includegraphics[width=0.8\textwidth]{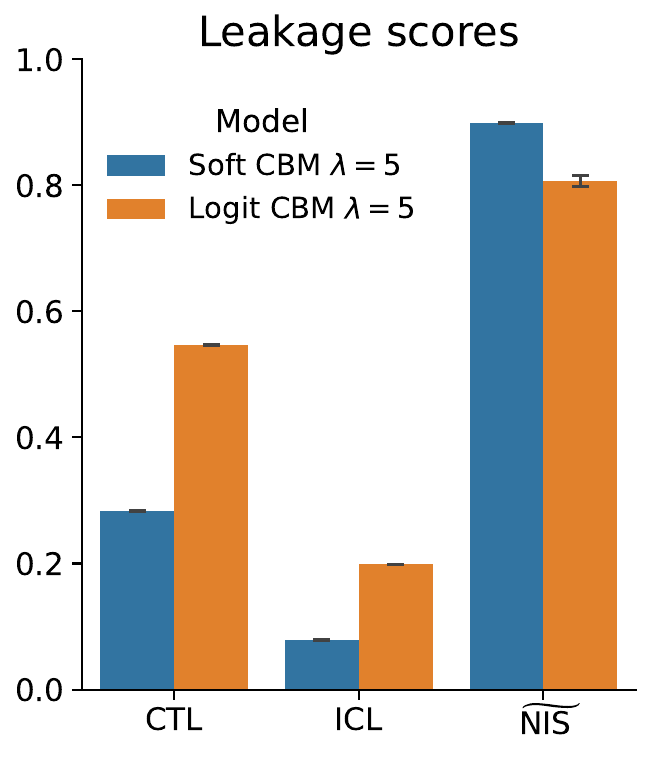} 
\end{minipage}
\hfill
\begin{minipage}[c]{0.32\textwidth}
\centering
    \includegraphics[width=0.8\textwidth]{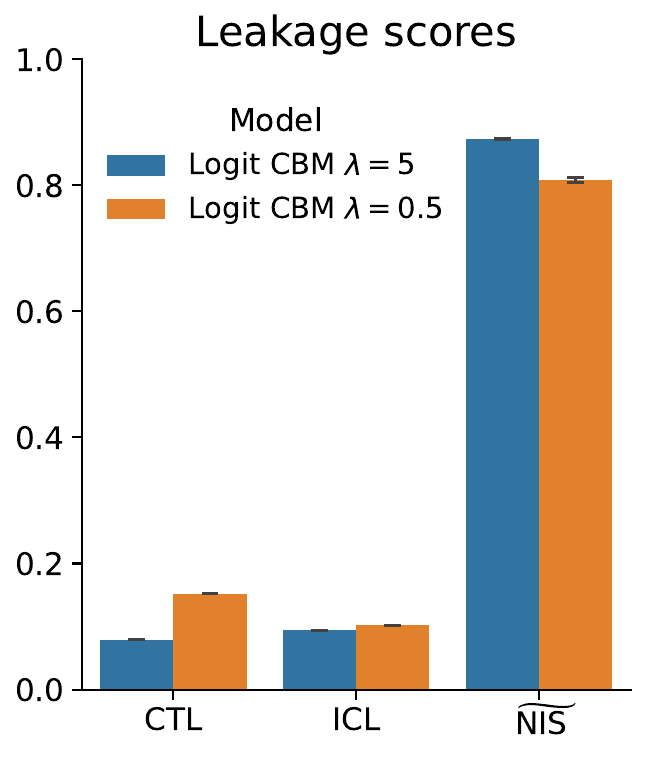} 
\end{minipage}
\begin{minipage}[c]{0.32\textwidth} 
\centering
    \includegraphics[width=0.8\textwidth]{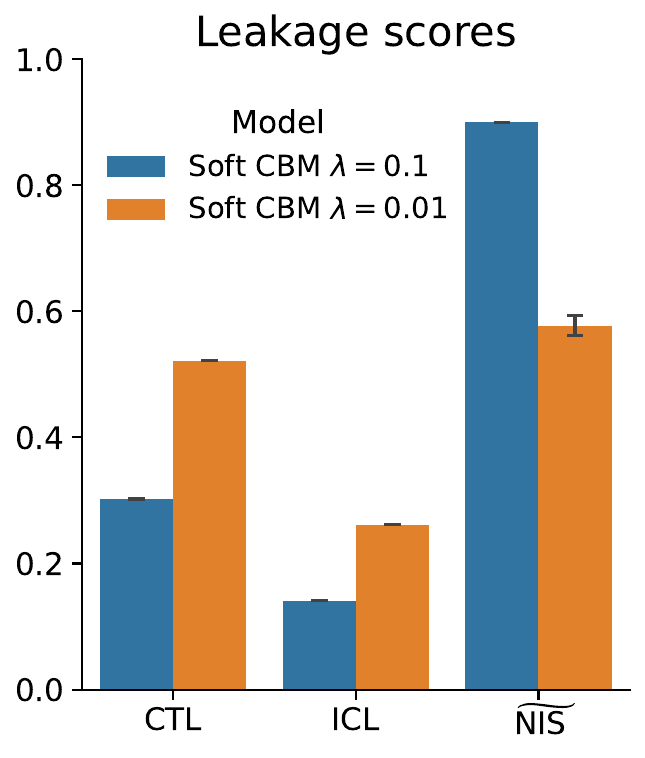} 
\end{minipage}
    \caption{
    $\widetilde{\text{NIS}}$ and leakage scores for the pairs of models assessed in Figure \ref{figure_leakage_scores_single_models} and Table \ref{table_evaluation_scores_single_models}.
    }
    \label{figure_NIS_single_models}
\end{figure}

Evaluating NIS on the pairs of models considered in Figure \ref{figure_leakage_scores_single_models}, this score appears to be anticorrelated with intervention performance and leakage (Figure \ref{figure_NIS_single_models}). Furthermore, it takes proportionally high values regardless of the model and dataset. This is confirmed by its behaviour on hard CBMs (Figure \ref{figure_NIS_on_hard_CBMs}): the NIS is non-vanishing and high in all cases. These undesirable features indicate that NIS is not a suitable measure of leakage.

\section{Concept-wise leakage scores}
\label{App_concept_wise_leakage_scores}

\begin{figure}[t]
\begin{minipage}[c]{0.325\textwidth} 
\centering
\includegraphics[width=\textwidth]{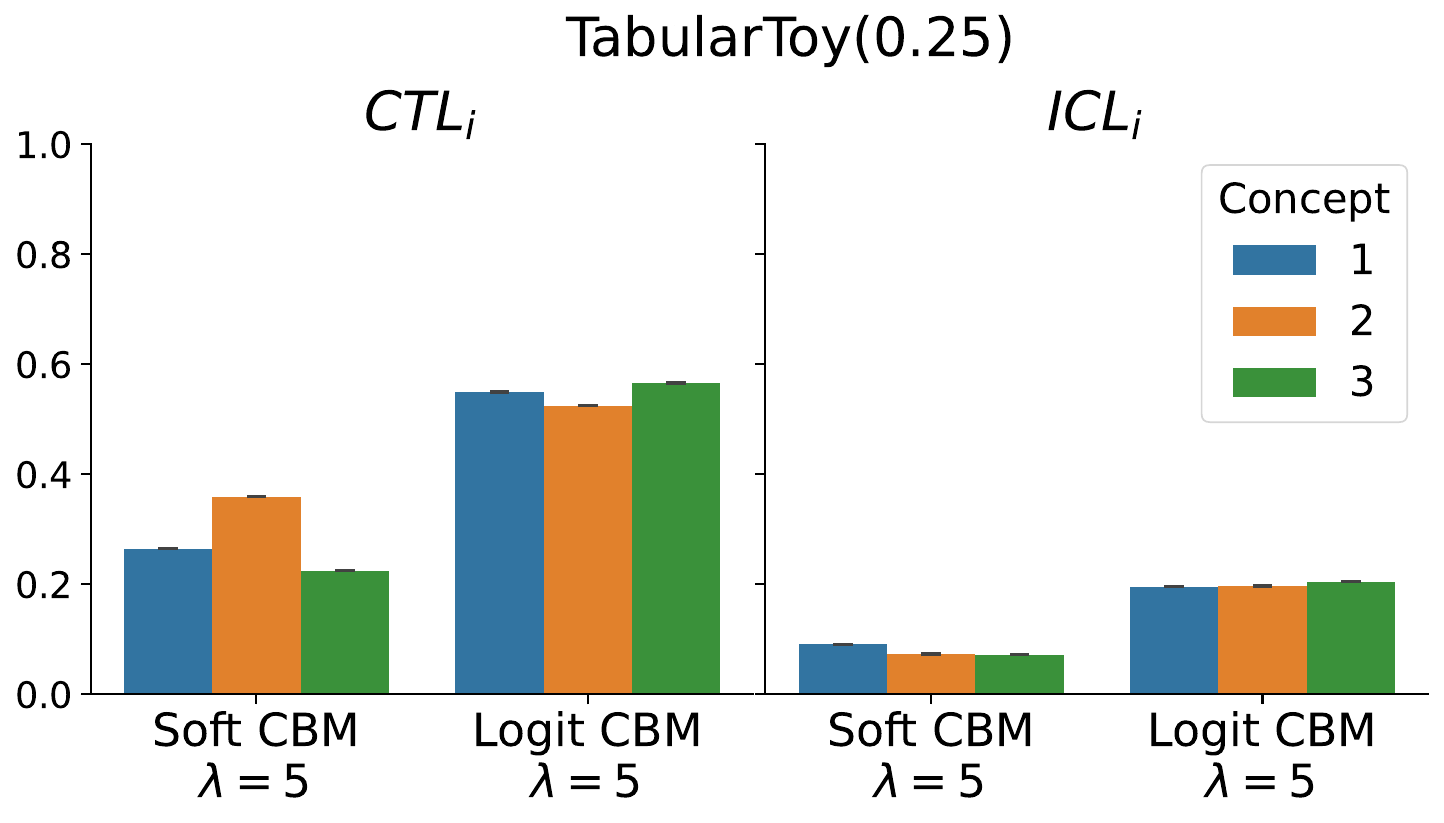} 
\end{minipage}
\begin{minipage}[c]{0.325\textwidth} 
\centering
\includegraphics[width=\textwidth]{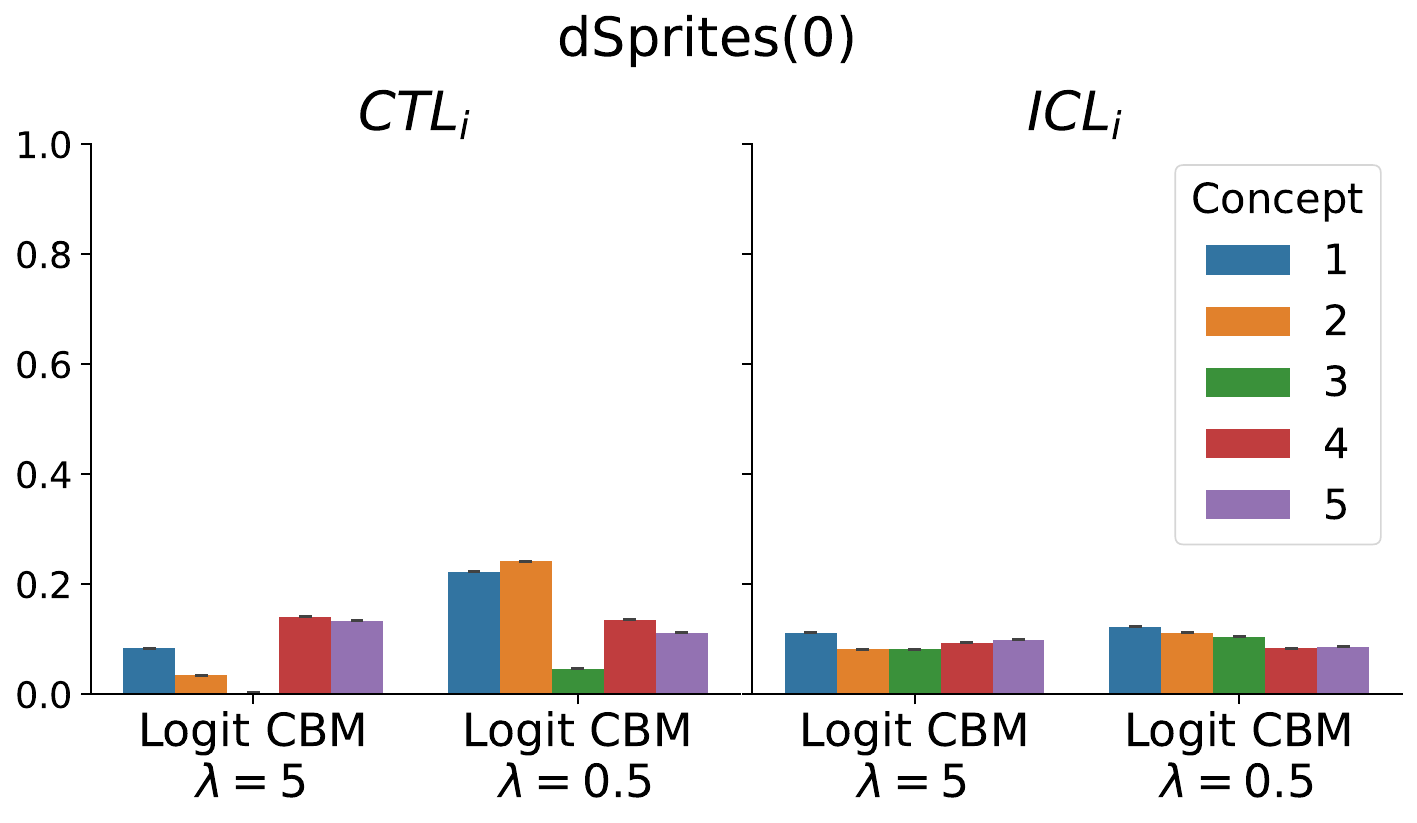} 
\end{minipage}
\begin{minipage}[c]{0.325\textwidth} 
\centering
\includegraphics[width=\textwidth]{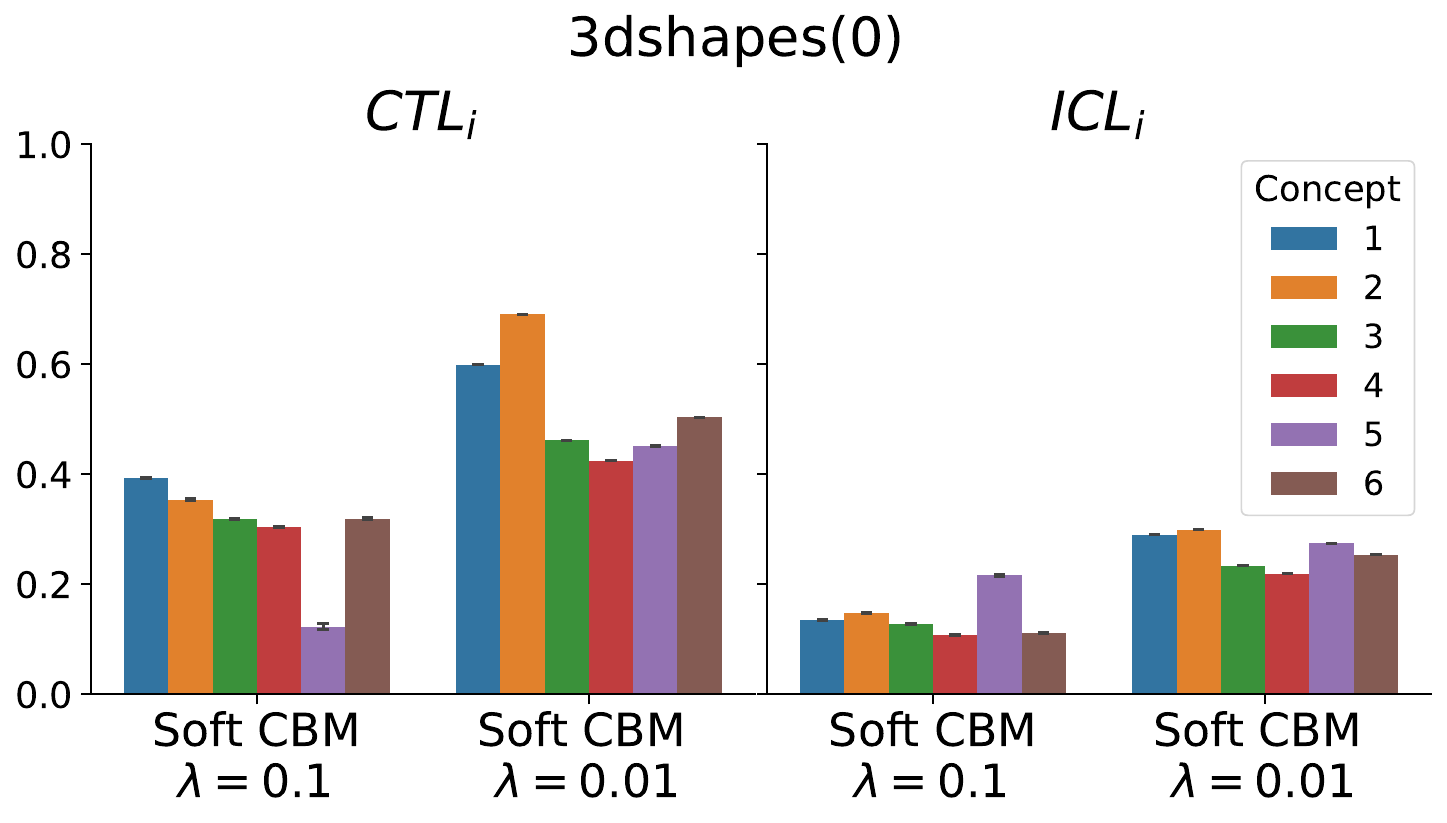} 
\end{minipage}
\caption{
Concept-wise leakage scores evaluated on the pairs of models in Figure \ref{figure_leakage_scores_single_models}.
    }
    \label{figure_concept_wise_leakage_scores}
\end{figure}

The concept-wise leakage scores defined in \eqref{eq_CTL_i} and \eqref{eq_ICL_i_ICL} provide more granular information on the leakage encoded in each concept. In Figure \ref{figure_concept_wise_leakage_scores} we display the $CTL_i$ and $ICL_i$ scores for the pairs of models analysed in Figure \ref{figure_leakage_scores_single_models} and Table \ref{table_evaluation_scores_single_models}. Note in particular that (i) leakage is not necessarily learnt homogeneously across concepts, and (ii) comparing two models with different levels of interpretability, certain concepts may exhibit comparable amounts of leakage (such as $c_4$ in the dSprites(0) example), while for others leakage may be significantly different.
Concept-wise scores are sensitive to both phenomena, making them valuable indicators of per-concept leakage risk upon intervention and therefore useful measures for model design.

\section{Details of the correlation between leakage scores and intervention performance}
\label{App_correlation_leakage_scores_int}

The models considered for this computation were
\begin{itemize}
    \item for the TabularToy(0.25), TabularToy(0.75), dSprites(0), dSprites(4) datasets: soft and logit CBMs with $\lambda$ = 0.01, 0.1, 0.5, 1, 5, 10;
    \item for the 3dshapes(0), 3dshapes(5) datasets: soft and logit CBMs with $\lambda$ = 0.001, 0.005, 0.01, 0.05, 0.1, 0.5, 1.
    \item for the CUB, HAM10K datasets: soft and logit CBMs with $\lambda$ = 0.001, 0.005, 0.01, 0.05, 0.1, 0.5, 1, 5, 10.
\end{itemize}
Each class of models was trained over 5 folds, amounting to a total of 60 evaluated models for each of the first set of four datasets, and to a total of 70 models for each of the second set of two datasets. We use Rubin's rule \citep[see][]{rubin2004multiple, BarnardRubin99} to combine within-imputation and between-imputation variances
and obtain the best-estimate Pearson $r$ and $p$-value.

\section{Behaviour of CBMs at high $\lambda$}
\label{App_high_lambda}

\begin{figure}[t]
\begin{minipage}[c]{0.325\textwidth} 
\centering
\includegraphics[width=\textwidth]{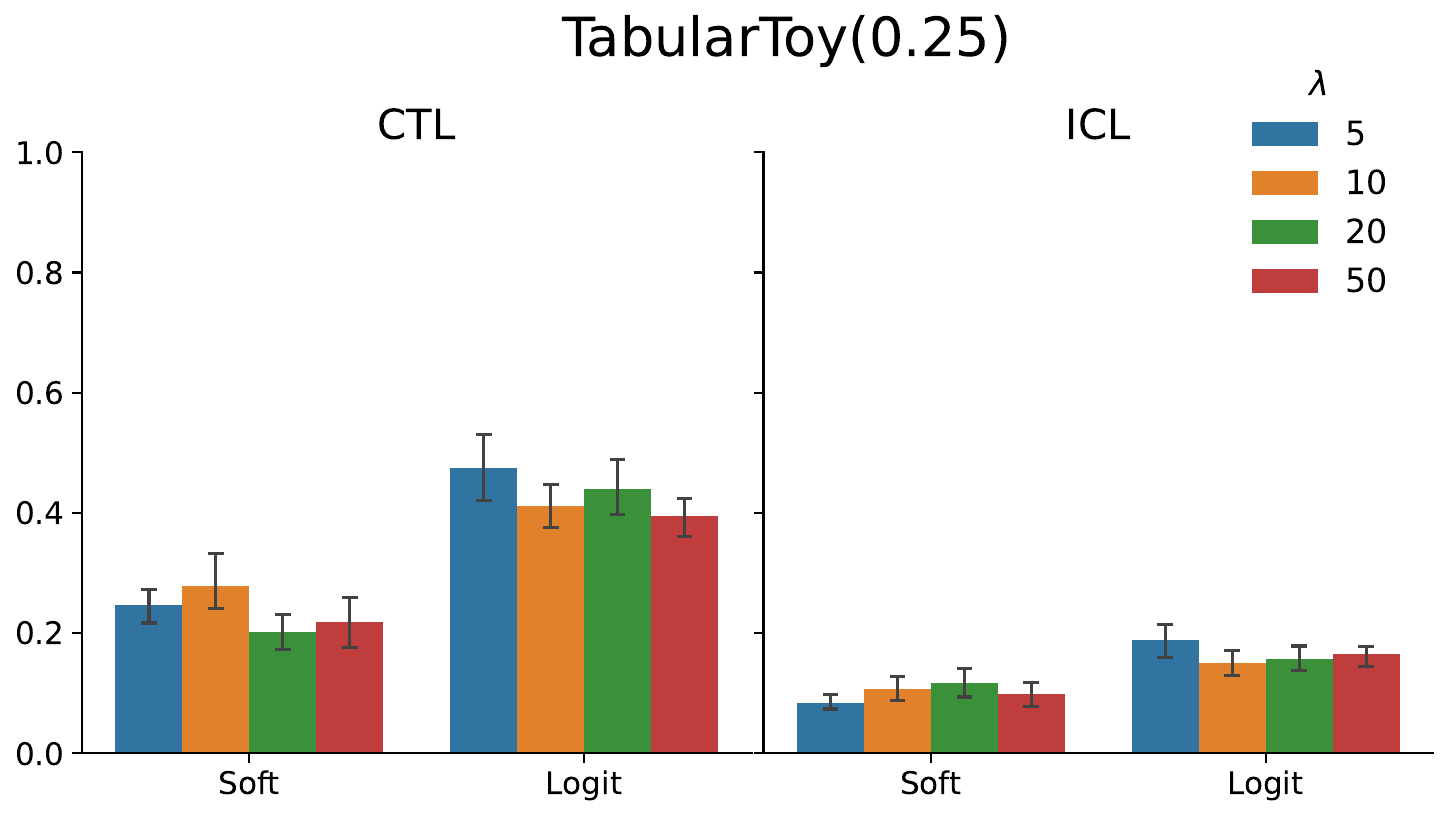} 
\end{minipage}
\begin{minipage}[c]{0.325\textwidth} 
\centering
\includegraphics[width=\textwidth]{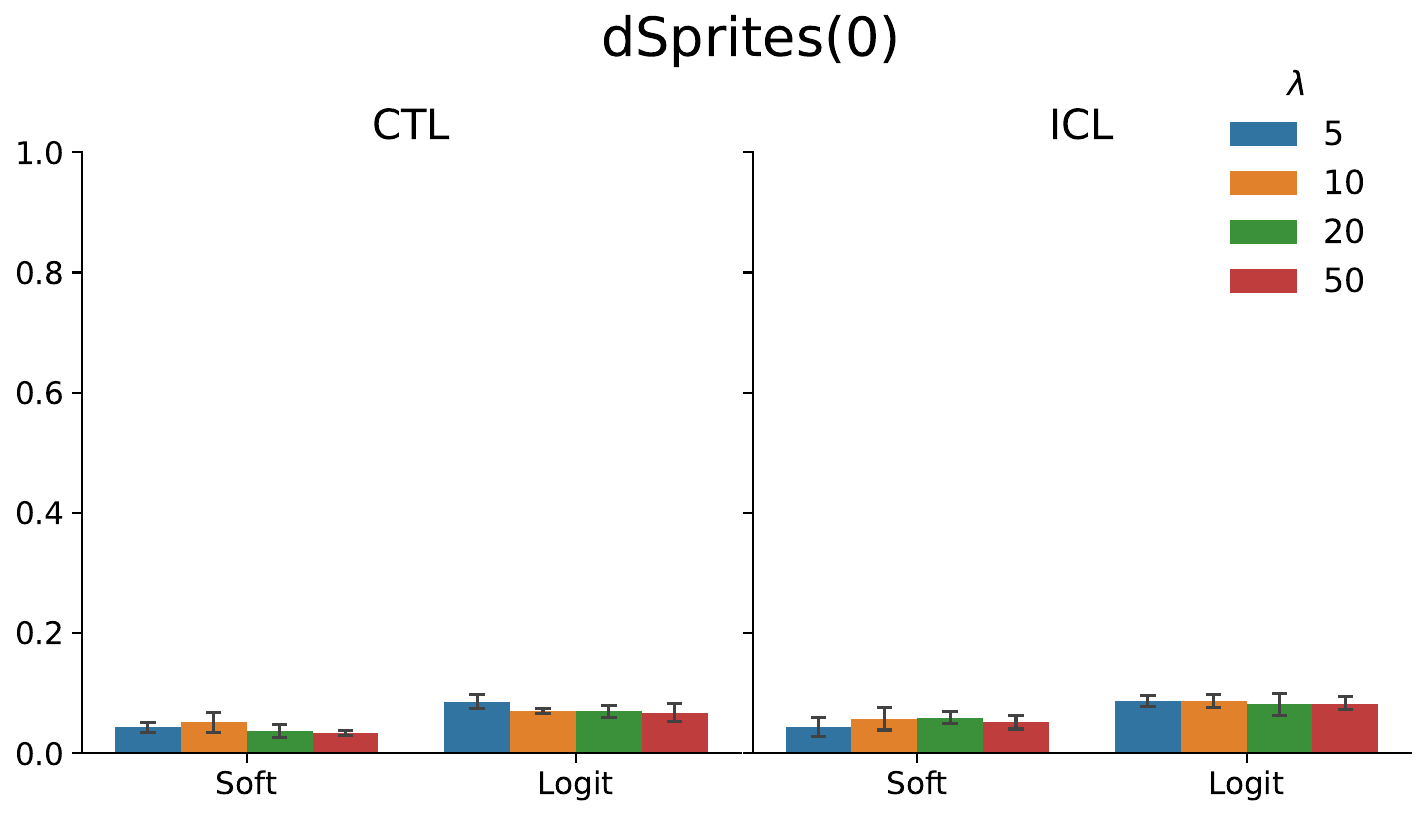} 
\end{minipage}
\begin{minipage}[c]{0.325\textwidth} 
\centering
\includegraphics[width=\textwidth]{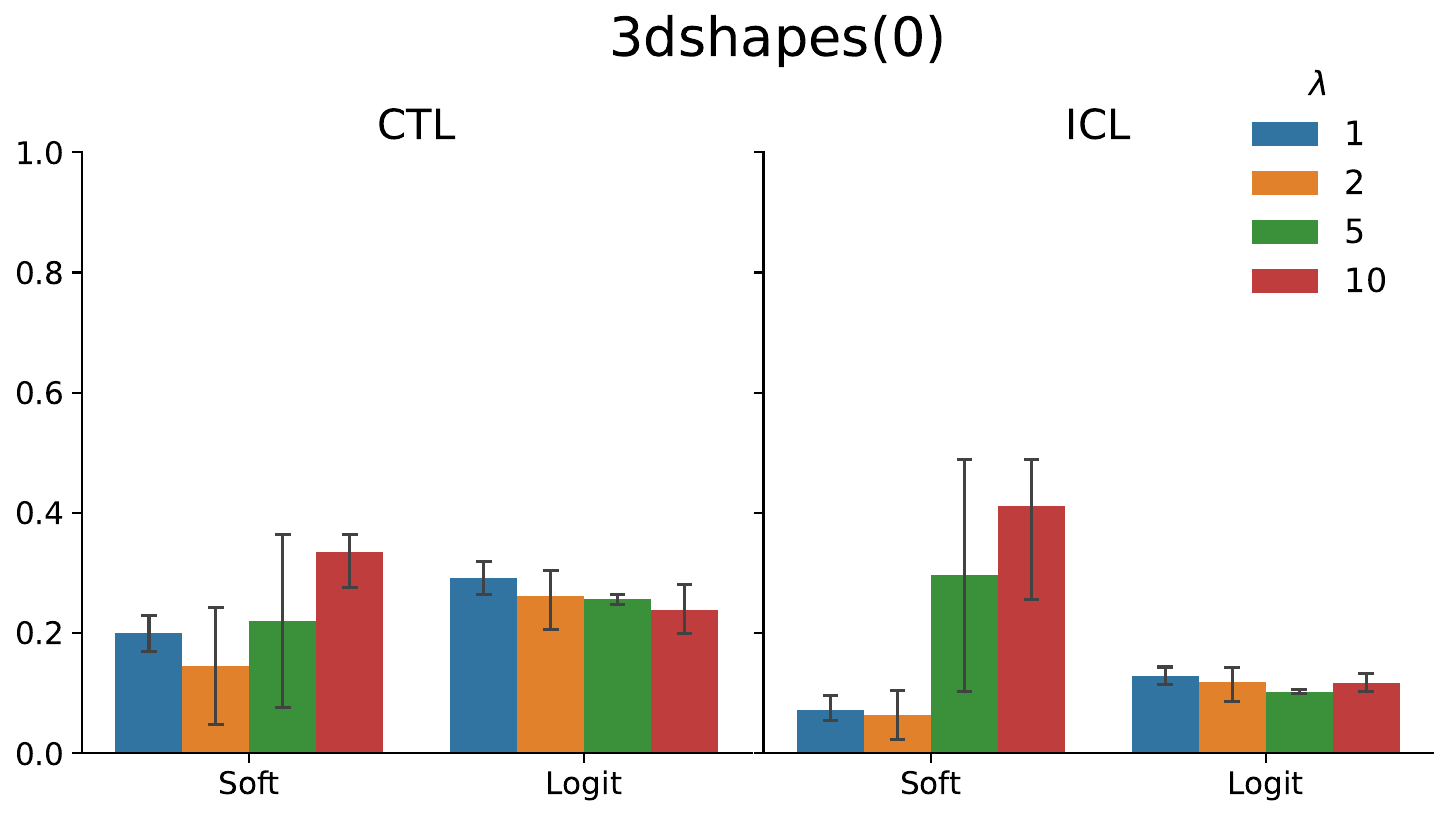} 
\end{minipage}

\begin{center}
\begin{minipage}[c]{0.325\textwidth} 
\centering
\includegraphics[width=\textwidth]{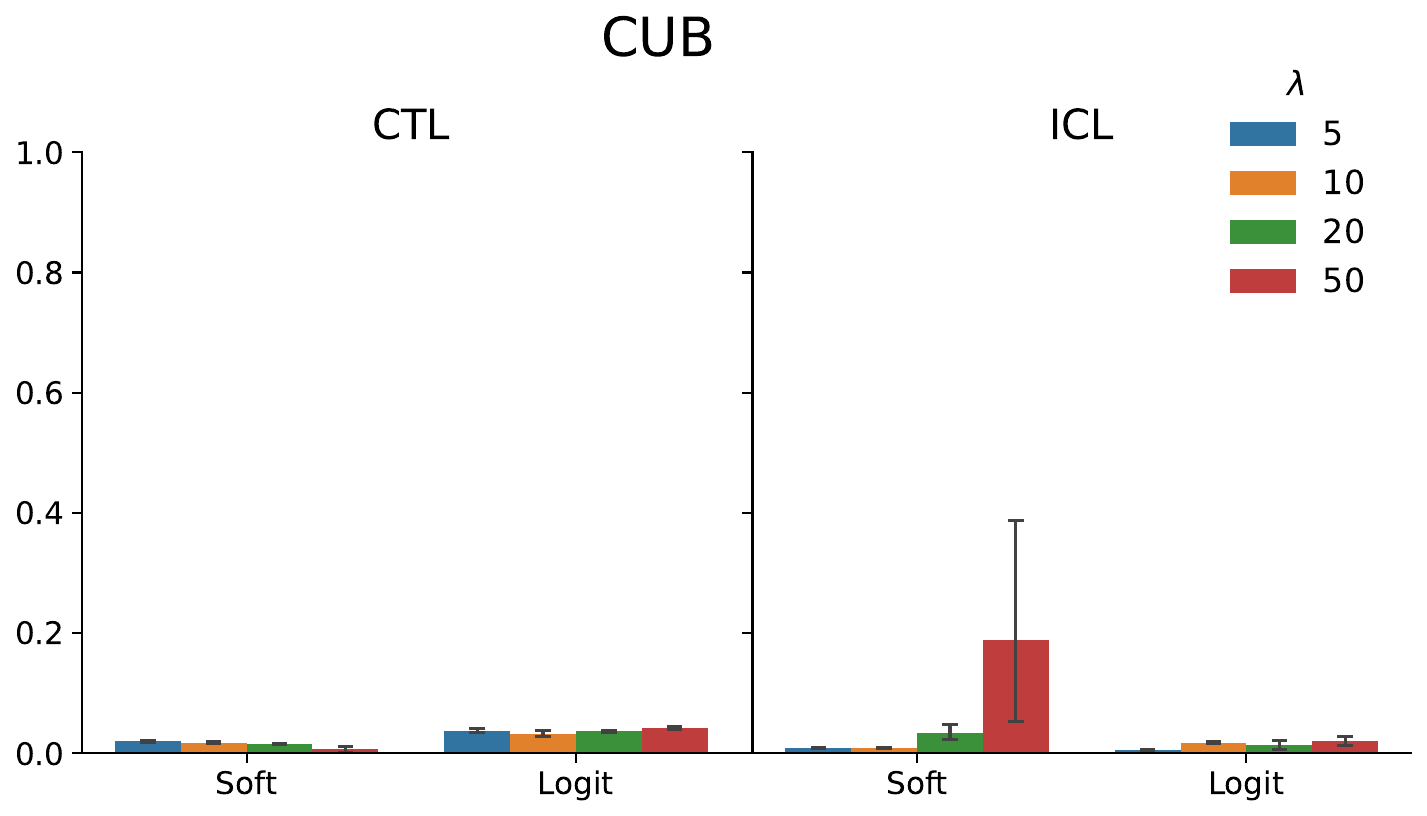} 
\end{minipage}
\begin{minipage}[c]{0.325\textwidth} 
\centering
\includegraphics[width=\textwidth]{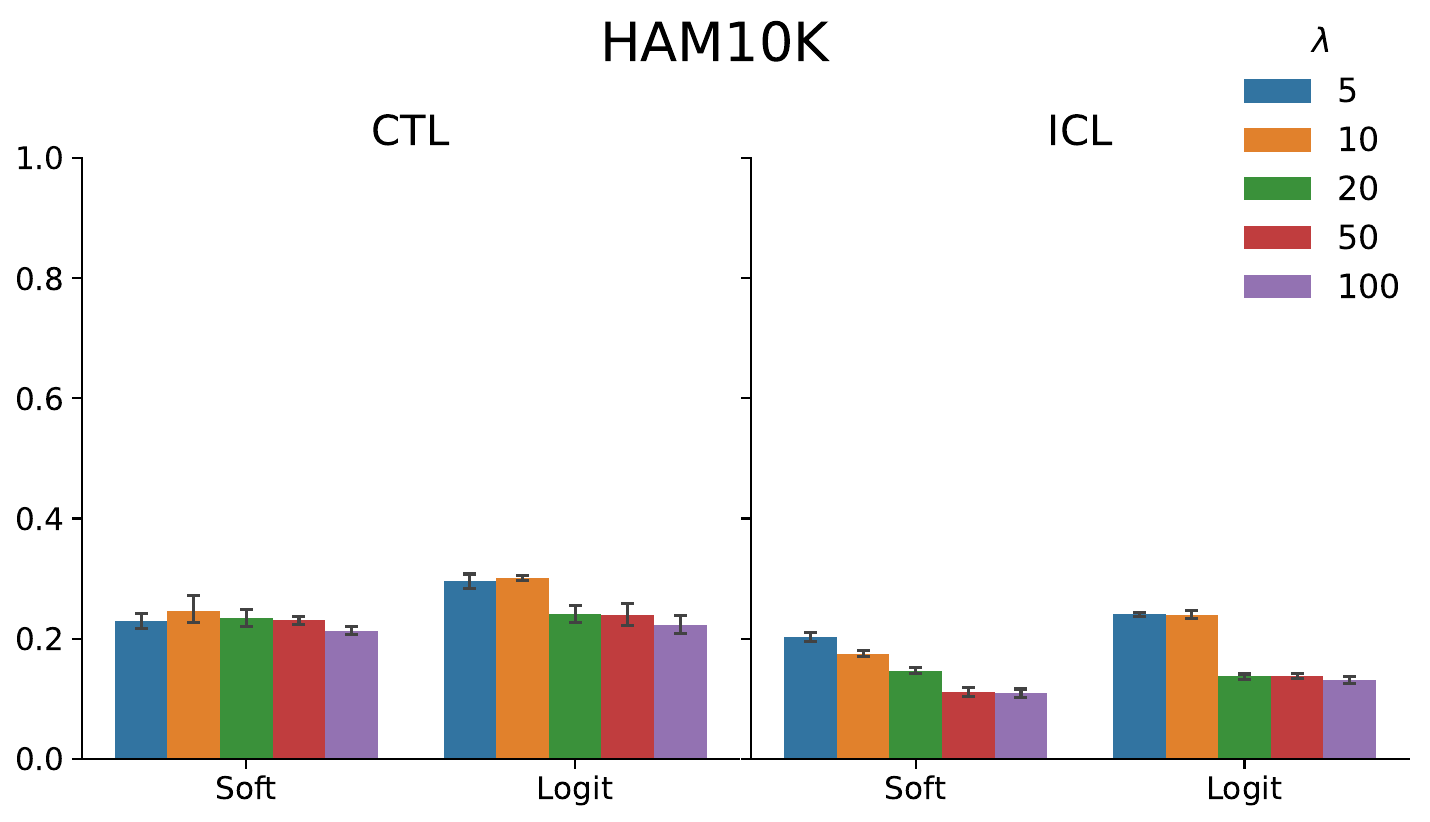} 
\end{minipage}
\end{center}

\caption{
Leakage scores evaluated on soft and logit CBMs at high $\lambda$.
    }
    \label{figure_soft_vs_logit_high_lambda}
\end{figure}

In soft and logit CBMs leakage is minimal for values of $\lambda > \lambda_{min}$, where $\lambda_{min}$ is specific to the considered model class (Figure \ref{figure_soft_vs_logit_high_lambda}). As $\lambda$ is increased leakage scores remain essentially constant to start. However, at higher values of $\lambda \gtrsim \lambda_{max}$ concept supervision may become too strong, at the expense of task learning. In such cases, 
\begin{enumerate}
    \item task performance may experience a drop. For instance, models trained on TabularToy(0.25) reach a task accuracy above 90\% in only 3 out of 5 folds for each $\lambda = 10, 20, 50$;
    \item leakage may raise again, to well above the minimal value. The soft models trained on 3dshapes(0) in Figure \ref{figure_soft_vs_logit_high_lambda} are a compelling example of this. 
\end{enumerate}
Finally, note that the minimal attainable leakage in logit CBMs is higher than in soft CBMs for any given dataset and choice of $\lambda$, in line with the expectations.

\section{Performance and scores of models with over-expressive concept representations, incomplete concept sets and misspecified heads}
\label{App_accuracies_incomplete_concepts_misspecified_head}

\begin{figure}[t]
\begin{minipage}[c]{0.325\textwidth} 
\centering
\includegraphics[width=\textwidth]{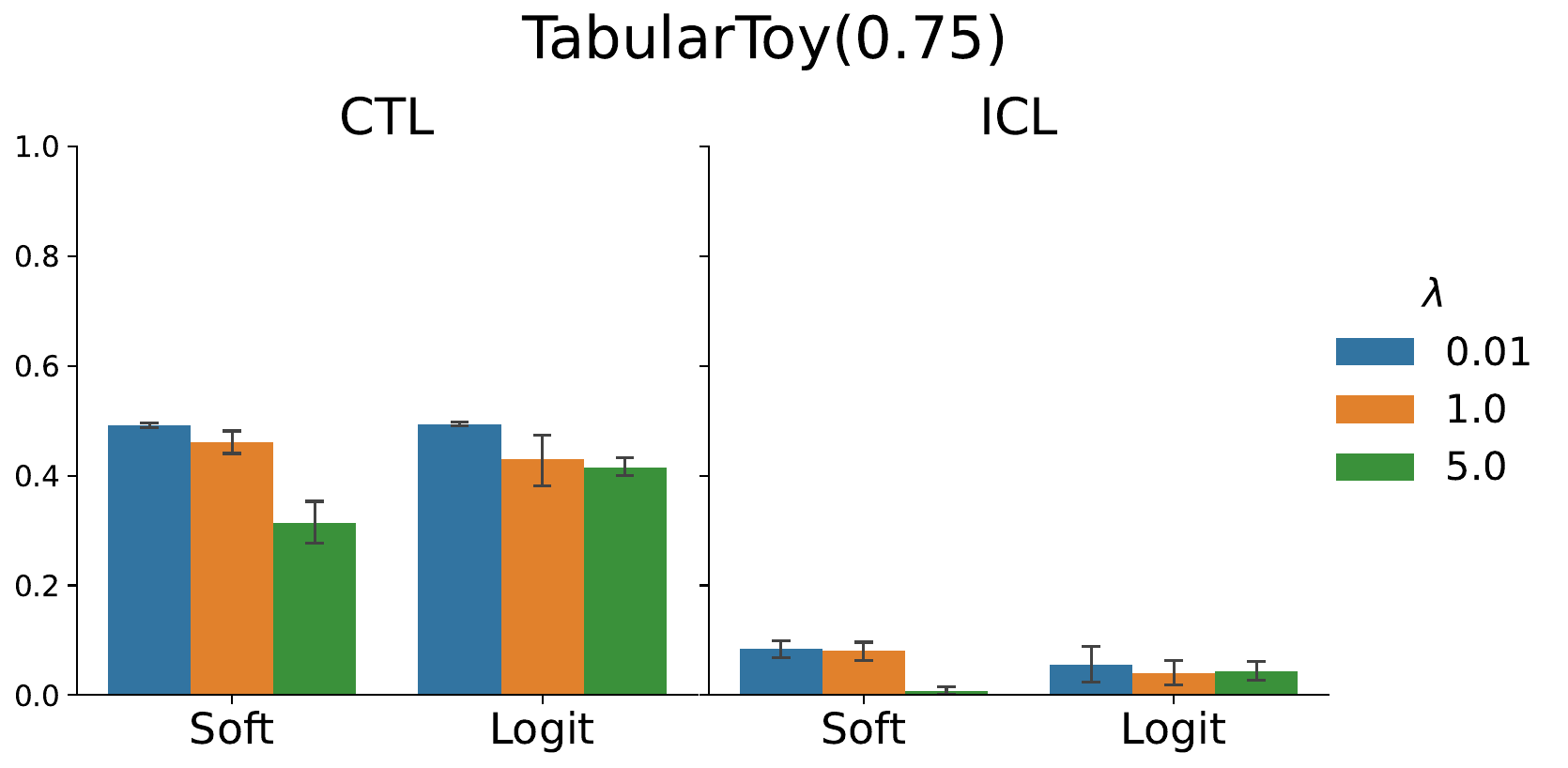} 
\end{minipage}
\begin{minipage}[c]{0.325\textwidth} 
\centering
\includegraphics[width=\textwidth]{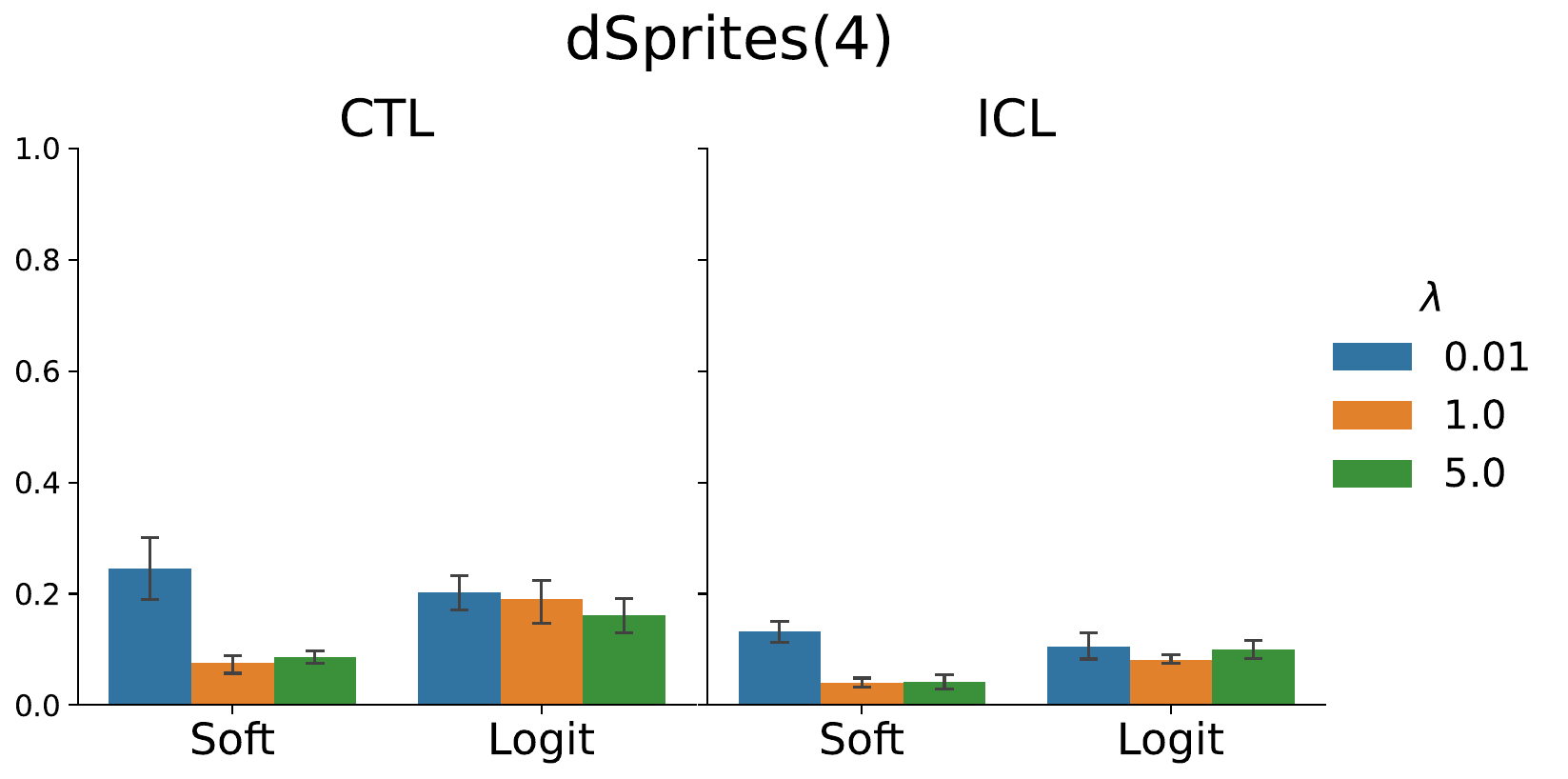} 
\end{minipage}
\begin{minipage}[c]{0.325\textwidth} 
\centering
\includegraphics[width=\textwidth]{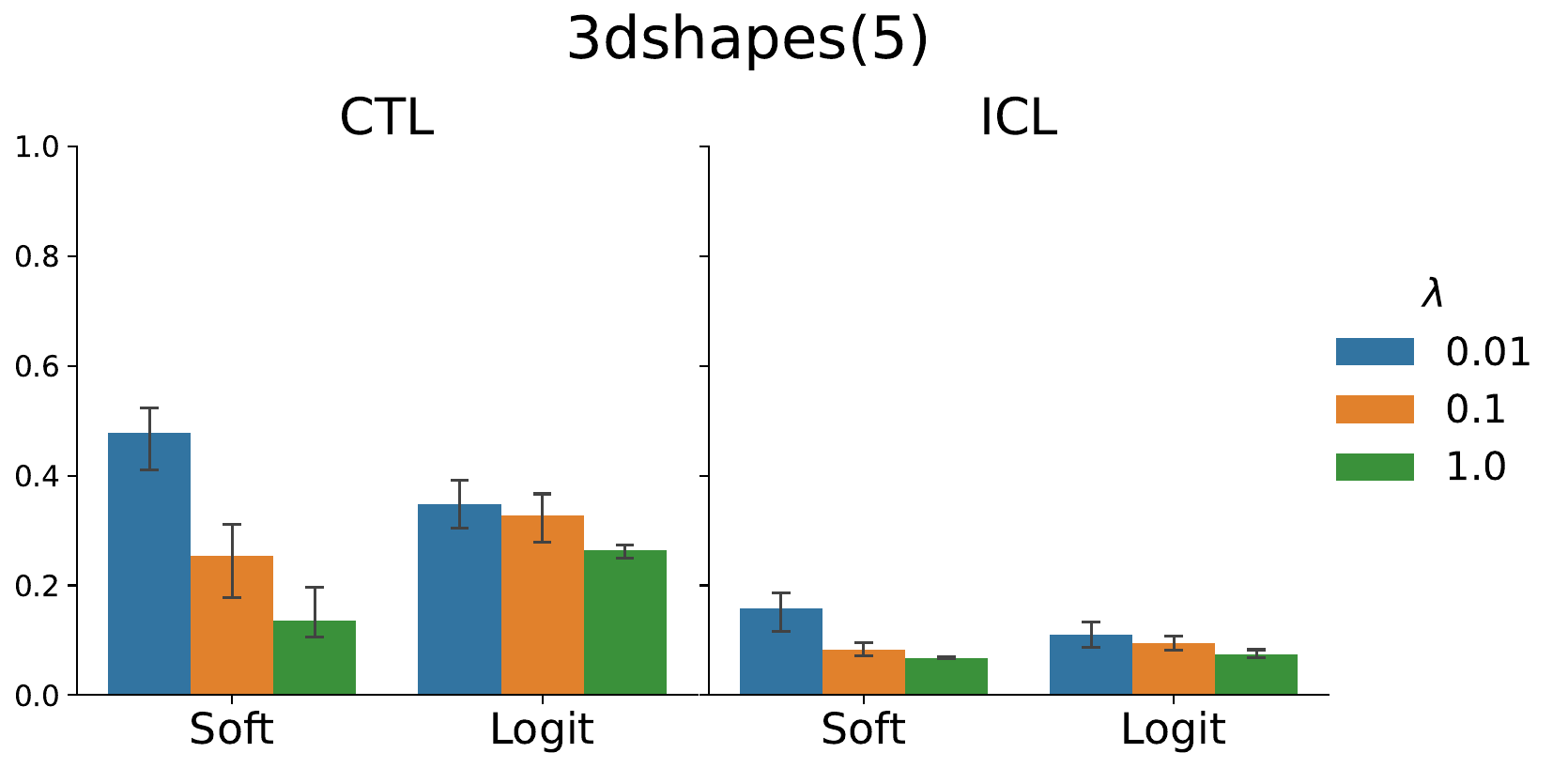} 
\end{minipage}
\caption{
Leakage scores evaluated for soft and logit models with different levels of concept supervision across datasets with high ground-truth interconcept MIs.
    }
    \label{figure_causes_soft_vs_logit_highCorr}
\end{figure}

\begin{figure}[t]
\begin{minipage}[c]{0.325\textwidth} 
\centering
\includegraphics[width=\textwidth]{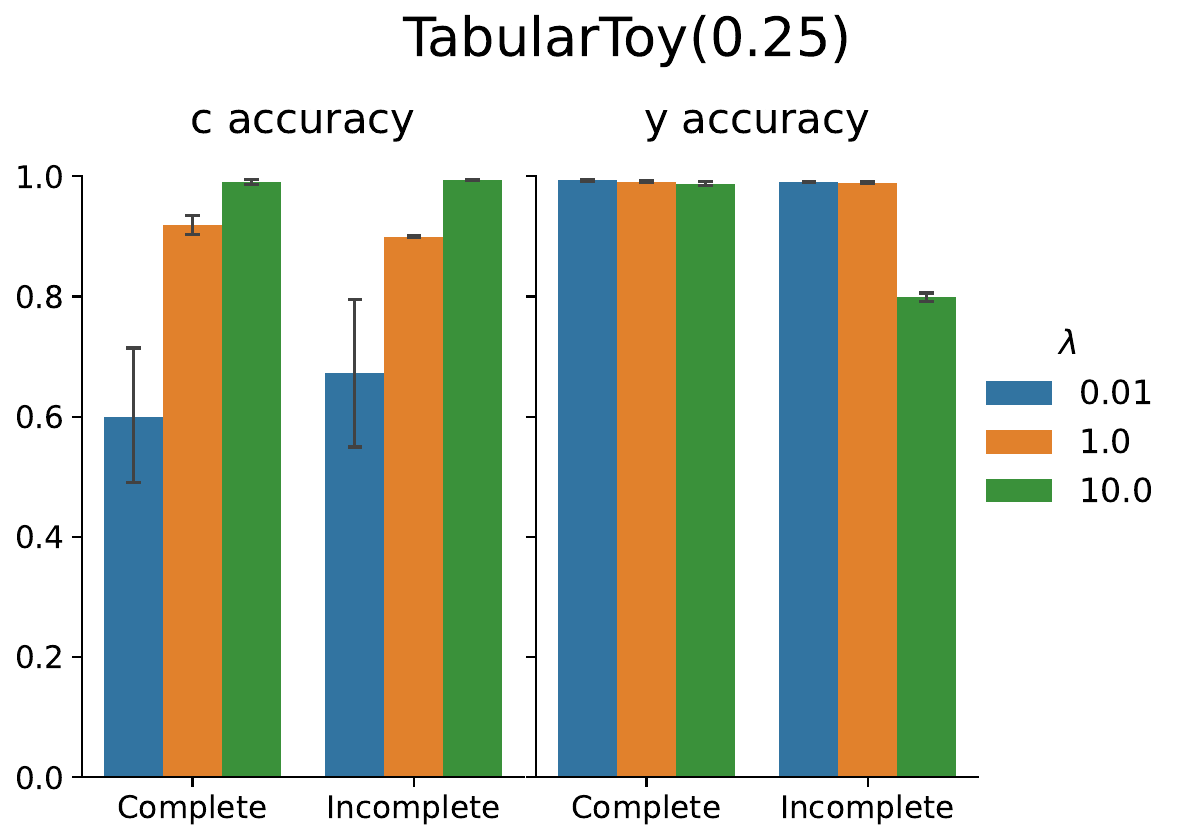} 
\end{minipage}
\begin{minipage}[c]{0.325\textwidth} 
\centering
\includegraphics[width=\textwidth]{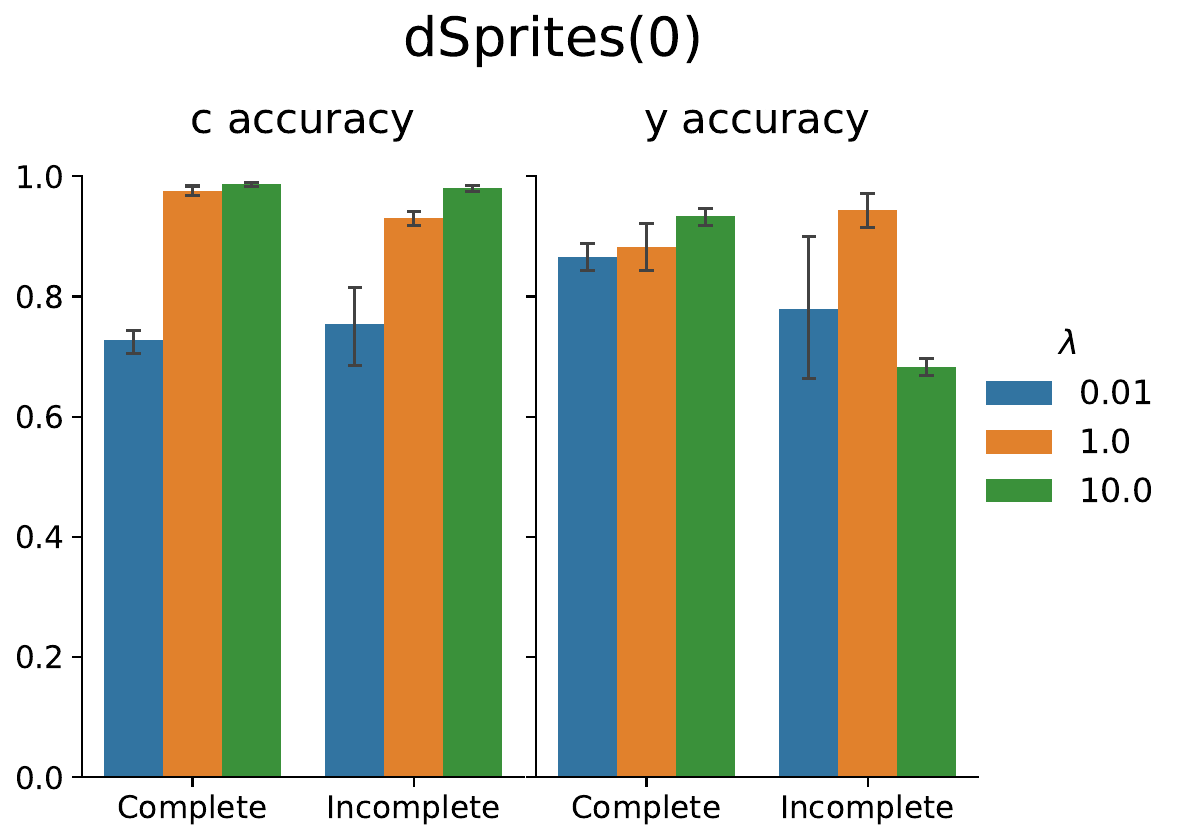} 
\end{minipage}
\begin{minipage}[c]{0.325\textwidth} 
\centering
\includegraphics[width=\textwidth]{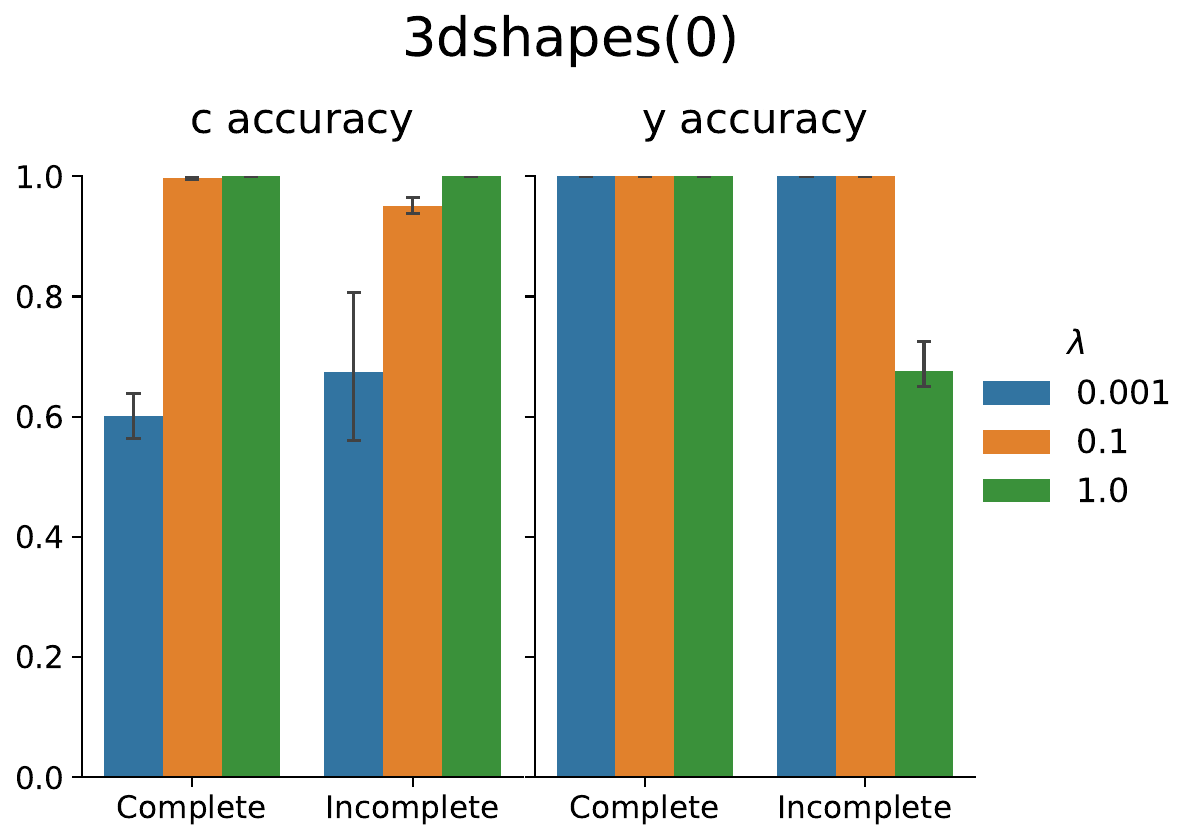} 
\end{minipage}

\begin{center}
    \begin{minipage}[c]{0.325\textwidth} 
\centering
\includegraphics[width=\textwidth]{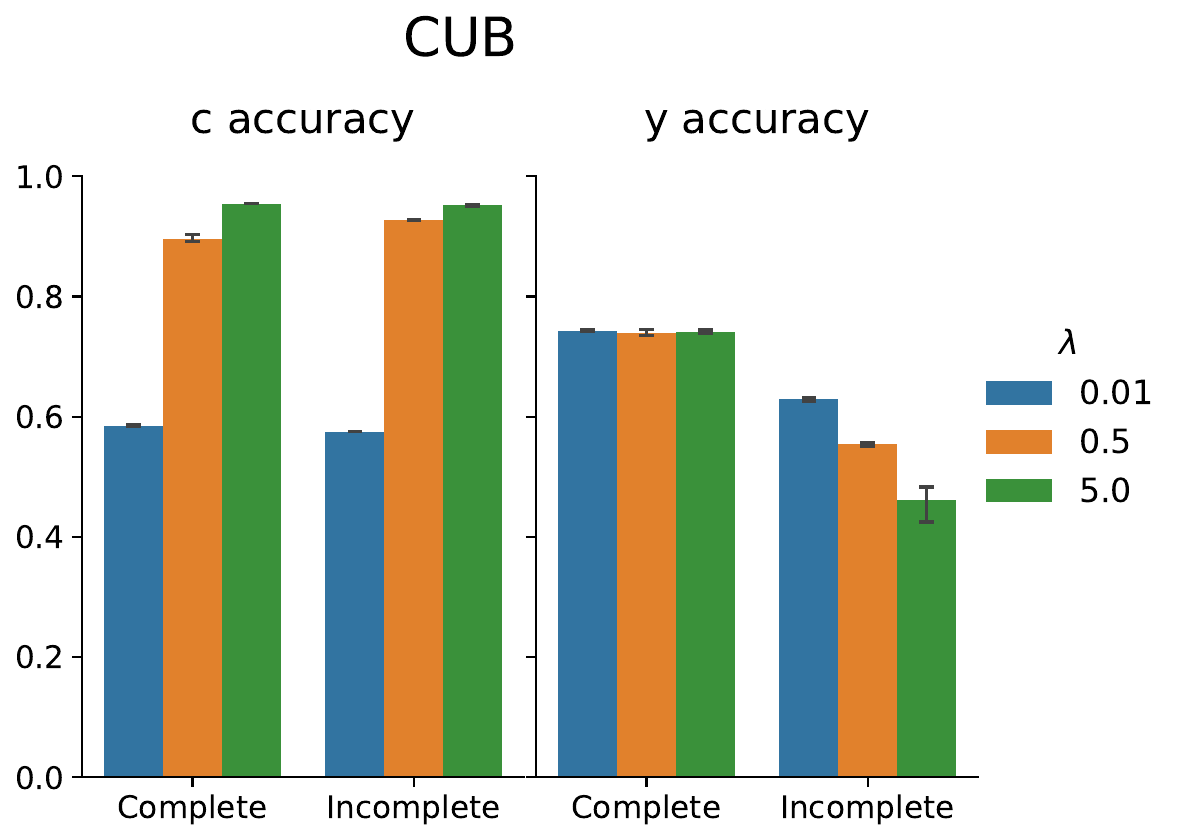} 
\end{minipage}
\begin{minipage}[c]{0.325\textwidth} 
\centering
\includegraphics[width=\textwidth]{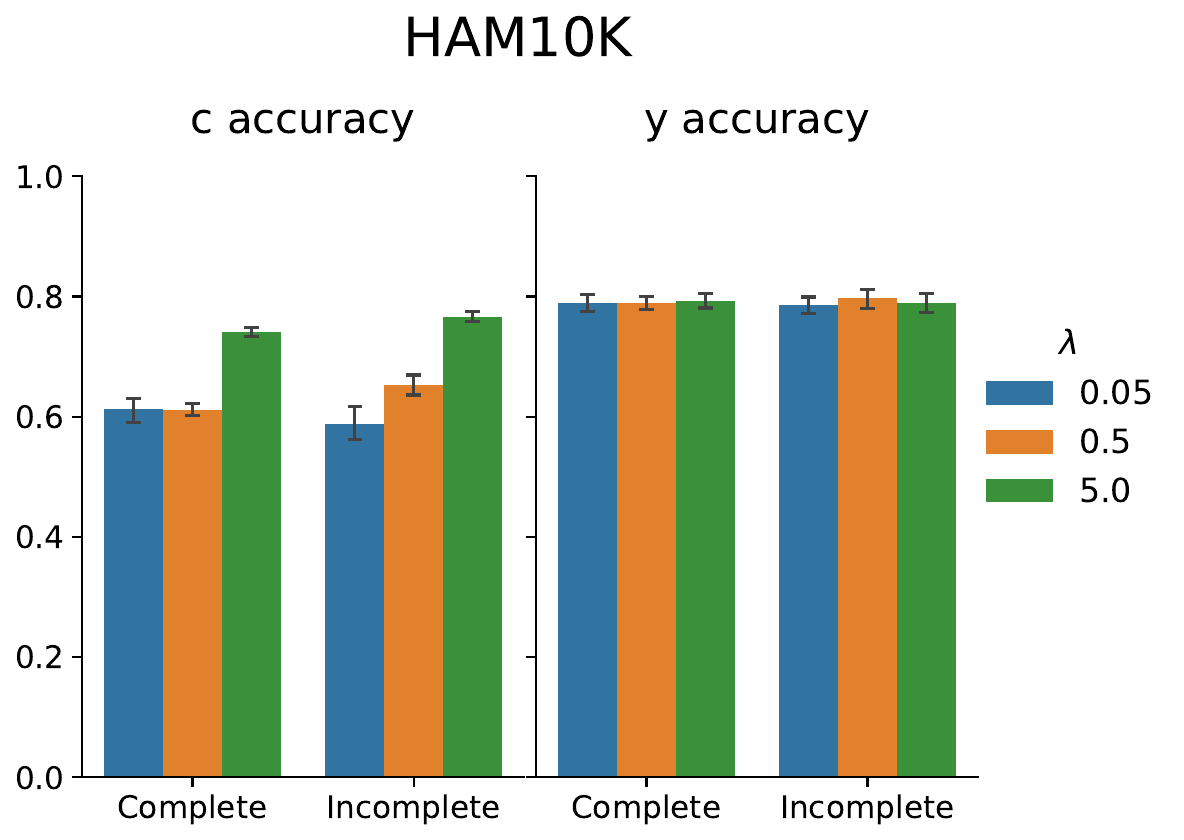} 
\end{minipage}
\end{center}
\caption{
Concepts and task accuracies for the models analysed in Figure \ref{figure_causes_incomplete_concept_set} on datasets with complete and incomplete sets of concepts.
    }
    \label{figure_causes_incomplete_concept_set_accs}
\end{figure}

\begin{figure}[t]
\begin{minipage}[c]{0.325\textwidth} 
\centering
\includegraphics[width=\textwidth]{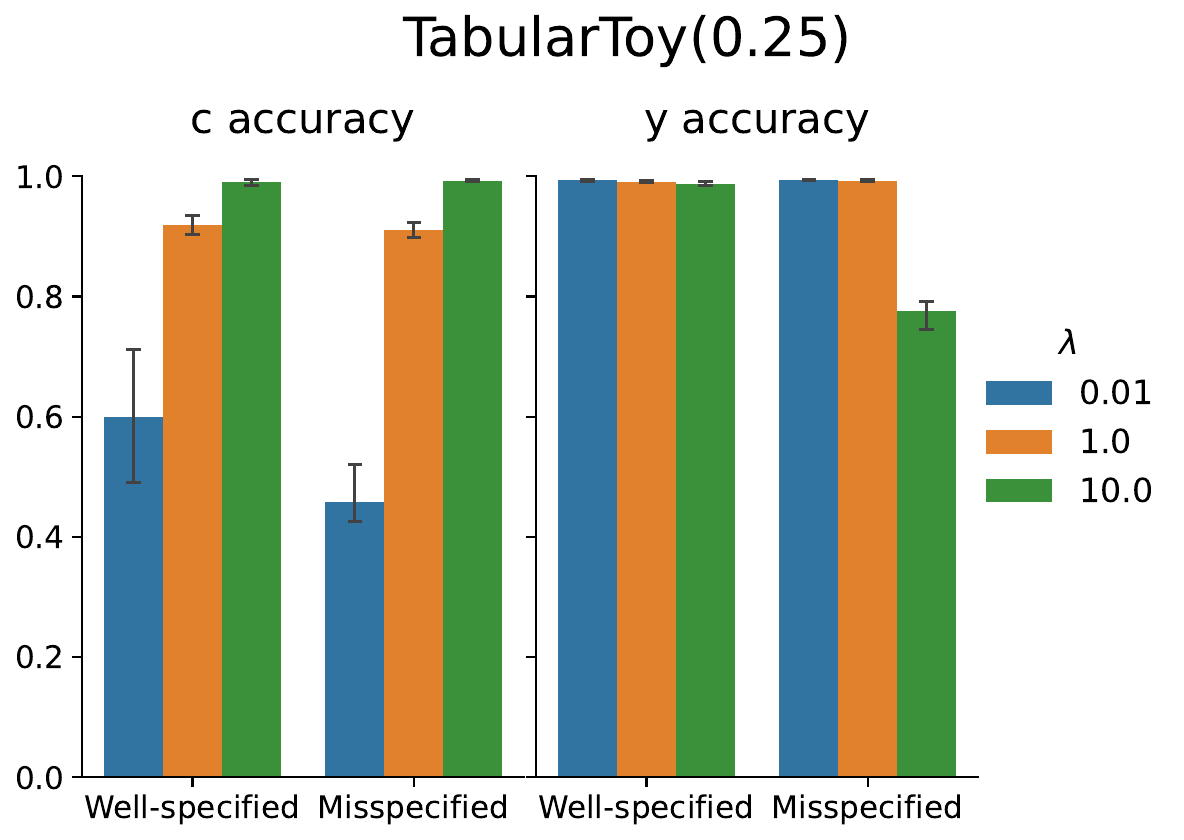} 
\end{minipage}
\begin{minipage}[c]{0.325\textwidth} 
\centering
\includegraphics[width=\textwidth]{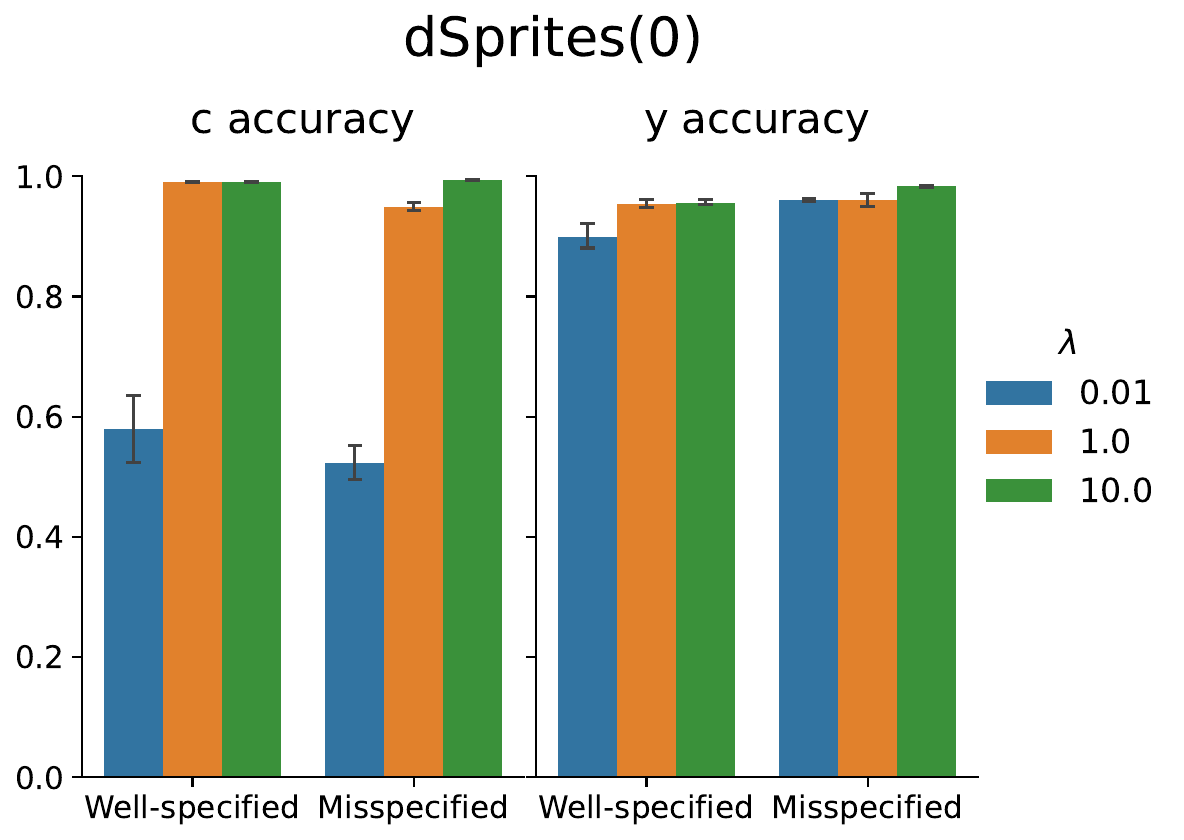} 
\end{minipage}
\begin{minipage}[c]{0.325\textwidth} 
\centering
\includegraphics[width=\textwidth]{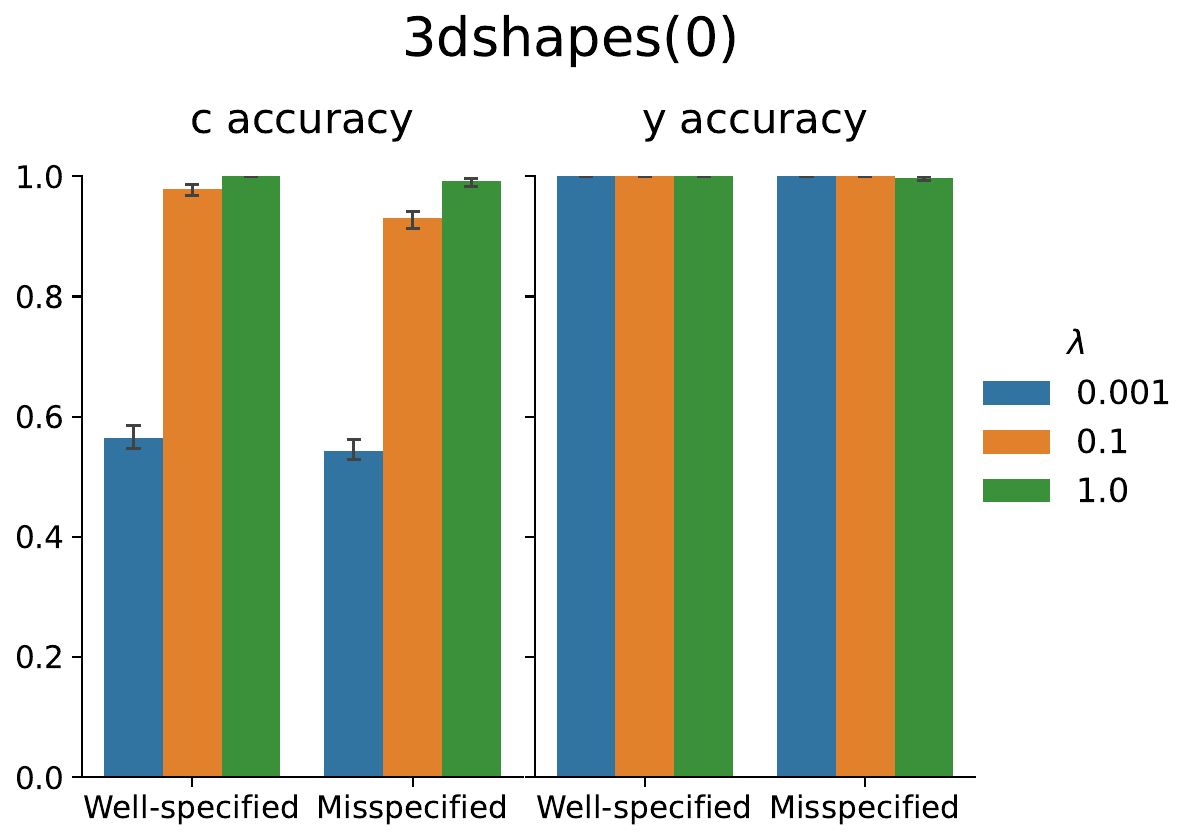} 
\end{minipage}
\caption{
Concepts and task accuracies evaluated for models with a linear head analysed in Figure \ref{figure_causes_c2y_misspecification} trained on datasets with linear and non-linear tasks as functions of concepts.
    }
    \label{figure_causes_c2y_misspecification_accs}
\end{figure}

In Figure \ref{figure_causes_soft_vs_logit_highCorr} we present the leakage scores across datasets with high ground-truth interconcept MI.
In Figure \ref{figure_causes_incomplete_concept_set_accs} we show the task and concept accuracies of the soft CBMs trained on the complete and incomplete sets of concepts discussed in Section \ref{sec_causes_of_leakage}. Note in particular the decrease in task accuracy at high values of $\lambda$ that occurs in the case of incomplete concept sets. At low $\lambda$, leakage is sufficient to ensure task accuracy remains essentially constant and approximately equivalent to that of the case of complete concept sets.
Figure \ref{figure_causes_c2y_misspecification_accs} shows the task and concept accuracies of the well-specified and misspecified soft CBMs evaluated in Figure \ref{figure_causes_c2y_misspecification} and discussed in Section \ref{sec_causes_of_leakage}.

\begin{figure}[t]
\begin{minipage}[c]{0.23\textwidth} 
\centering \small \sffamily
$\;\;\;\;\;$ TabularToy(0.25)
\vspace{0.2cm}
\end{minipage}
\hfill
\begin{minipage}[c]{0.23\textwidth} 
\centering \small \sffamily
dSprites(0) \hspace{0.67cm}
\vspace{0.2cm}
\end{minipage}
\begin{minipage}[c]{0.23\textwidth}
\centering \small \sffamily
 3dshapes(0) \hspace{0.35cm}
\vspace{0.2cm}
\end{minipage}
\begin{minipage}[c]{0.23\textwidth}
\centering \small \sffamily
 \hspace{0.35cm}HAM10K
\vspace{0.2cm}
\end{minipage}

\begin{minipage}[c]{0.24\textwidth} 
\centering
\includegraphics[width=0.83\textwidth]{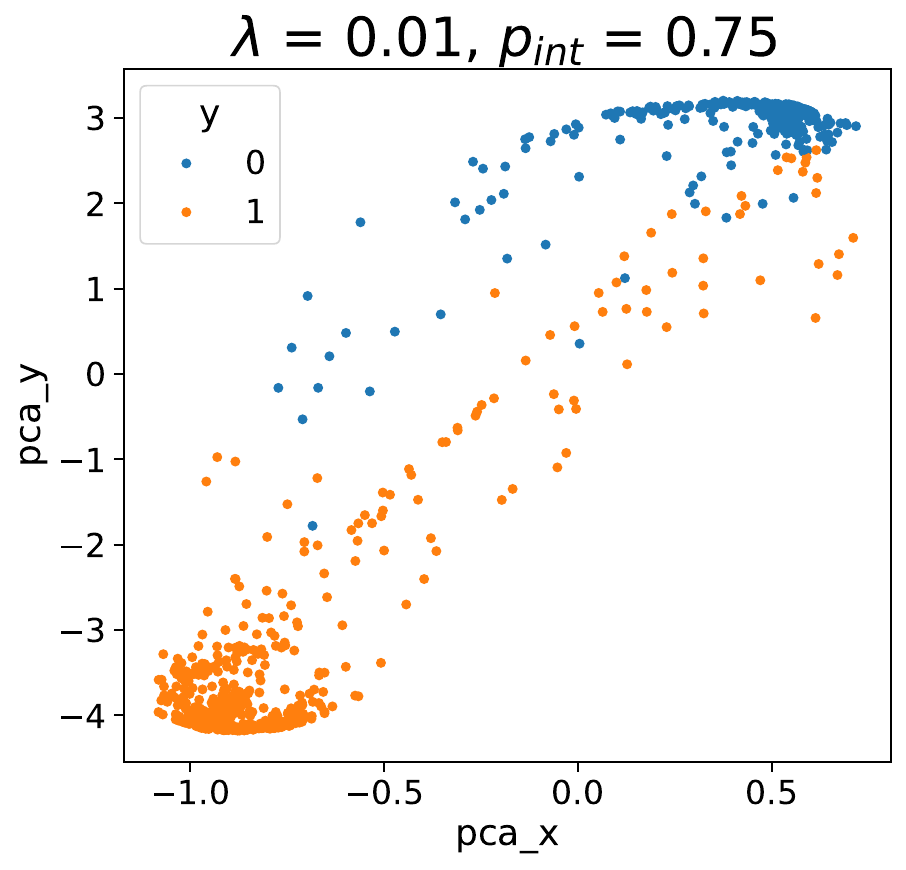} 
\end{minipage}
\hfill
\begin{minipage}[c]{0.24\textwidth} 
\centering
    \includegraphics[width=0.96\textwidth]{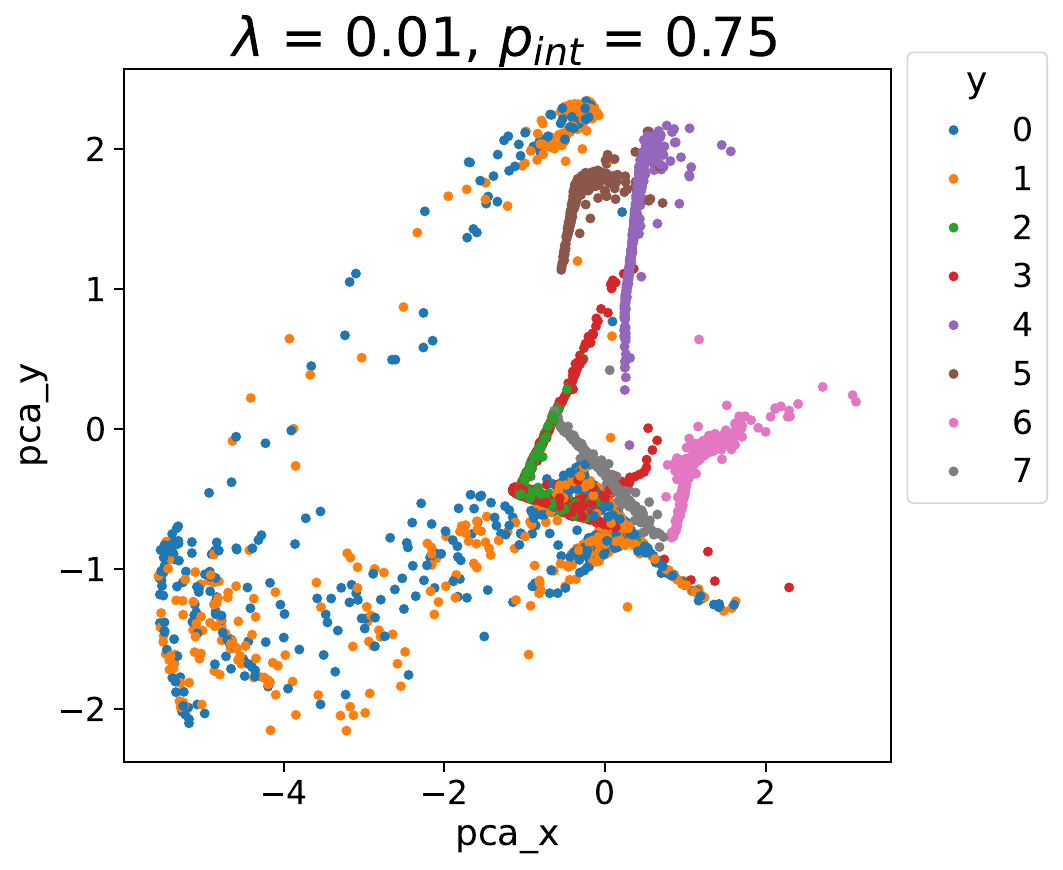} 
\end{minipage}
\begin{minipage}[c]{0.24\textwidth} 
\centering
    \includegraphics[width=1\textwidth]{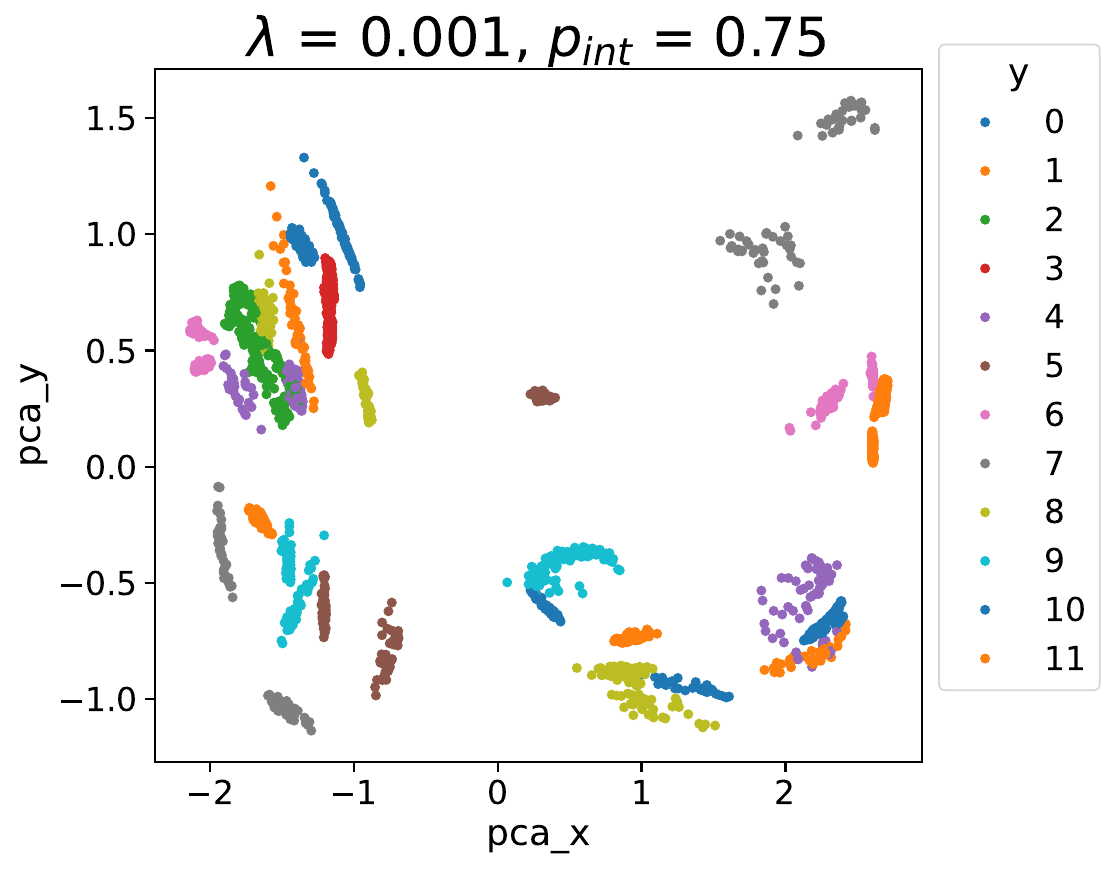} 
\end{minipage}
\begin{minipage}[c]{0.24\textwidth} 
\centering
    \includegraphics[width=0.84\textwidth]{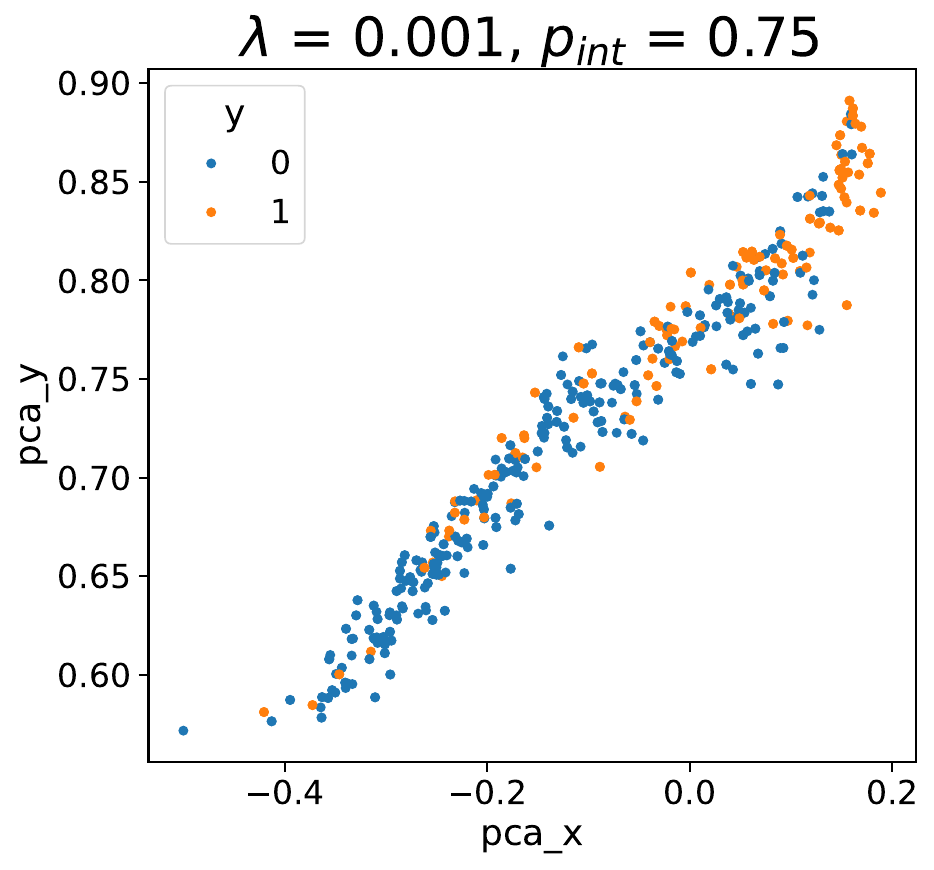} 
\end{minipage}

\begin{minipage}[c]{0.24\textwidth} 
\centering
\includegraphics[width=0.83\textwidth]{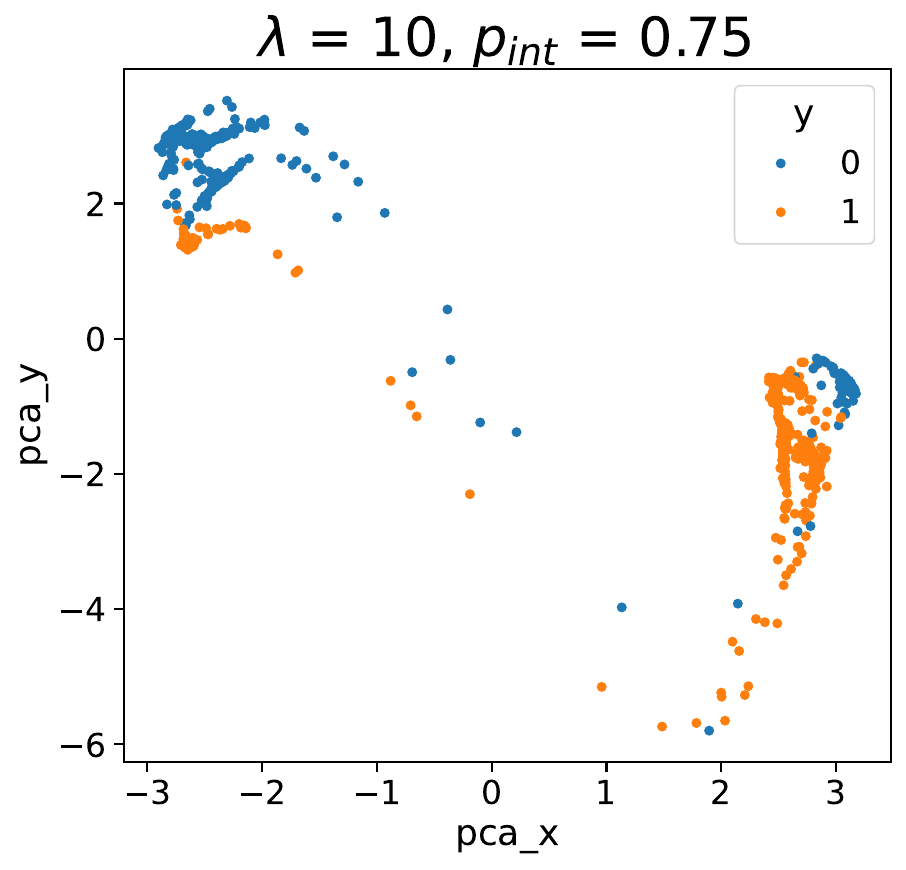} 
\end{minipage}
\hfill
\begin{minipage}[c]{0.24\textwidth} 
\centering
    $\;$\includegraphics[width=0.96\textwidth]{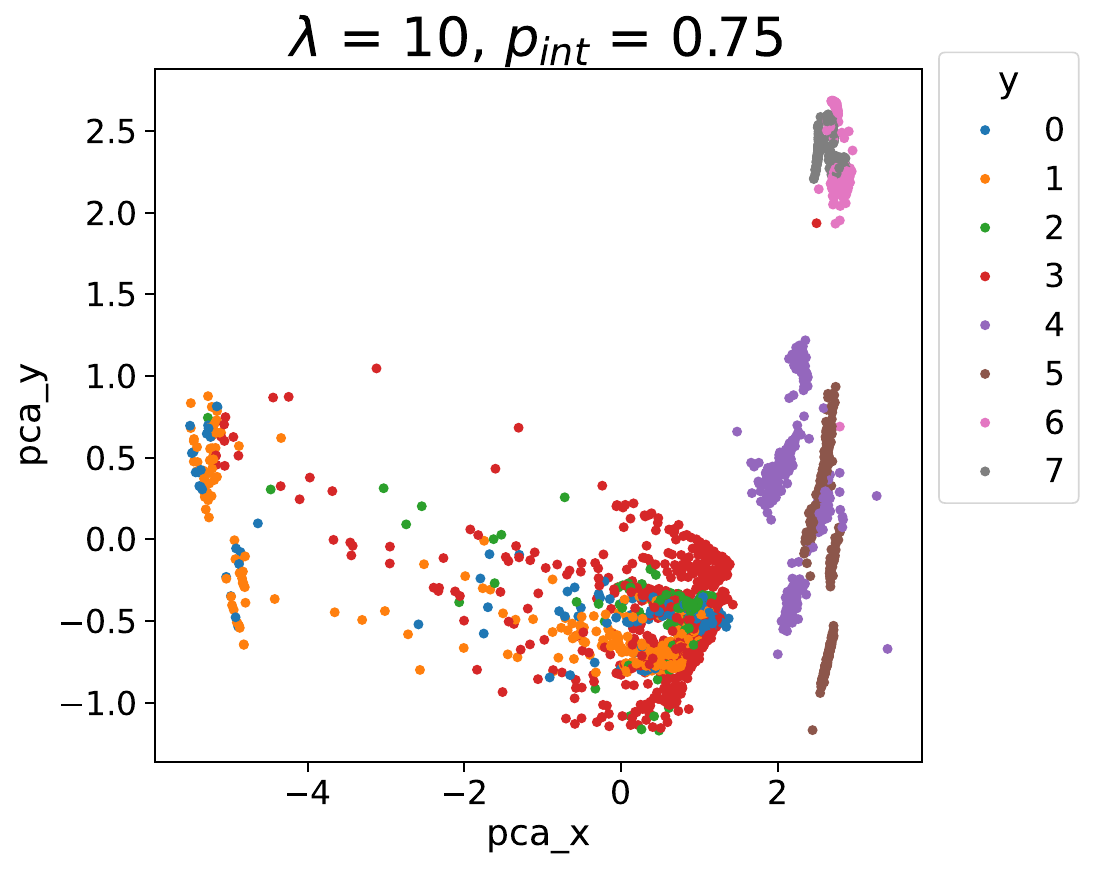} 
\end{minipage}
\begin{minipage}[c]{0.24\textwidth} 
\centering
    \includegraphics[width=0.99\textwidth]{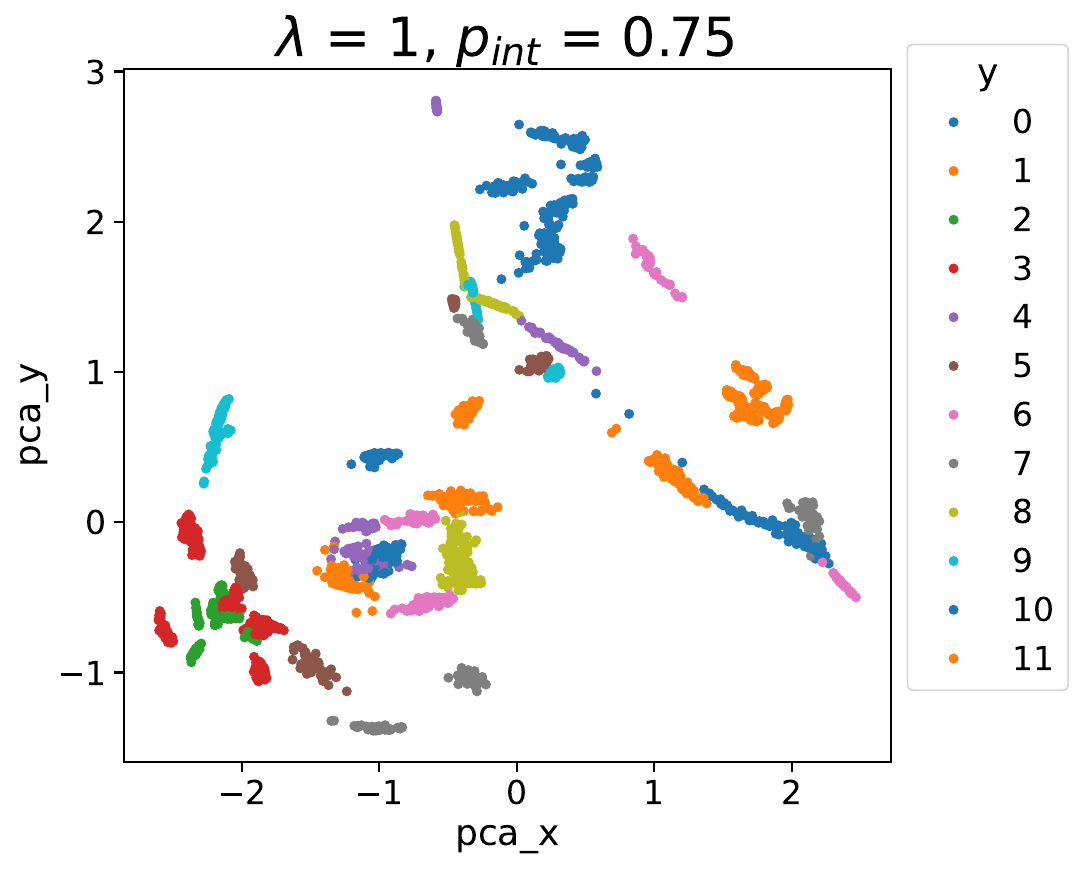} 
\end{minipage}
\begin{minipage}[c]{0.24\textwidth} 
\centering
    \includegraphics[width=0.84\textwidth]{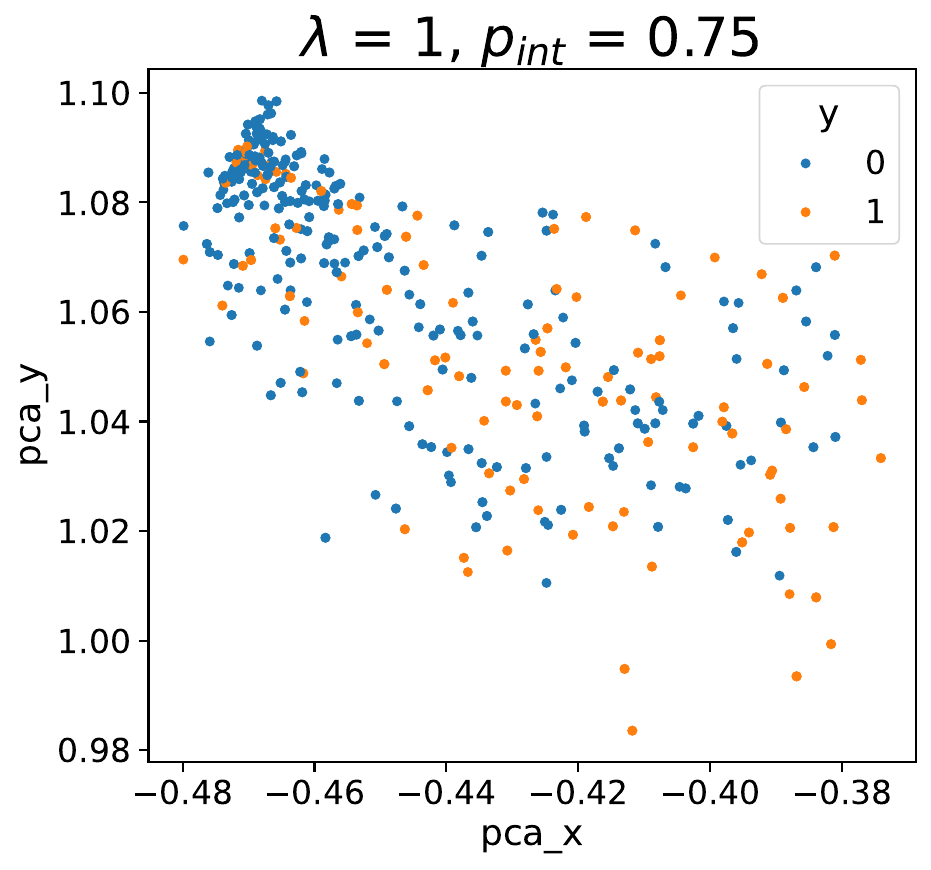} 
\end{minipage}
    \caption{
    2-dimensional PCA projections of the weighted embeddings $\bm{\hat{c}^{w}}_1$ for the first concept across datasets and for different values of $\lambda$ at high $p_{int}=0.75$. The colouring indicates the value of the ground-truth task label.
    }
    \label{figure_PCAs_y_app}
    \vskip 0.5cm
\end{figure}

\begin{figure}[t]
\begin{minipage}[c]{0.23\textwidth} 
\centering \small \sffamily
$\;\;\;\;\;$ TabularToy(0.25)
\vspace{0.2cm}
\end{minipage}
\hfill
\begin{minipage}[c]{0.23\textwidth} 
\centering \small \sffamily
dSprites(0) \hspace{0.67cm}
\vspace{0.2cm}
\end{minipage}
\begin{minipage}[c]{0.23\textwidth}
\centering \small \sffamily
 3dshapes(0) \hspace{0.35cm}
\vspace{0.2cm}
\end{minipage}
\begin{minipage}[c]{0.23\textwidth}
\centering \small \sffamily
 HAM10K \hspace{0.18cm}
\vspace{0.2cm}
\end{minipage}

\begin{minipage}[c]{0.24\textwidth} 
\centering
\includegraphics[width=0.99\textwidth]{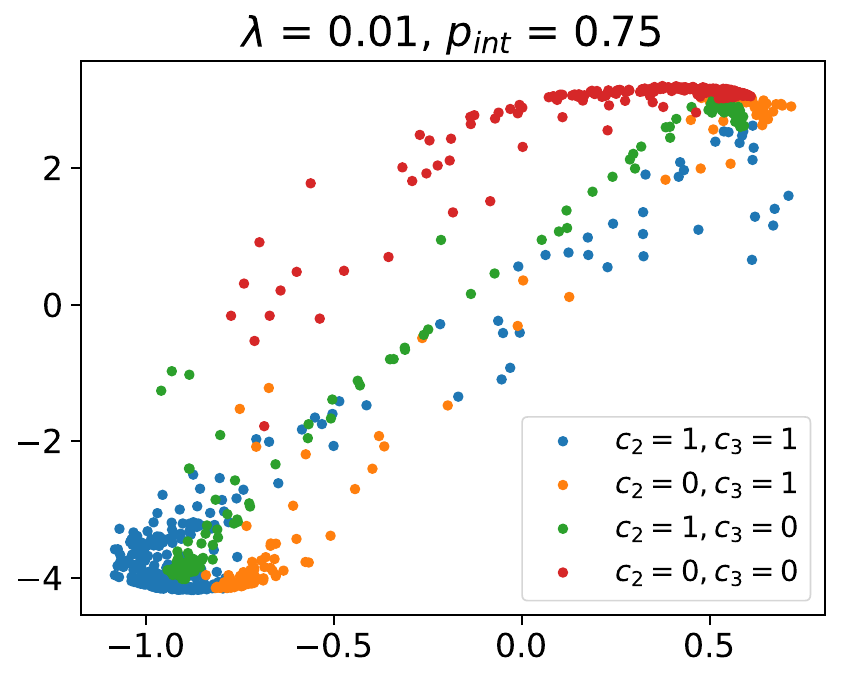} 
\end{minipage}
\begin{minipage}[c]{0.24\textwidth} 
\centering
    \includegraphics[width=1\textwidth]{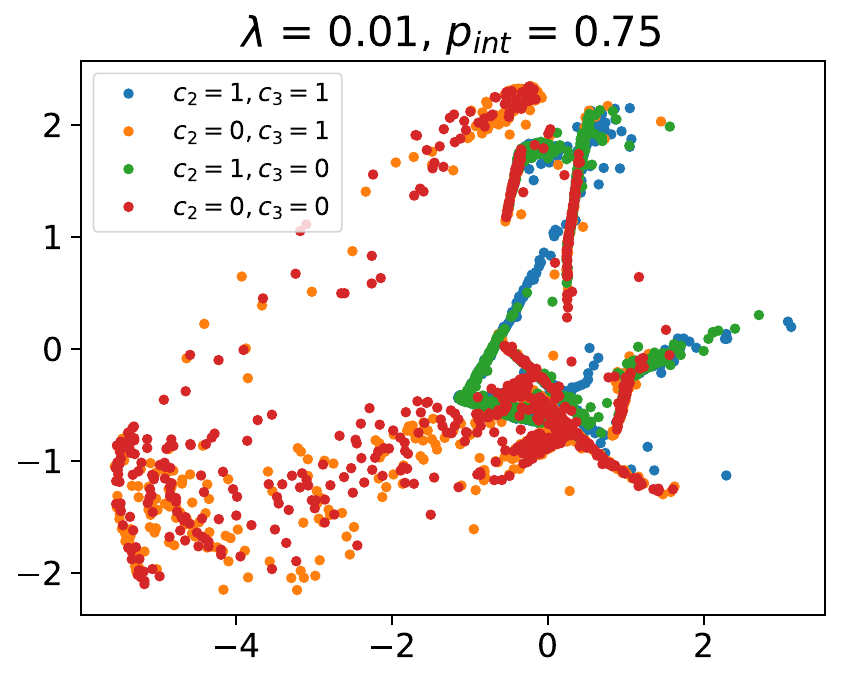} 
\end{minipage}
\begin{minipage}[c]{0.24\textwidth} 
\centering
    \includegraphics[width=1\textwidth]{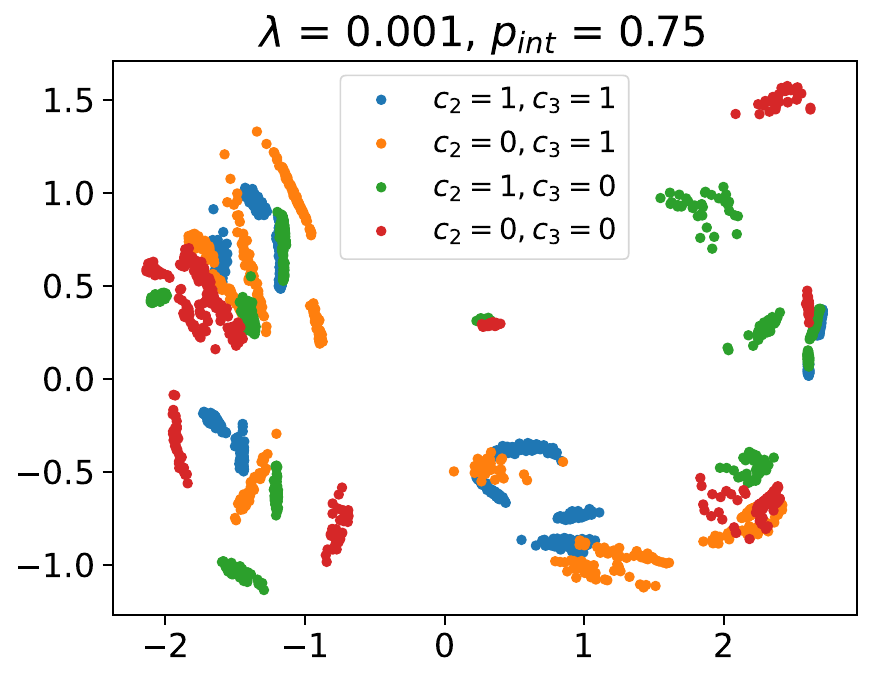} 
\end{minipage}
\begin{minipage}[c]{0.24\textwidth} 
\centering
\includegraphics[width=0.99\textwidth]{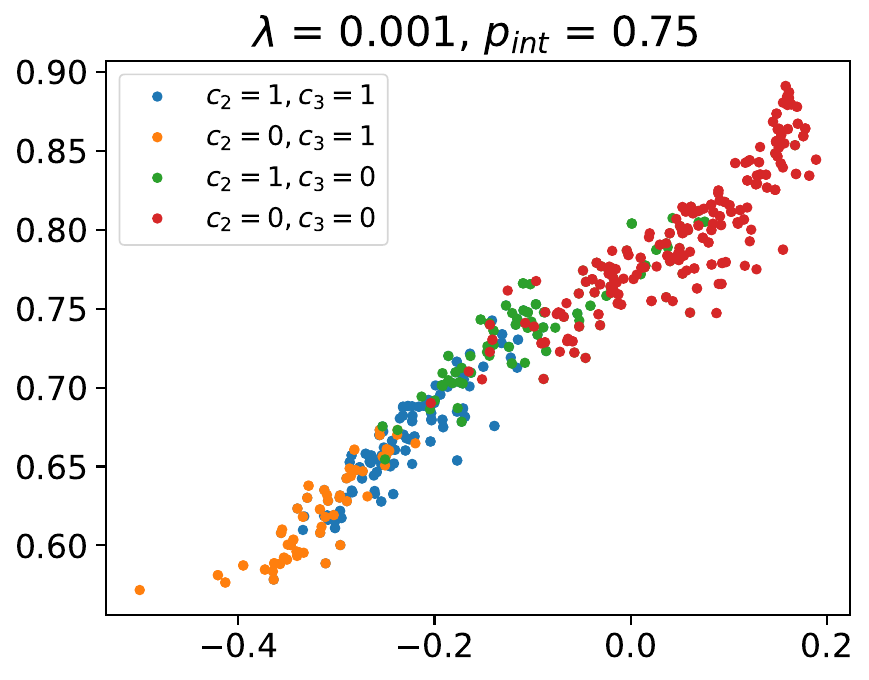} 
\end{minipage}

\begin{minipage}[c]{0.24\textwidth} 
\centering
\includegraphics[width=0.98\textwidth]{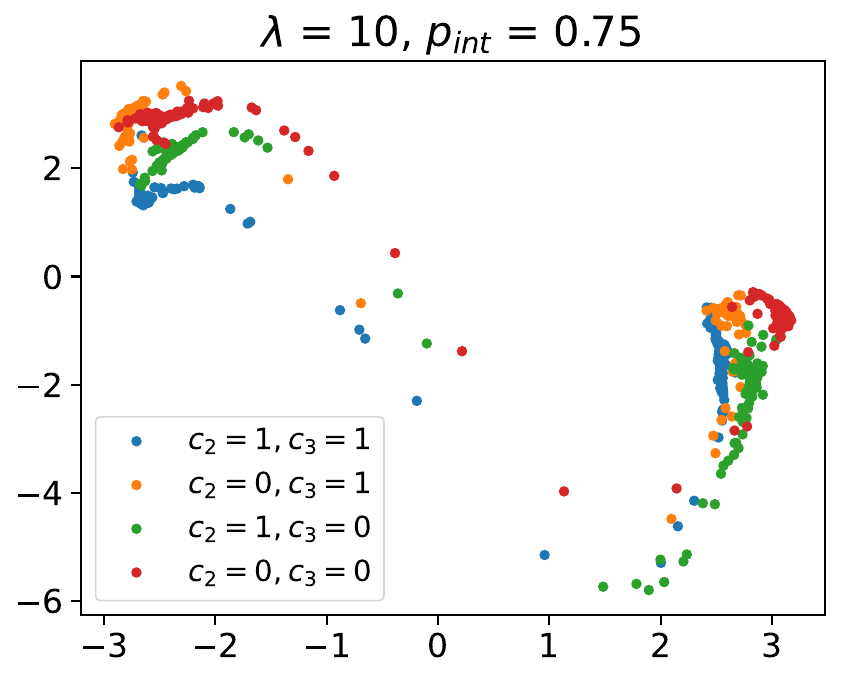} 
\end{minipage}
\begin{minipage}[c]{0.24\textwidth} 
\centering
    $\;$\includegraphics[width=0.99\textwidth]{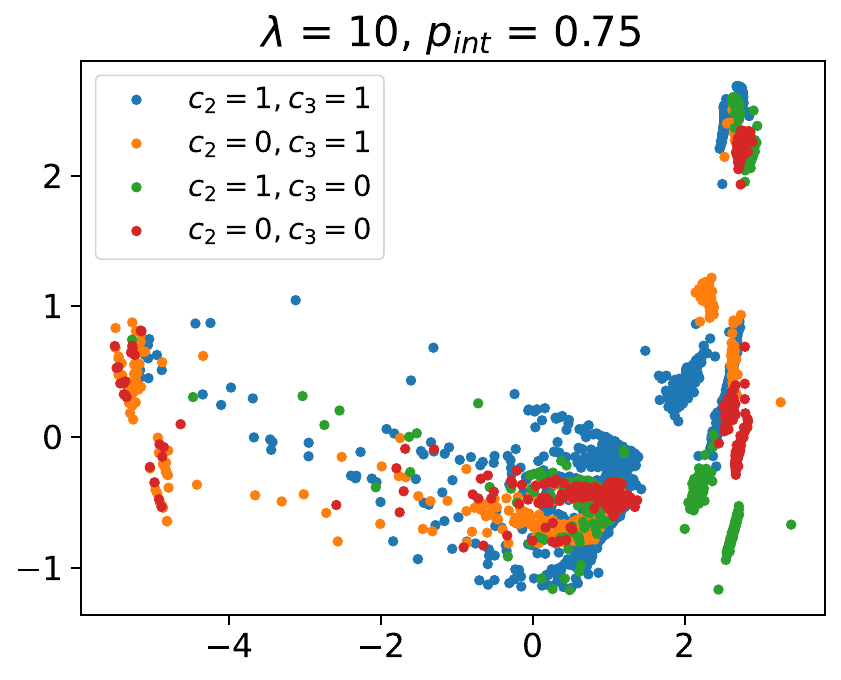} 
\end{minipage}
\begin{minipage}[c]{0.24\textwidth} 
\centering
    \includegraphics[width=0.99\textwidth]{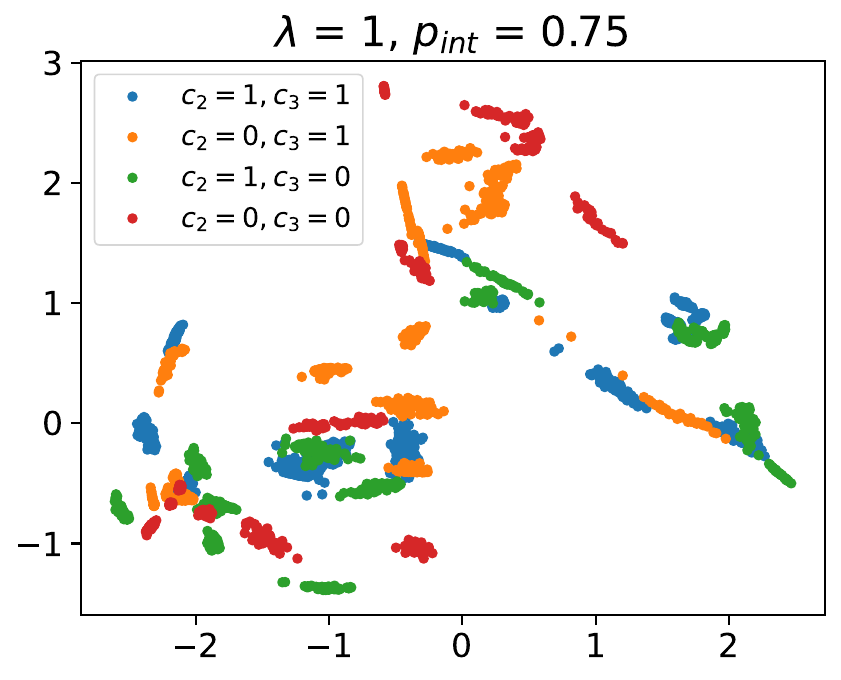} 
\end{minipage}
\begin{minipage}[c]{0.24\textwidth} 
\centering
\includegraphics[width=1\textwidth]{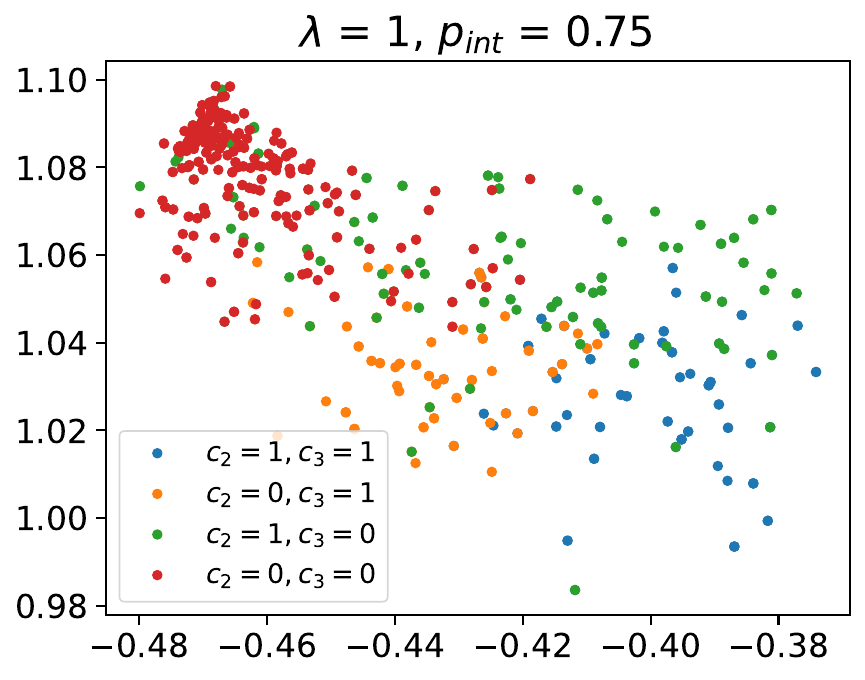} 
\end{minipage}
    \caption{
    2-dimensional PCA projections of the weighted embeddings $\bm{\hat{c}^{w}}_1$ for the first concept across datasets and for different values of $\lambda$ at high $p_{int}=0.75$. The colouring indicates the ground-truth value of concepts 2 and 3.
    }
    \label{figure_PCAs_c_app}
\end{figure}

\section{Further results on CEMs}
\label{App_further_CEM_noninterpret}

\paragraph{Embedding structure at non-vanishing $p_{int}$.}
 In Figures \ref{figure_PCAs_y_app} and \ref{figure_PCAs_c_app} we display the PCA projections of the weighted vectors $\bm{\hat{c}^{w}}_1$ with colouring based on the ground-truth value of the task label and of concepts 2 and 3 respectively, at non-vanishing values of $p_{int}$ and for low and high values of $\lambda$.

\paragraph{Concept accuracy and intervention performance are not measures of leakage in CEMs.} As demonstrated in Section \ref{sec_interpretability_CEMs}, CEMs generally encode high amounts of concepts-task leakage, and from this perspective, concept accuracy is only a measure of how predictive the vectors $\bm{\hat{c}^{w}}_i$ are of concept $i$, regardless of the leaked information about the task in the weighted embeddings. Furthermore, as shown in Figure \ref{figure_CEM_ICL},
interconcept leakage increases with concept supervision. Thus $c_{acc}$ typically correlates with interconcept leakage in CEMs.

\begin{figure}[t]
\centering
\begin{minipage}[c]{0.19\textwidth} 
\centering
\includegraphics[width=0.95\textwidth]{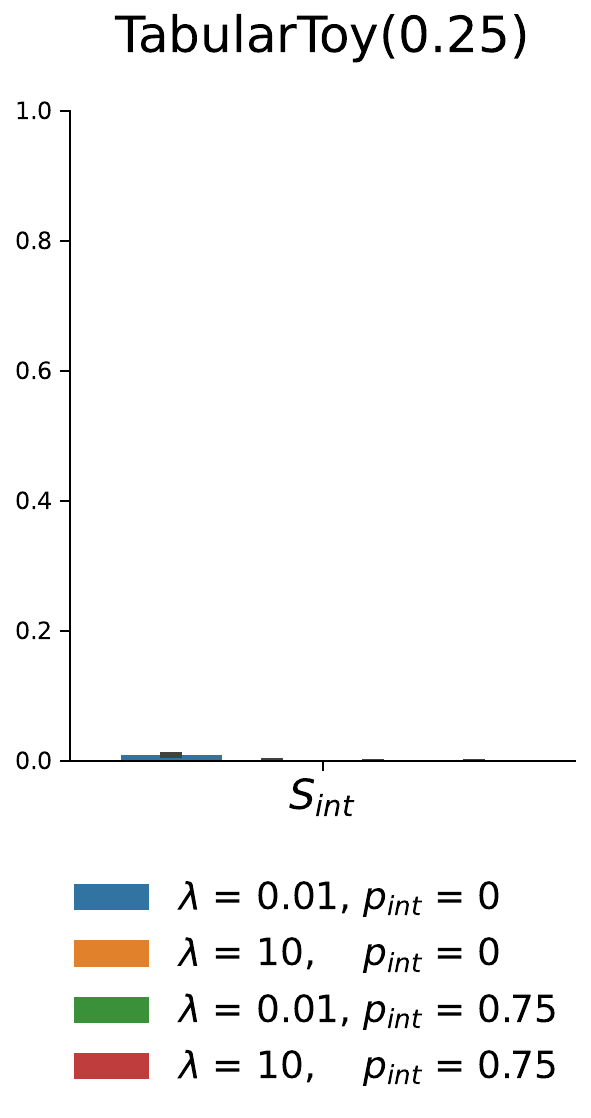} 
\end{minipage}
\begin{minipage}[c]{0.19\textwidth} 
\centering
\includegraphics[width=1\textwidth]{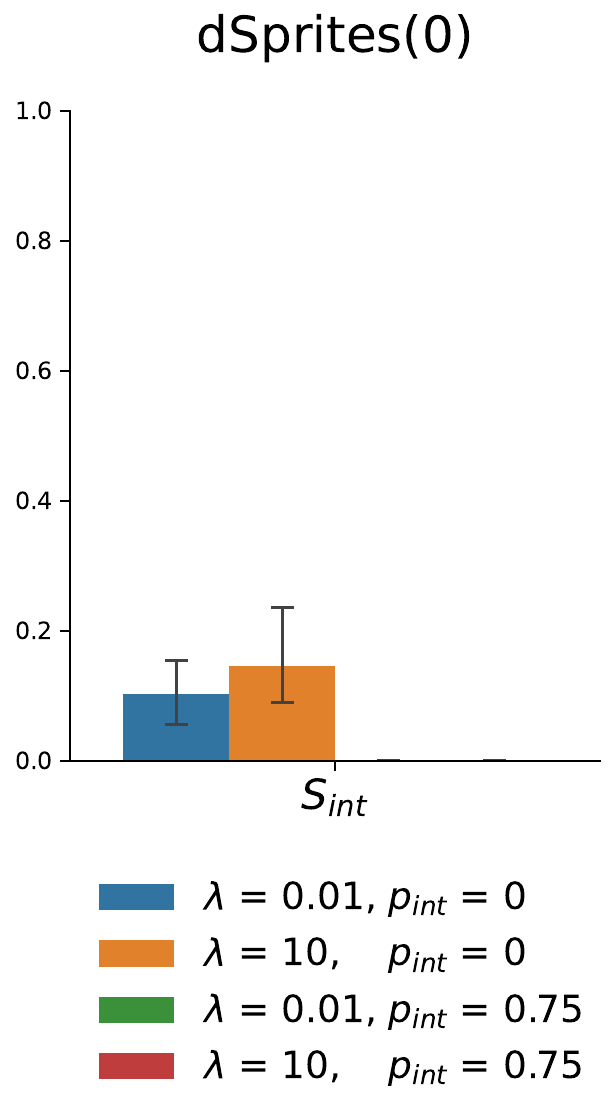} 
\end{minipage}
\begin{minipage}[c]{0.19\textwidth} 
\centering
\includegraphics[width=0.95\textwidth]{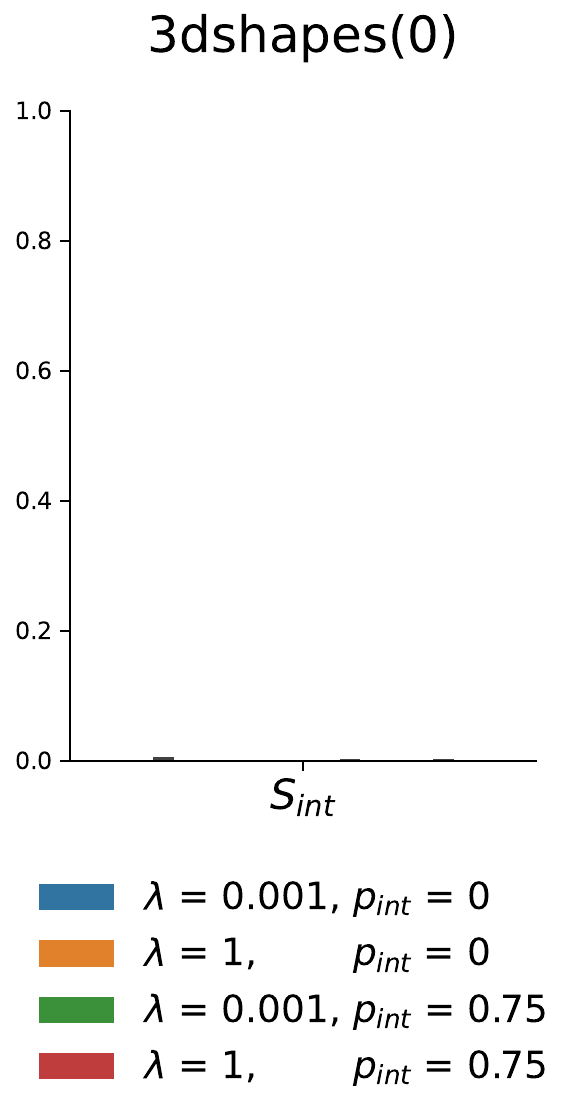} 
\end{minipage}
\begin{minipage}[c]{0.19\textwidth} 
\centering
\includegraphics[width=0.95\textwidth]{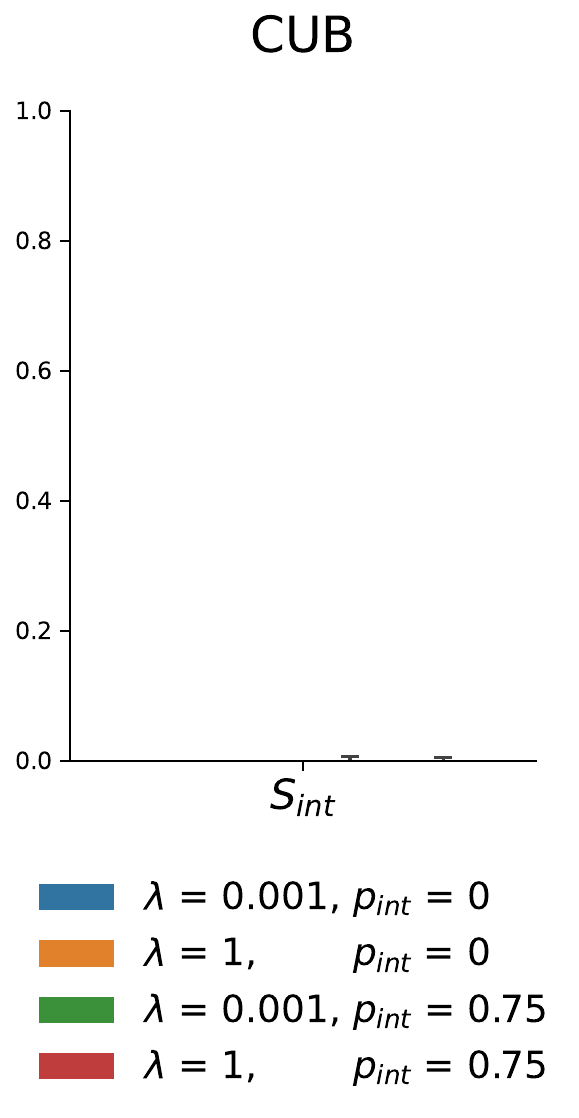} 
\end{minipage}
\begin{minipage}[c]{0.19\textwidth} 
\centering
\includegraphics[width=0.95\textwidth]{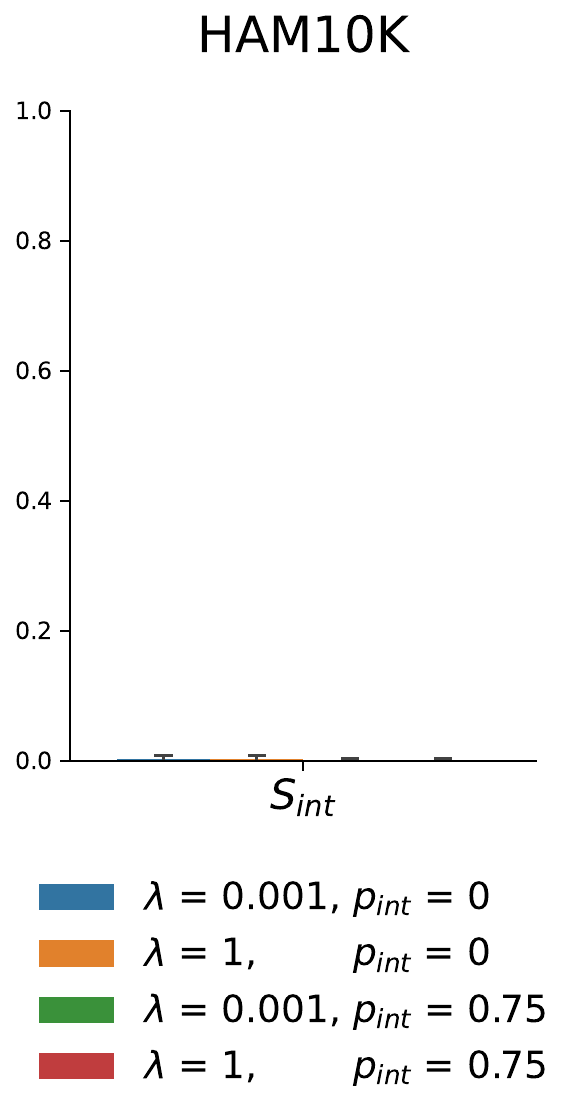} 
\end{minipage}
\caption{
The $\textrm{S}_{int} $ score for CEMs with low and high values of $\lambda$ and $p_{int}$.
}
    \label{figure_CEM_Sint}  
\end{figure}

Intervention performance is not an indicator of leakage either; indeed it can be high in models with severe leakage (see Figure \ref{figure_CEM_Sint}). In particular, CEMs with high concepts-task leakage in both $\bm{\hat{c}^+}_i$ and $\bm{\hat{c}^-}_i$, or where alignment leakage is present are able to achieve a vanishing $\textrm{S}_{int} $.
This motivates the definition and adoption of more sensitive information-theoretic measures for CEMs, such as \eqref{eq_I_CT_CEM}, \eqref{eq_I_IC_CEM} and \eqref{eq_I_align_CEM}.

\vskip 0.2in

\bibliography{bibliography}

\end{document}